\documentclass{article}

\usepackage{microtype}
\usepackage{graphicx}
\usepackage{subcaption}
\usepackage{booktabs} 

\usepackage{hyperref}

\usepackage[main,final]{neurips_2026}

\usepackage{amsmath}
\usepackage{amssymb}
\usepackage{mathtools}
\usepackage{amsthm}
\usepackage{bm}

\usepackage[capitalize,noabbrev]{cleveref}

\newcommand{\R}{\mathbb{R}} 
\newcommand{\X}{\mathbb{X}} 

\usepackage{bm}
\newcommand{\vct}[1]{\bm{#1}} 
\newcommand{\mtx}[1]{\bm{#1}} 
\newcommand{\tns}[1]{\bm{\mathcal{#1}}} 

\usepackage{amsmath,amssymb}

\newcommand{\inner}[3][]{\ifx q#1q\langle #2, #3 \rangle\else\langle #2, #3 \rangle_{#1}\fi} 
\newcommand{\norm}[2][]{\ifx q#1q\left\lVert#2\right\rVert\else\left\lVert#2\right\rVert_{#1}\fi} 

\usepackage[utf8]{inputenc} 
\usepackage{tikz}
\usepackage{csquotes}
\usetikzlibrary{calc,arrows.meta}
\MakeOuterQuote{"}
\usepackage{multirow}
\usepackage{makecell}
\usepackage{tabularx}
\usetikzlibrary{decorations.pathreplacing}
\usepackage{enumitem}
\setlist[description]{labelindent=0pt, labelsep=0.5em, leftmargin=0pt}
\usepackage{xcolor}
\usepackage{siunitx}
\usepackage{wrapfig}
\usepackage{pgfplots}
\pgfplotsset{compat=1.18}
\newcolumntype{Y}{>{\centering\arraybackslash}X}
\usepackage{array}
\newcolumntype{C}[1]{>{\centering\arraybackslash}p{#1}}

\theoremstyle{plain}

\theoremstyle{definition}

\theoremstyle{remark}

\usepackage[T1]{fontenc}    
\usepackage{hyperref}       
\usepackage{url}            
\usepackage{booktabs}       
\usepackage{adjustbox}      
\usepackage{amsfonts}       
\usepackage{nicefrac}       
\usepackage{microtype}      
\usepackage{xcolor}         
\newif\ifshowrev
\showrevfalse
\ifshowrev
  \newcommand{\rev}[1]{{\color{blue}#1}}
  \newenvironment{revblock}{\color{blue}}{}
\else
  \newcommand{\rev}[1]{#1}
  \newenvironment{revblock}{}{}
\fi
\hypersetup{hidelinks}  

\usepackage[textsize=tiny]{todonotes}

\author{
Niels Vyncke \quad Pooya Ashtari \quad Aleksandra Pižurica\\
Department of Telecommunications and Information Processing\\
Ghent University, Belgium\\
\texttt{\{niels.vyncke, pooya.ashtari, aleksandra.pizurica\}@ugent.be}
}

\title{LARGO: Low-Rank Hypernetwork for Handling Missing Modalities}

\begin{document}
\maketitle
\begin{abstract}
Addressing missing modalities is an important challenge in multimodal image analysis and often relies on complex architectures that do not transfer easily to different datasets without architectural modifications or hyperparameter tuning. While most existing methods tackle this problem in \emph{feature} space by engineering representations that are robust to missing inputs, we instead operate in \emph{weight} space. We propose LARGO, a hypernetwork that compresses the $2^N-1$ dedicated missing-modality models into a single network by modelling the convolutional weights using the Canonical Polyadic (CP) tensor decomposition. Extensive experimental validation on BraTS 2018 (4 modalities, 15 scenarios) and ISLES 2022 (3 modalities, 7 scenarios) shows that our method ranks first in 47 out of 52 configurations, achieving average Dice improvements of +0.68\% and +2.53\% over state-of-the-art baselines (mmFormer, M\textsuperscript{3}AE, ShaSpec, SimMLM). A proof-of-concept experiment on avMNIST suggests that LARGO may extend beyond medical imaging to heterogeneous non-medical modalities. Code is available at \url{https://github.com/nielsvyncke/LARGO}.
\end{abstract}
\section{Introduction}
\label{sec:intro}

Multimodal imaging combines complementary information from different acquisition protocols to improve accuracy~\citep{baltruvsaitis2018multimodal,bayoudh2022survey}. In brain tumor segmentation, T2-weighted imaging reveals edema through fluid sensitivity, while contrast-enhanced T1-weighted imaging highlights active tumor regions through structural disruption~\citep{brats1,brats2,brats3}. The synergy of multiple modalities enables precise delineation of tumor subregions that would be ambiguous from any single modality alone.

However, acquiring all modalities is not always feasible. Equipment availability, scan time constraints, patient conditions, and clinical protocols frequently result in incomplete modality sets~\citep{wu2024deepmultimodallearningmissing,azad2022medicalimagesegmentationmri}. Standard deep learning models cannot handle such missing inputs without explicit design considerations, creating a critical gap between research benchmarks---which assume complete data---and clinical deployment scenarios.

Existing approaches to missing modalities face a trade-off between \textit{specialization}---optimal performance on a specific subset---and \textit{scalability}---efficient support of all $2^N-1$ combinations. Single-model methods such as zero-imputation~\citep{pemberton_multi-class_2023} or shared-encoder architectures~\citep{havaei_hemis_2016,dorent_hetero-modal_2019,zhang_mmformer_2022} force one network to cover all combinations, compromising specific configurations; generative approaches~\citep{chartsias_multimodal_2018,dar_image_2019} risk hallucinating clinically dangerous features; and knowledge distillation~\citep{maheshwari_missing_2023,wang_learnable_2023} requires multi-stage training. Training dedicated models per combination maximizes specialization but scales exponentially in storage and compute.

\begin{figure*}[!t]
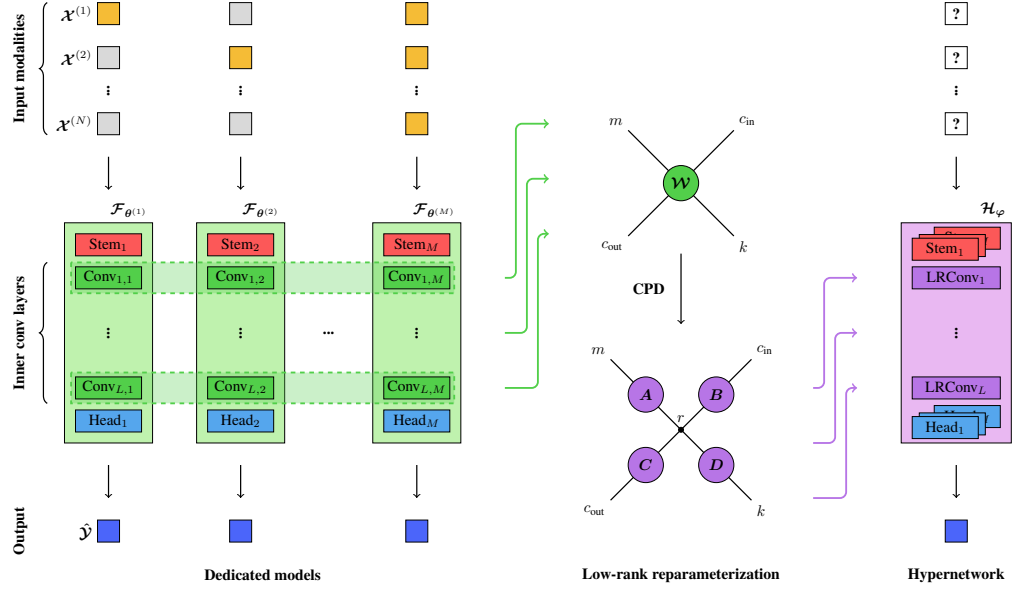

\centering
\begin{subfigure}{\linewidth}
    \centering
    \resizebox{0.95\linewidth}{!}{\input{Fig/methodology/overview-a}}
    \caption{\rev{Network level: the $M$ dedicated models are compressed into one hypernetwork.}}
    \label{fig:hypernetwork}
\end{subfigure}
\\[0.5em]
\begin{subfigure}{\linewidth}
    \centering
    \resizebox{\linewidth}{!}{
\begin{tikzpicture}[font=\small]
    \input{Fig/methodology/palette}
    \draw[panel] (0,0) rectangle (5.6,4.1);
    \node[num] at (0.4,3.7) {1};
    \node[ptitle] at (0.7,3.7) {Stack dedicated kernels};
    \foreach \i/\c in {3/Leaf, 2/Leaf!80!white, 1/Leaf!60!white, 0/Leaf!40!white} {
        \draw[kern, fill=\c] ({0.4+0.35*\i},{0.8+0.35*\i}) rectangle ({1.8+0.35*\i},{2.2+0.35*\i});
    }
    \node[lab] at (1.1,1.5) {$\tns{V}^{[1]}$};
    \node[lab] at (1.45,2.37) {$\tns{V}^{[2]}$};
    \node[lab] at (1.8,2.72) {$\cdots$};
    \node[lab] at (2.15,3.07) {$\tns{V}^{[M]}$};
    \draw[flow] (2.0,0.6) -- (3.05,1.65);
    \node[note, anchor=north east] at (1.95,0.6) {model mode $m$};
    \draw[flow] (3.0,2.15) -- (3.65,2.15) node[midway, above, note] {stack};
    \begin{scope}[shift={(4.45,2.3)}]
        \draw[leg] (0,0) -- (-0.55,0.55) node[lab, anchor=south east] {$m$};
        \draw[leg] (0,0) -- (0.55,0.55) node[lab, anchor=south west] {$c_\text{in}$};
        \draw[leg] (0,0) -- (-0.55,-0.55) node[lab, anchor=north east] {$c_\text{out}$};
        \draw[leg] (0,0) -- (0.55,-0.55) node[lab, anchor=north west] {$k$};
        \node[Wfac] at (0,0) {$\tns{W}$};
    \end{scope}
    \node[note, anchor=north east] at (5.45,1.15) {$\tns{W}\in\R^{M\times C_\text{in}\times C_\text{out}\times K}$};

    \draw[big] (5.75,2.05) -- (6.15,2.05);
    \draw[panel] (6.3,0) rectangle (11.7,4.1);
    \node[num] at (6.7,3.7) {2};
    \node[ptitle] at (7.0,3.7) {Factorize with CP};
    \node[lab] at (8.6,3.2) {$\tns{W}=\sum_{r=1}^{R}\vct{a}_r\circ\vct{b}_r\circ\vct{c}_r\circ\vct{d}_r$};
    \draw[leg] (7.5,1.75) -- (6.8,1.75) node[lab, anchor=east] {$m$};
    \draw[leg] (7.5,1.75) -- (8.8,1.75);
    \draw[leg] (10.2,2.55) .. controls (9.4,2.55) .. (8.8,1.75);
    \draw[leg] (10.2,1.75) -- (8.8,1.75);
    \draw[leg] (10.2,0.95) .. controls (9.4,0.95) .. (8.8,1.75);
    \draw[leg] (10.2,2.55) -- (10.9,2.55) node[lab, anchor=west] {$c_\text{in}$};
    \draw[leg] (10.2,1.75) -- (10.9,1.75) node[lab, anchor=west] {$c_\text{out}$};
    \draw[leg] (10.2,0.95) -- (10.9,0.95) node[lab, anchor=west] {$k$};
    \fill[black] (8.8,1.75) circle (0.05);
    \node[lab, anchor=south] at (8.8,1.8) {$r$};
    \node[Afac] at (7.5,1.75) {$\mtx{A}$};
    \node[Bfac] at (10.2,2.55) {$\mtx{B}$};
    \node[Bfac] at (10.2,1.75) {$\mtx{C}$};
    \node[Bfac] at (10.2,0.95) {$\mtx{D}$};
    \node[note, anchor=north] at (7.5,1.25) {model factor\\$\mtx{A}\in\R^{M\times R}$};
    \node[note, anchor=north west] at (6.5,0.55) {kernel factors $\mtx{B},\mtx{C},\mtx{D}$};

    \draw[big] (11.85,2.05) -- (12.25,2.05);
    \draw[panel] (12.4,0) rectangle (17.9,4.1);
    \node[num] at (12.8,3.7) {3};
    \node[ptitle] at (13.1,3.7) {Reconstruct only $\tns{W}^{(m)}$};
    \node[lab, anchor=west] at (12.65,3.15) {available modalities $\{\bullet\,\circ\,\bullet\,\circ\}\;\Rightarrow\; m$};
    \draw[kern, fill=AmberLight!45!white] (12.7,0.95) rectangle (13.7,2.65);
    \foreach \y in {1.15,1.35,...,2.5} {\draw[Amber!70!white] (12.7,\y) -- (13.7,\y);}
    \draw[fill=Amber, draw=Amber!60!black, thick] (12.7,1.7) rectangle (13.7,1.9);
    \node[lab, anchor=south] at (13.2,2.67) {$\mtx{A}$};
    \node[note, anchor=north] at (13.2,0.9) {$M$ rows};
    \draw[flow] (13.75,1.8) -- (14.55,1.8) node[midway, above, lab, Amber!60!black] {$\mtx{A}_{m,:}$};
    \node[fac, fill=Amber!85!white, minimum size=0.6cm, font=\footnotesize] at (14.9,1.8) {$\vct{a}_m$};
    \draw[leg] (15.2,1.8) -- (15.6,1.8);
    \draw[leg] (16.3,2.55) .. controls (15.85,2.55) .. (15.6,1.8);
    \draw[leg] (16.3,1.8) -- (15.6,1.8);
    \draw[leg] (16.3,1.05) .. controls (15.85,1.05) .. (15.6,1.8);
    \fill[black] (15.6,1.8) circle (0.05);
    \node[Bfac, minimum size=0.55cm, font=\footnotesize] at (16.3,2.55) {$\mtx{B}$};
    \node[Bfac, minimum size=0.55cm, font=\footnotesize] at (16.3,1.8) {$\mtx{C}$};
    \node[Bfac, minimum size=0.55cm, font=\footnotesize] at (16.3,1.05) {$\mtx{D}$};
    \draw[flow] (16.6,1.8) -- (16.95,1.8);
    \node[Wfac, minimum size=0.6cm, font=\footnotesize] at (17.3,1.8) {$\tns{W}^{(m)}$};
    \node[note, anchor=north] at (15.15,0.62) {$\tns{W}^{(m)}=\sum_{r} A_{m,r}\,\vct{b}_r\circ\vct{c}_r\circ\vct{d}_r$\\kernel of subset $m$ only};
\end{tikzpicture}%
}
    \caption{\rev{Layer level: the low-rank reparameterization block of (a) for one convolutional layer.}}
    \label{fig:conv-reparameterization}
\end{subfigure}
\caption{Illustration of LARGO. \rev{(a)} The dedicated models $\tns{F}_{\vct{\theta}^{(m)}}$ for each missing-modality case $m=1,...,M$ are compressed into a single hypernetwork $\tns{H}_{\vct{\varphi}}$, by representing the kernel weights $\tns{W}$ of their convolutional layers using CPD decomposition, which we refer to as low-rank convolutional layers (LRConv). The stem and head layers remain uncompressed. \rev{(b) For each such layer, (1) the kernels of the $M=2^N-1$ dedicated models are stacked along a new \emph{model} mode into $\tns{W}$; (2) $\tns{W}$ is factorized with the CP decomposition into a model factor $\mtx{A}$ and kernel factors $\mtx{B},\mtx{C},\mtx{D}$, trained jointly from scratch; (3) at inference, the available subset $m$ selects row $\mtx{A}_{m,:}$, and only the kernel $\tns{W}^{(m)}$ of that subset is reconstructed.}}
\label{fig:overview}
\end{figure*}

We propose a different approach that achieves \textit{both} specialization and scalability. Our key observation is that dedicated models for different modality combinations are not independent: models sharing access to the same modalities must learn similar patterns, implying substantial parameter redundancy across the family. We exploit this redundancy through tensor decomposition.

Our approach, LARGO (\textbf{L}ow-r\textbf{A}nk hype\textbf{R}network for missin\textbf{G} m\textbf{O}dalities), constructs a hypernetwork by stacking the convolutional weights of all $2^{N}-1$ dedicated models along a new \emph{model} dimension and applying Canonical Polyadic (CP) decomposition. The entire family is represented through compact factor matrices, with one factor indexing modality combinations, yielding a single network that implicitly contains all dedicated models at the parameter budget of just one.

Unlike feature-space methods, LARGO operates in \emph{weight} space and is the first to represent the exponential family of $2^{N}-1$ missing-modality models within a single compressed weight manifold---a formulation absent from both prior tensor-decomposition compression of individual networks~\citep{ashtari2020low,kim_compression_2015,lebedev_speeding_2014} and hypernetwork literature~\citep{ha_hypernetworks_2017,zhang_graph_2019}. It does not rely on auxiliary losses, teacher networks, or multi-stage training. \rev{The result is a jointly optimized low-rank manifold of $2^N-1$ subset-specific networks (\cref{fig:conv-reparameterization}).}

Our contributions are:
\begin{itemize}
    \item A novel hypernetwork formulation that jointly reparameterizes the kernel weights of the entire family of $2^N-1$ dedicated missing-modality models into a single parameter-efficient representation using canonical polyadic decomposition.
    \item State-of-the-art results on BraTS 2018 and ISLES 2022: LARGO ranks first in 47 out of 52 missing-modality scenarios, with +0.68\% and +2.53\% average Dice gains over mmFormer, M³AE, ShaSpec, and SimMLM.
    \item A proof-of-concept on avMNIST (image+audio) showing that LARGO extends to non-medical, heterogeneous-modality classification, with +0.44\% average accuracy over the baselines.
\end{itemize}

\paragraph{Notation.}
Vectors are denoted by bold lowercase (e.g., $\vct{x}$), matrices by bold uppercase (e.g., $\mtx{U}$), tensors by calligraphic bold uppercase (e.g., $\tns{W}$), and sets by blackboard bold uppercase (e.g., $\X$). Individual elements are indexed by subscripts in non-bold notation: $x_i$, $U_{ij}$, $\mathcal{W}_{ijk}$.
\section{Related work} \label{sec:related_work}

\rev{We review prior work along three directions: incomplete multimodal learning, brain tumor segmentation under missing modalities, and hypernetworks with tensor-based network compression.} Existing missing-modality methods mainly learn robust feature representations, impute missing inputs, or distill knowledge from complete-modality models~\citep{wu2024deepmultimodallearningmissing,azad2022medicalimagesegmentationmri}. In contrast, LARGO operates in \emph{weight} space, jointly compressing the family of dedicated models associated with all $2^N-1$ non-empty modality subsets through a shared low-rank parameterization.

\begin{figure*}[!t]
\centering
\resizebox{0.8\linewidth}{!}{
\newcommand{\modelcol}[1]{\textbf{#1}}
\newcommand{\modalityrow}[1]{
\begin{tikzpicture}
\path (0,0)--(0,-2.35) node[midway,rotate=90] {\textbf{#1}};
\end{tikzpicture}
}

\newcommand{\img}[2]{%
\begin{tikzpicture}
    \node[inner sep=0pt] at (0,0)
        {\includegraphics[width=0.15\linewidth,angle=-90,trim={35 40 35 40},clip]{#1}};
    \if\relax\detokenize{#2}\relax
    \node[
      anchor=south east,
      font=\bfseries,
      fill=white,
      fill opacity=0,
      text opacity=1
    ] at (0.5,-1.15) {#2};
  \else
    \node[
      anchor=south east,
      font=\bfseries,
      fill=white,
      fill opacity=0.7,
      text opacity=1
    ] at (0.5,-1.15) {#2};
  \fi
\end{tikzpicture}
}

\begin{tabular}{c c c c c c c}
 & \hspace{-3pt}\modelcol{GT} & \hspace{-3pt}\modelcol{mmFormer} & \hspace{-3pt}\modelcol{M$^3$AE} & \hspace{-3pt}\modelcol{ShaSpec} & \hspace{-3pt}\modelcol{SimMLM} & \hspace{-3pt}\modelcol{LARGO (Ours)} \\
 \vspace{0.1cm}\\

\modalityrow{FLAIR only} &
\img{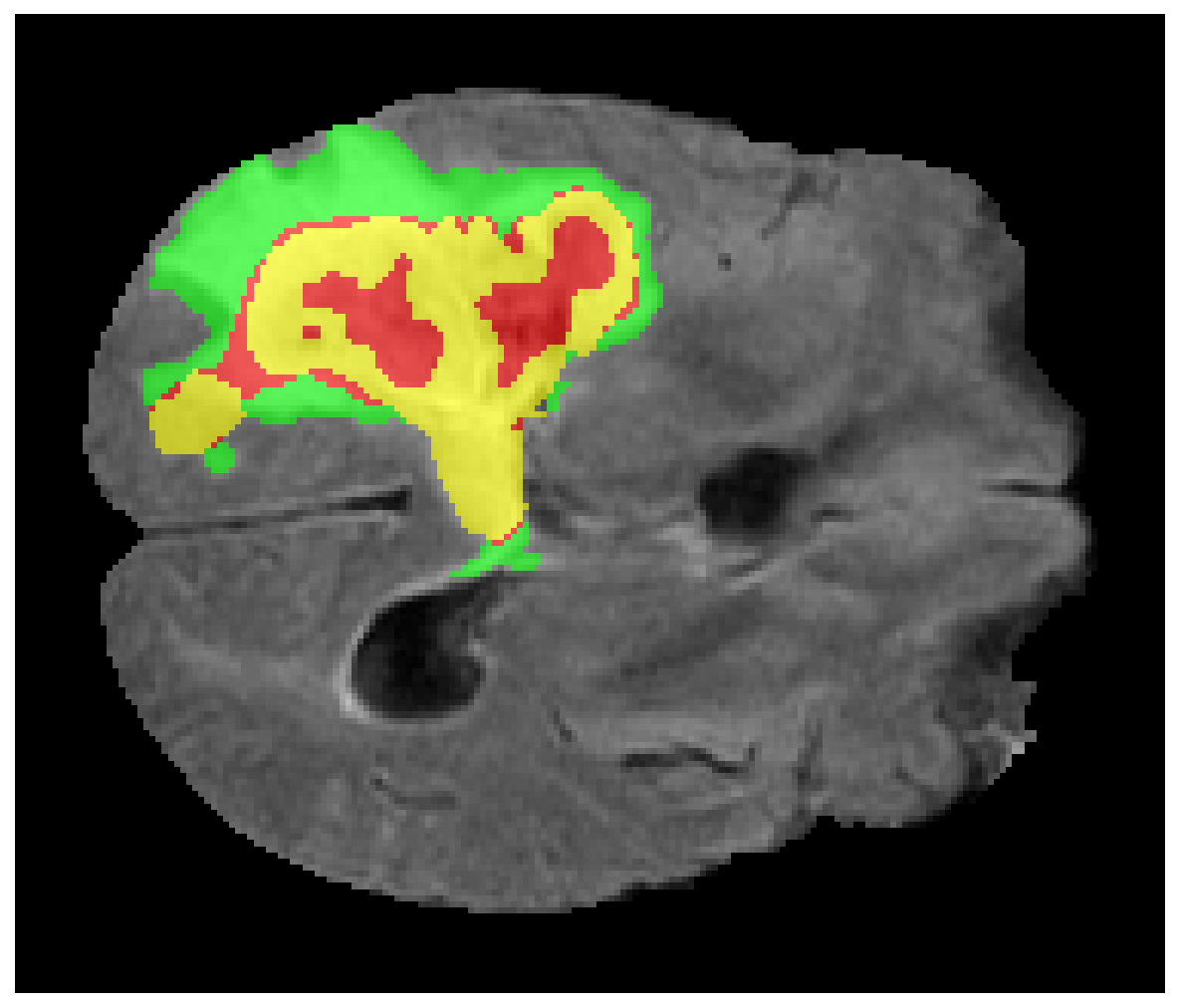}{} &
\img{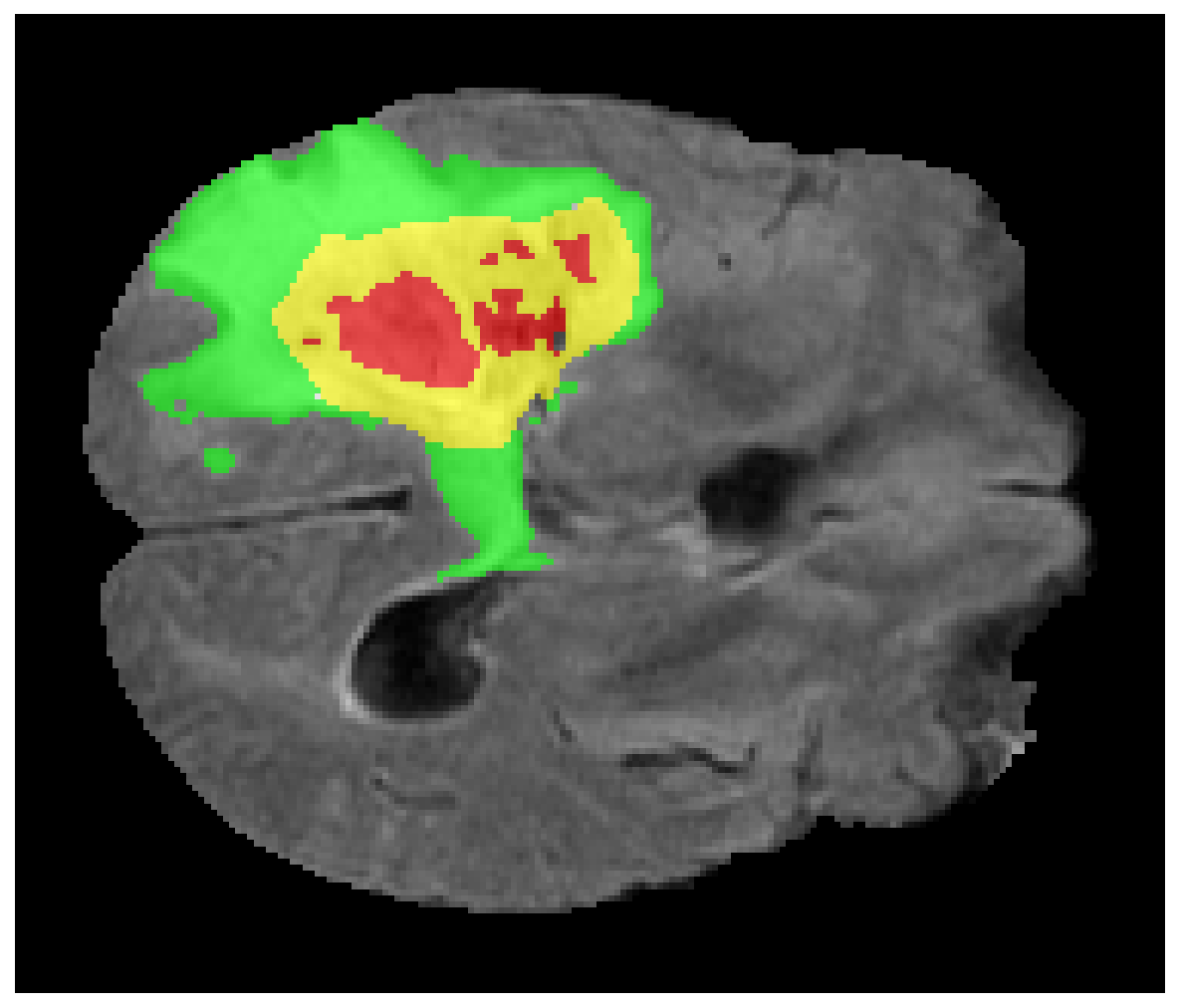}{80.34} &
\img{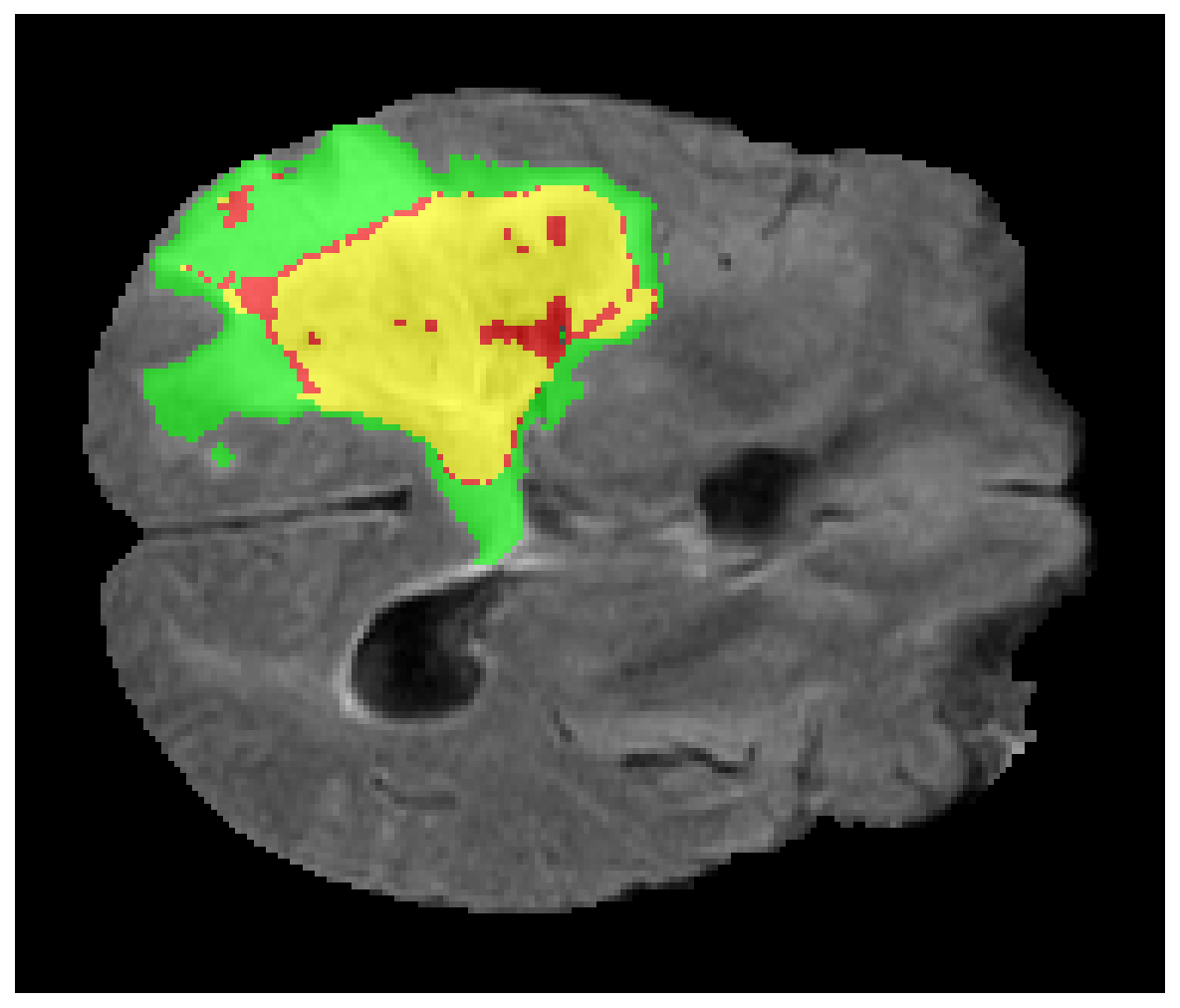}{80.74} &
\img{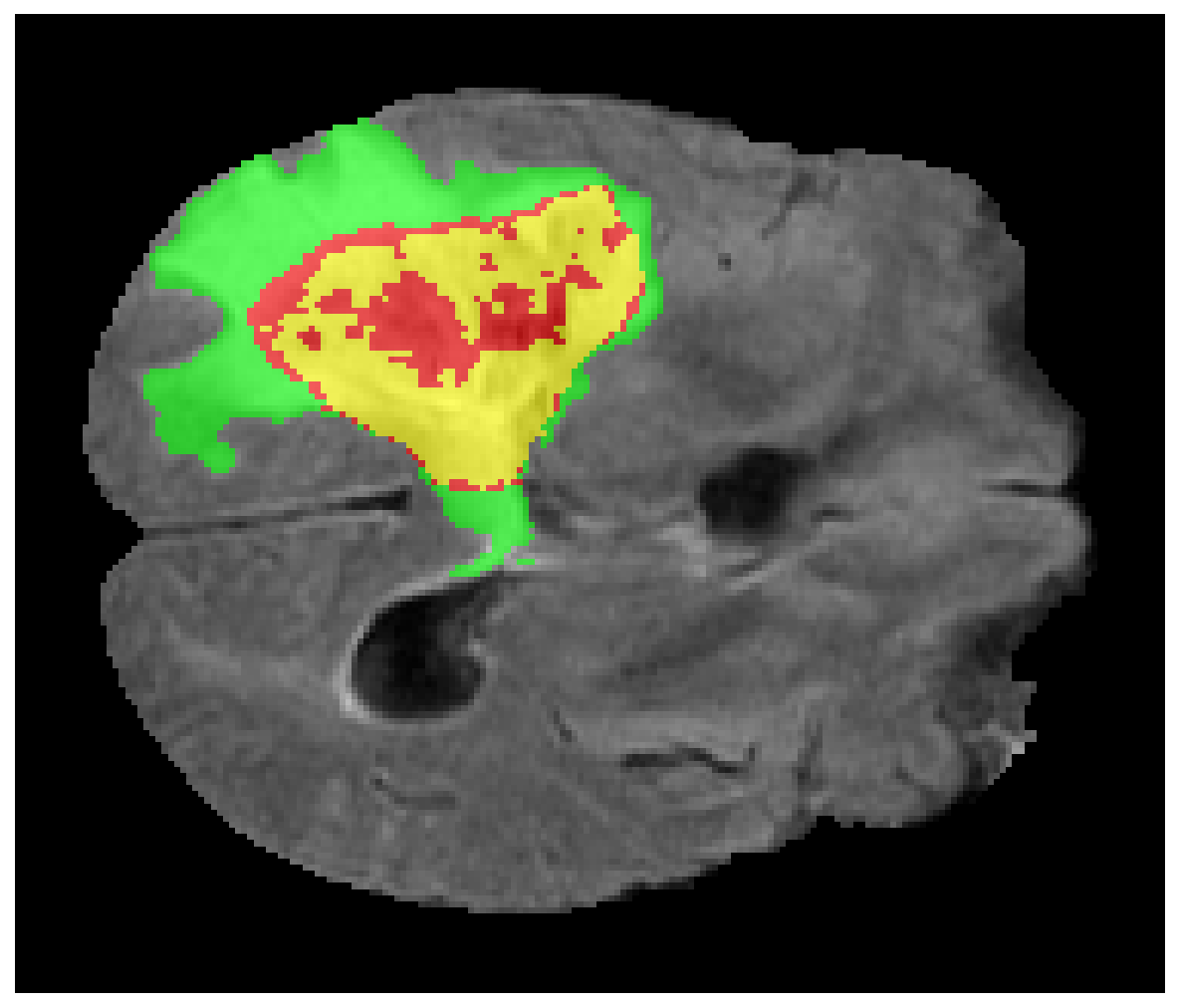}{81.99} &
\img{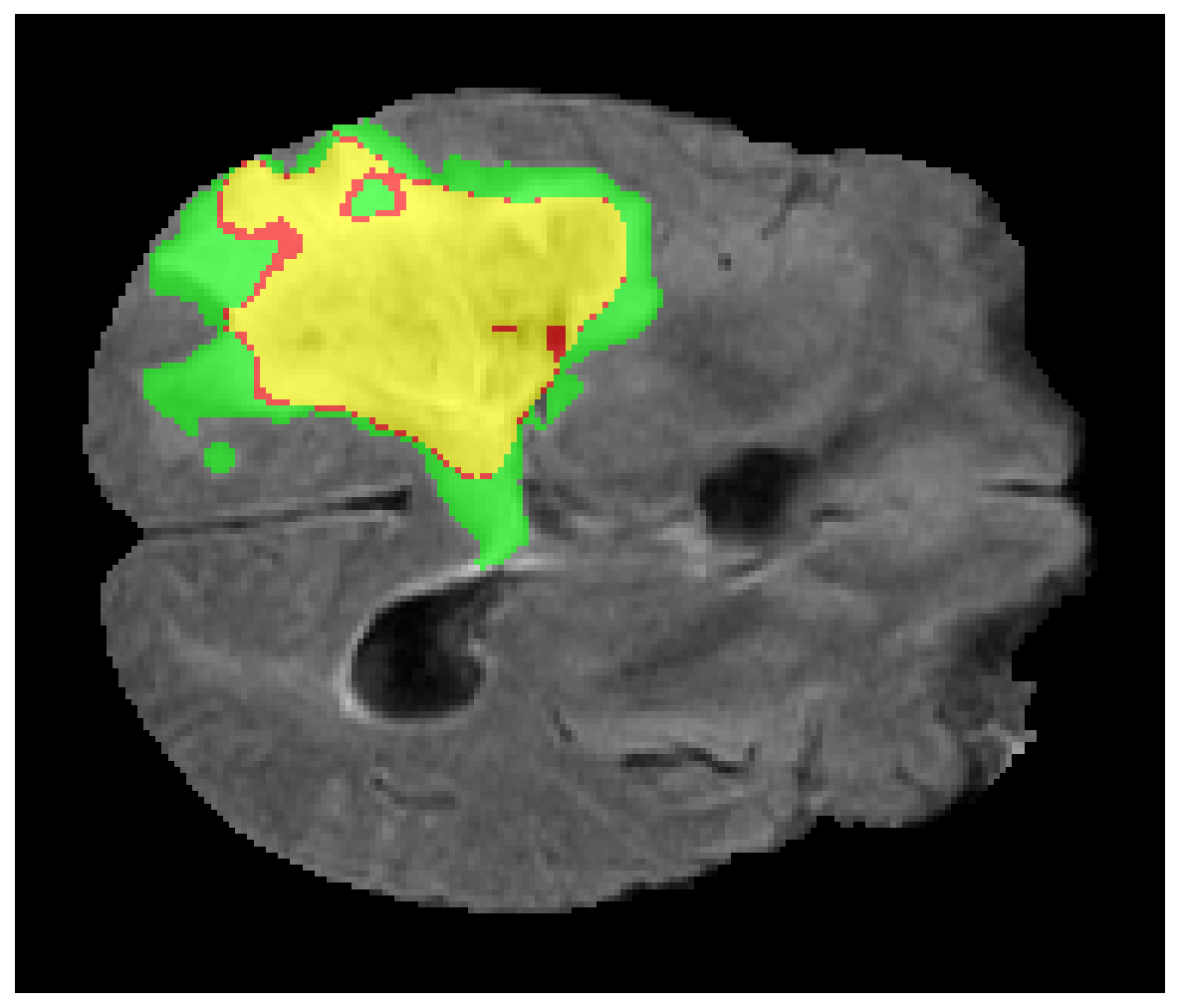}{79.74} &
\img{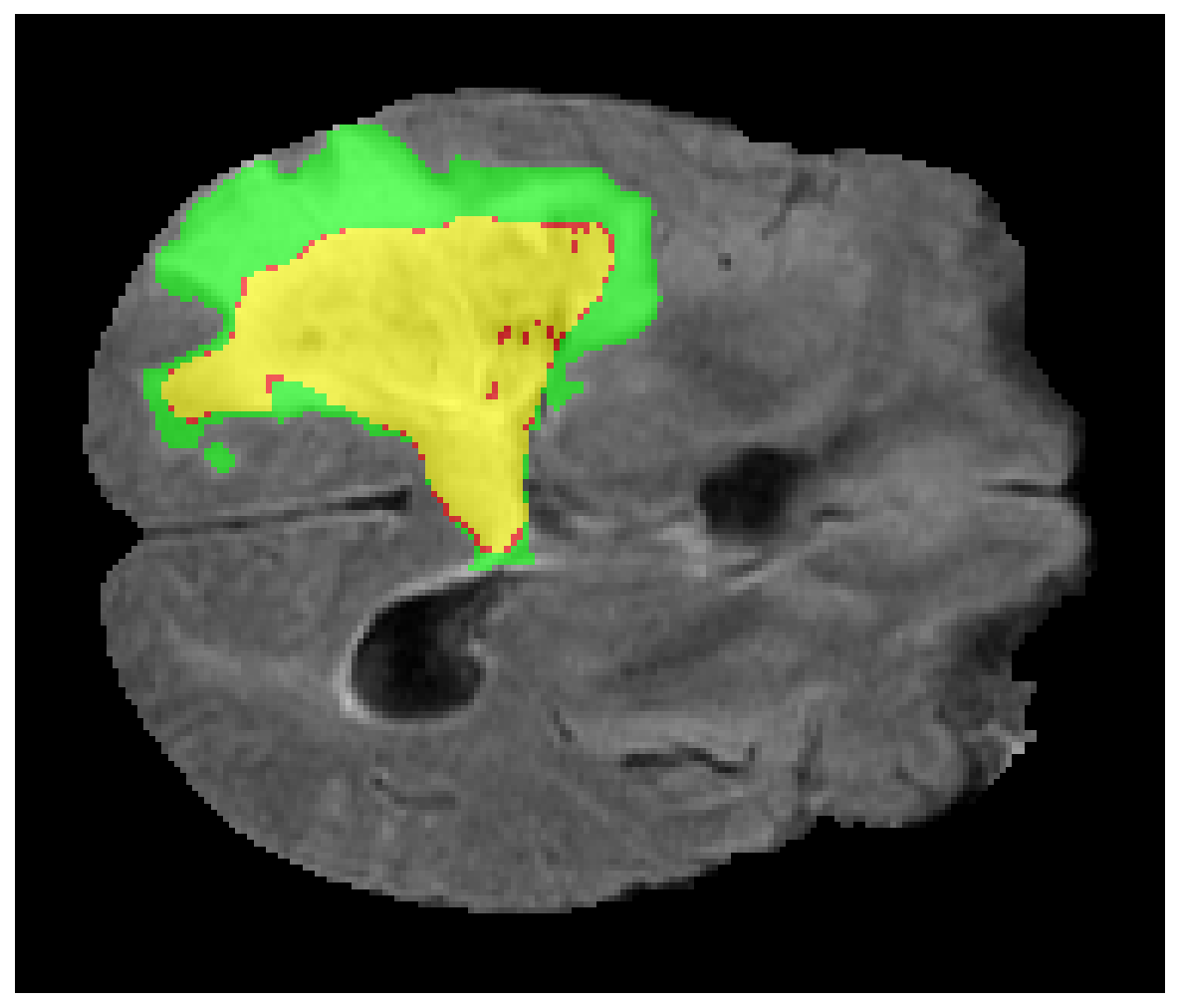}{84.94} \\

\modalityrow{T1 only} &
\img{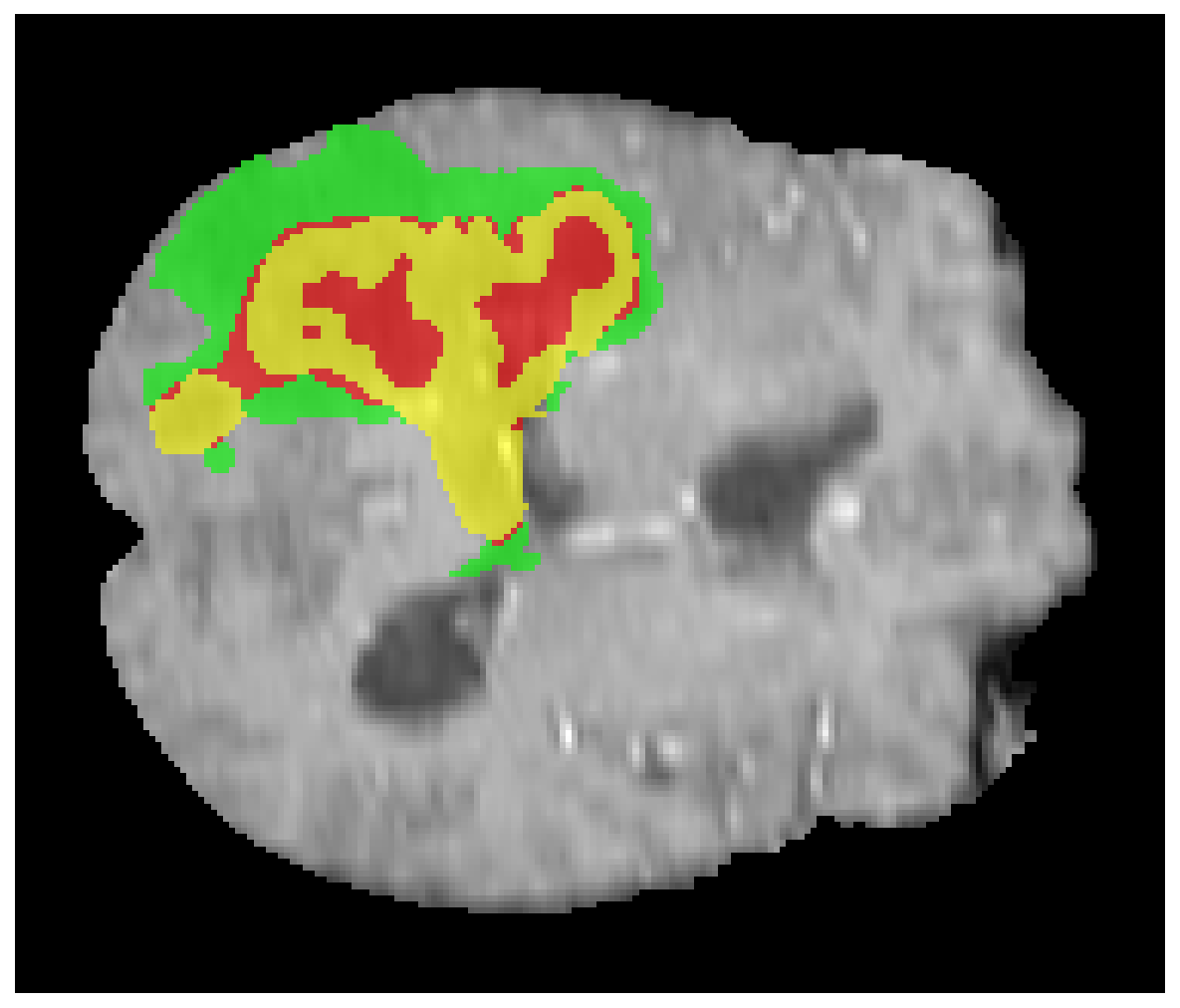}{} &
\img{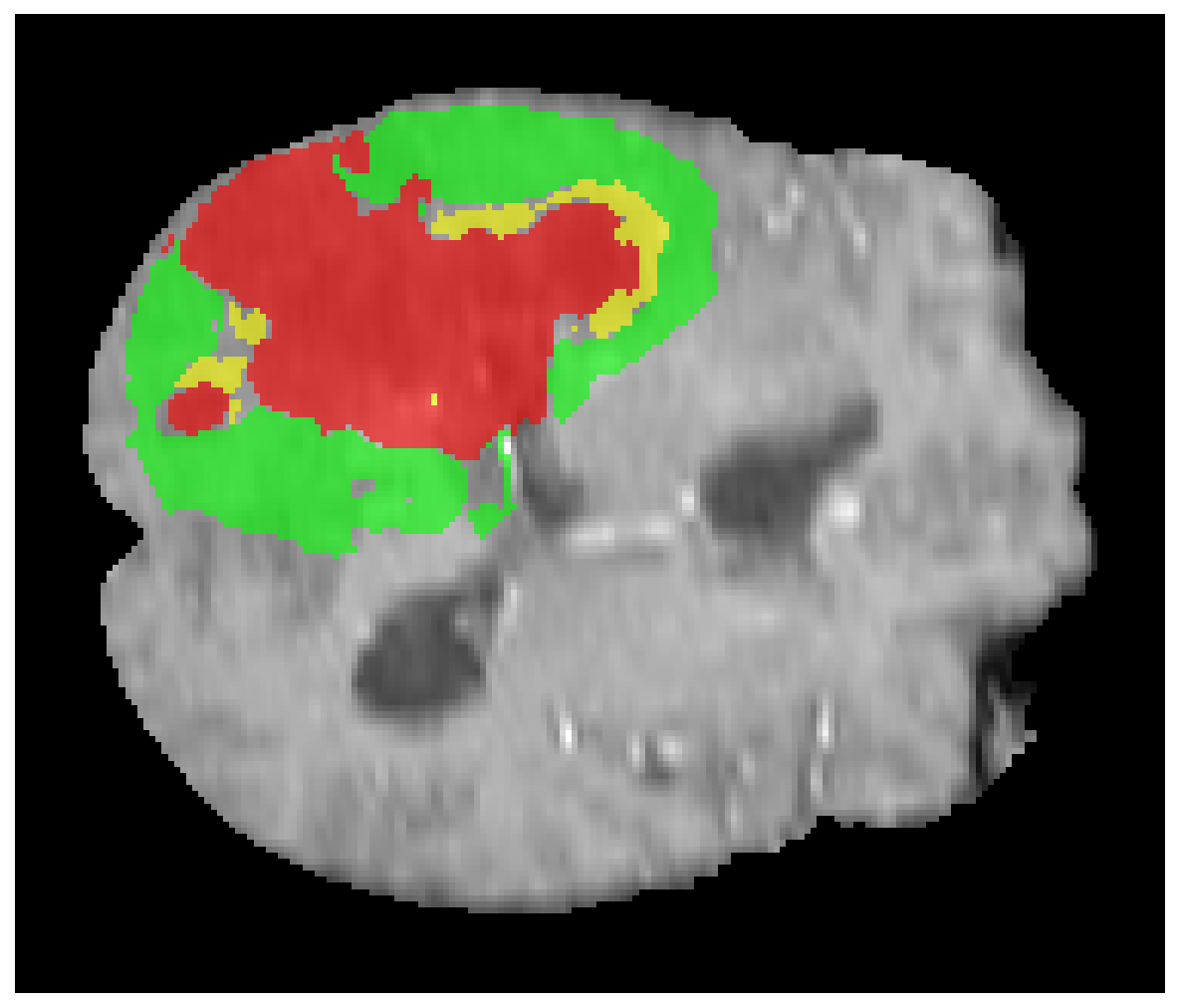}{54.68} &
\img{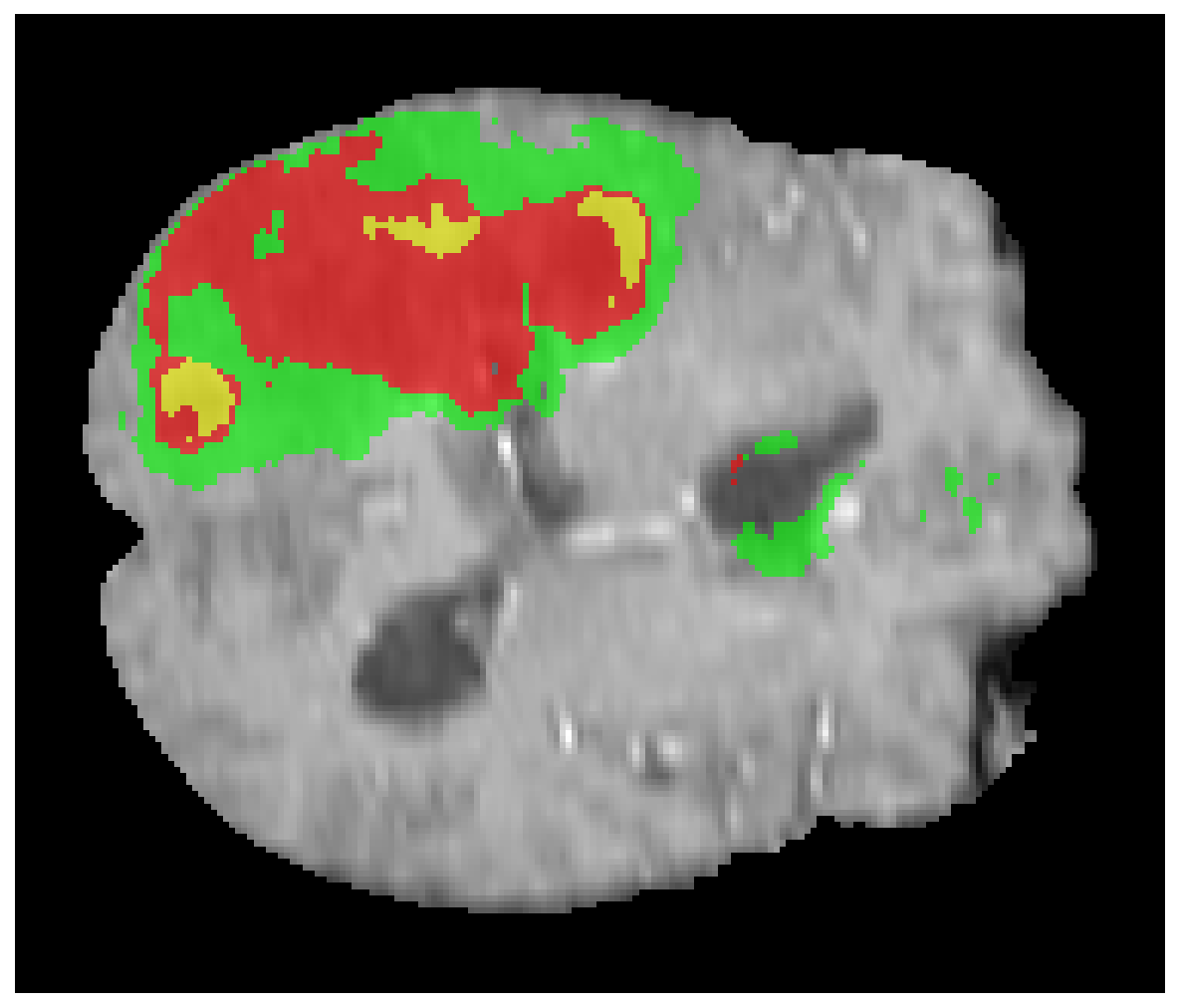}{58.73} &
\img{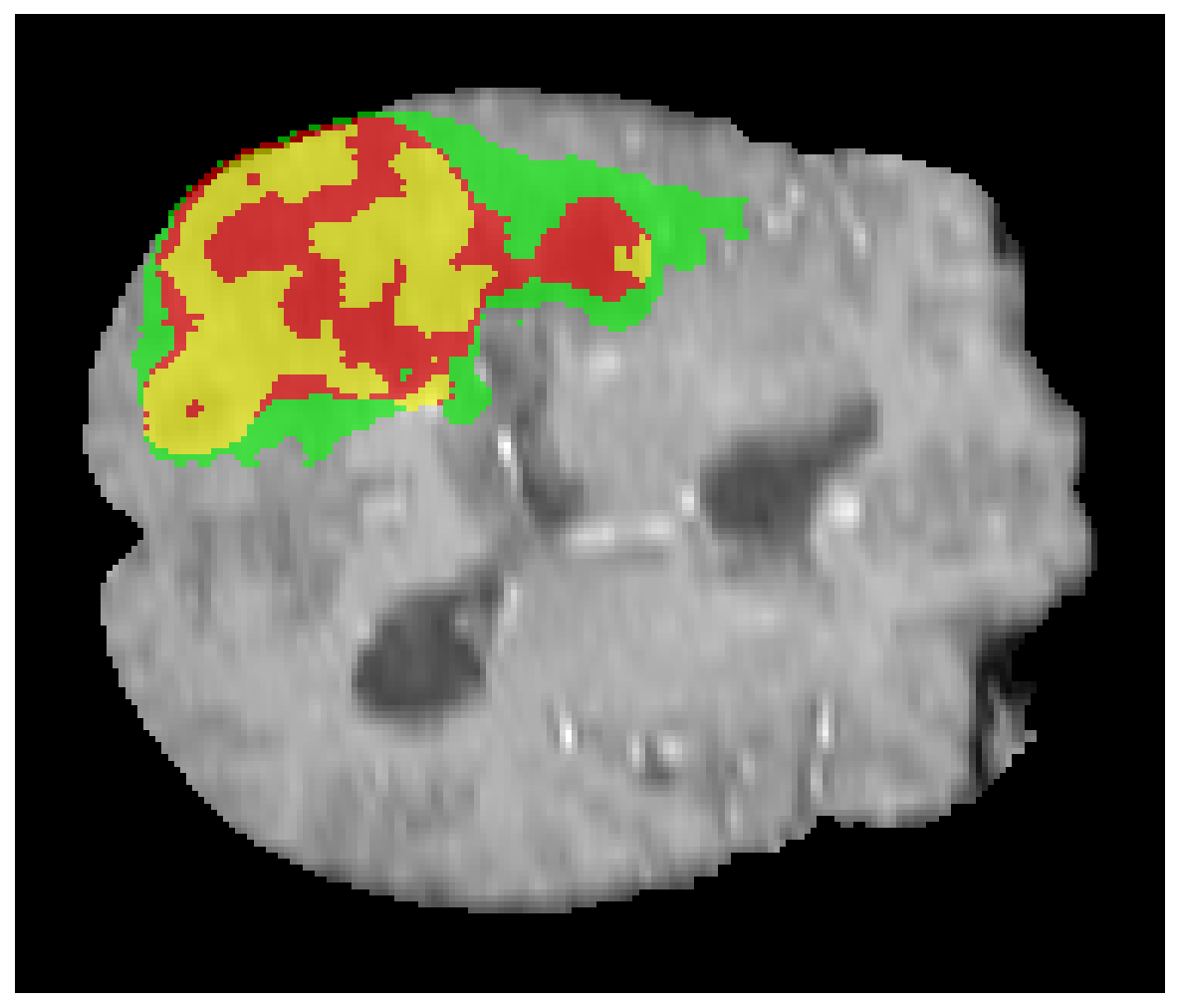}{62.44} &
\img{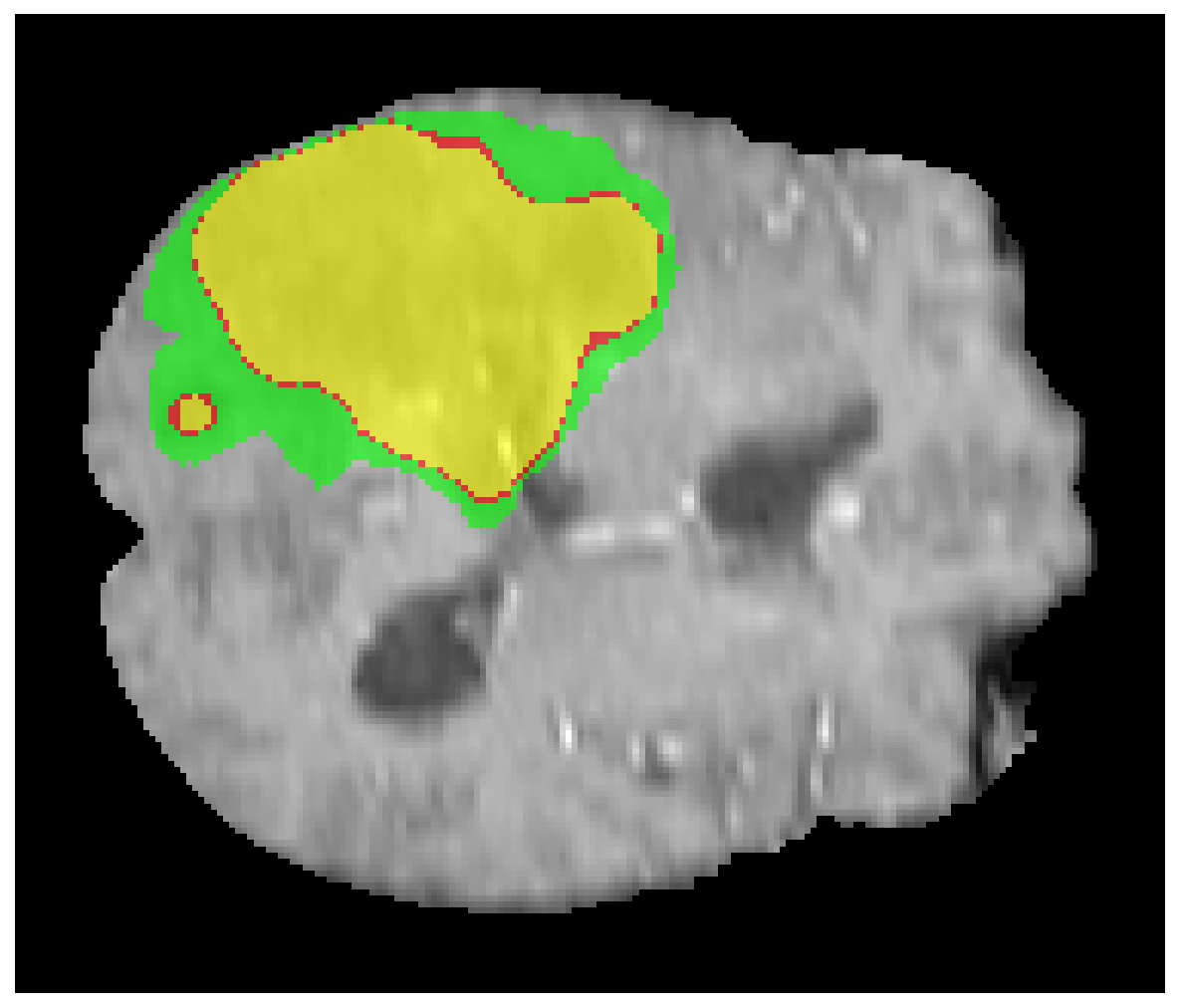}{74.48} &
\img{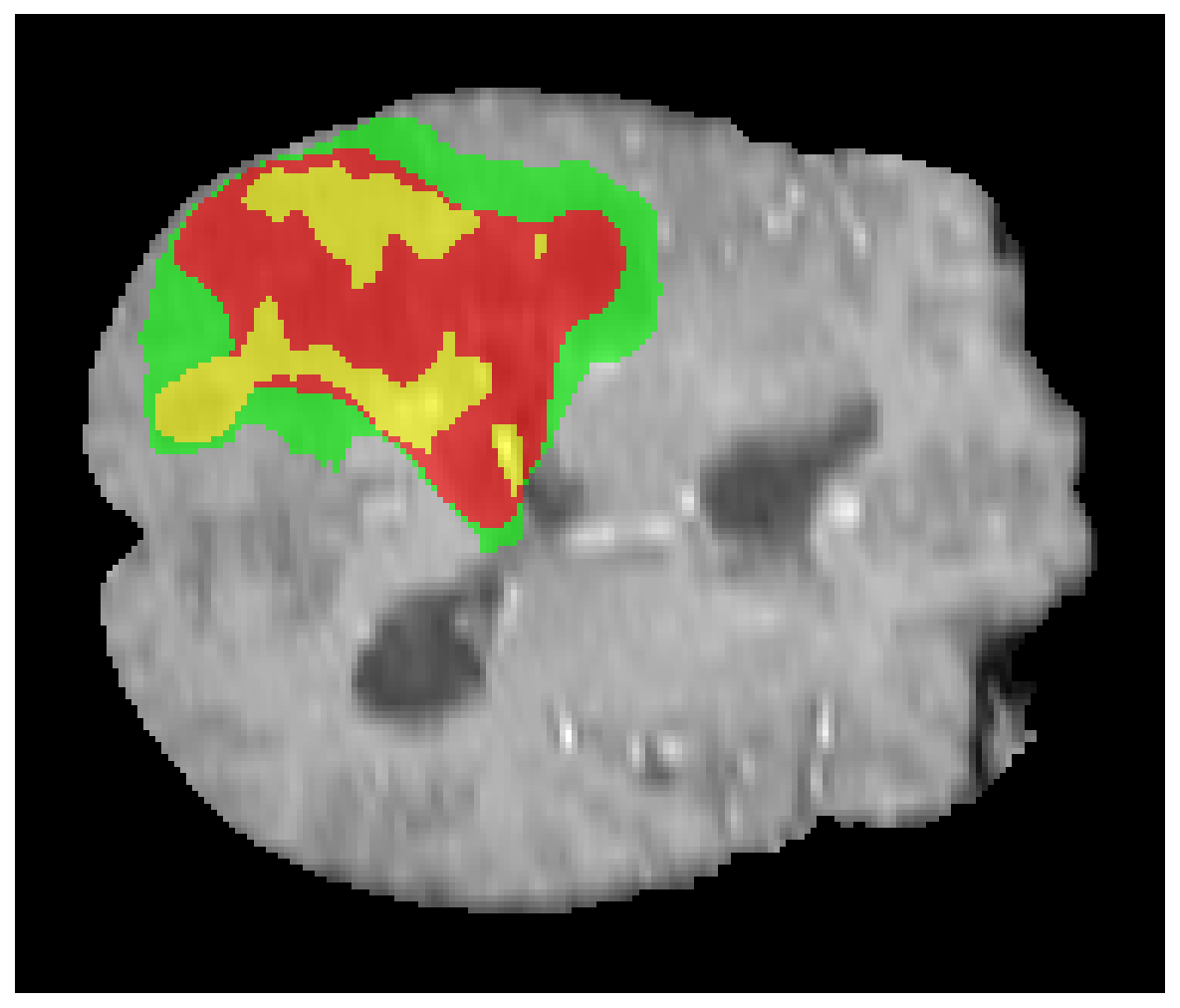}{73.37} \\

\modalityrow{T1c only} &
\img{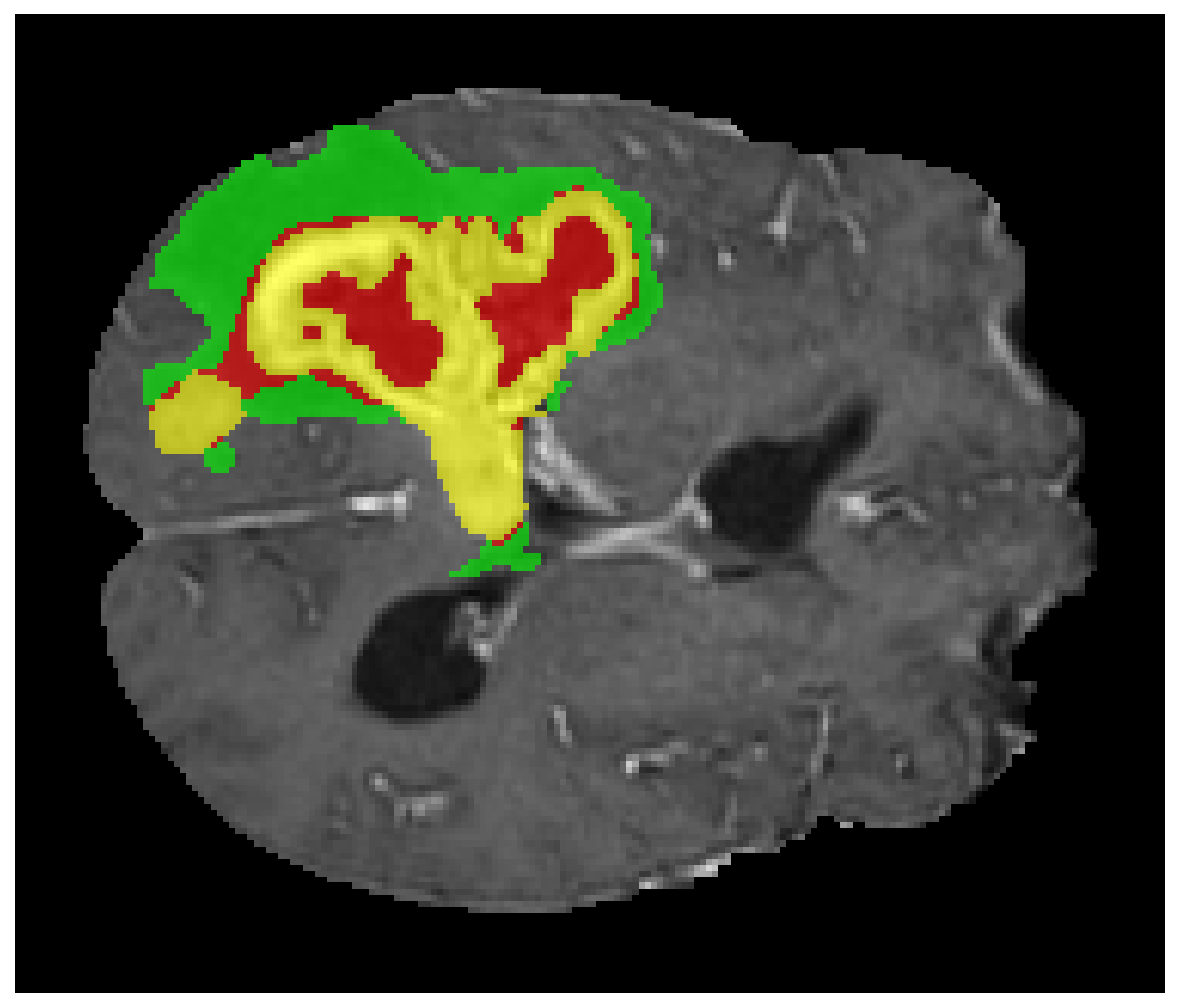}{} &
\img{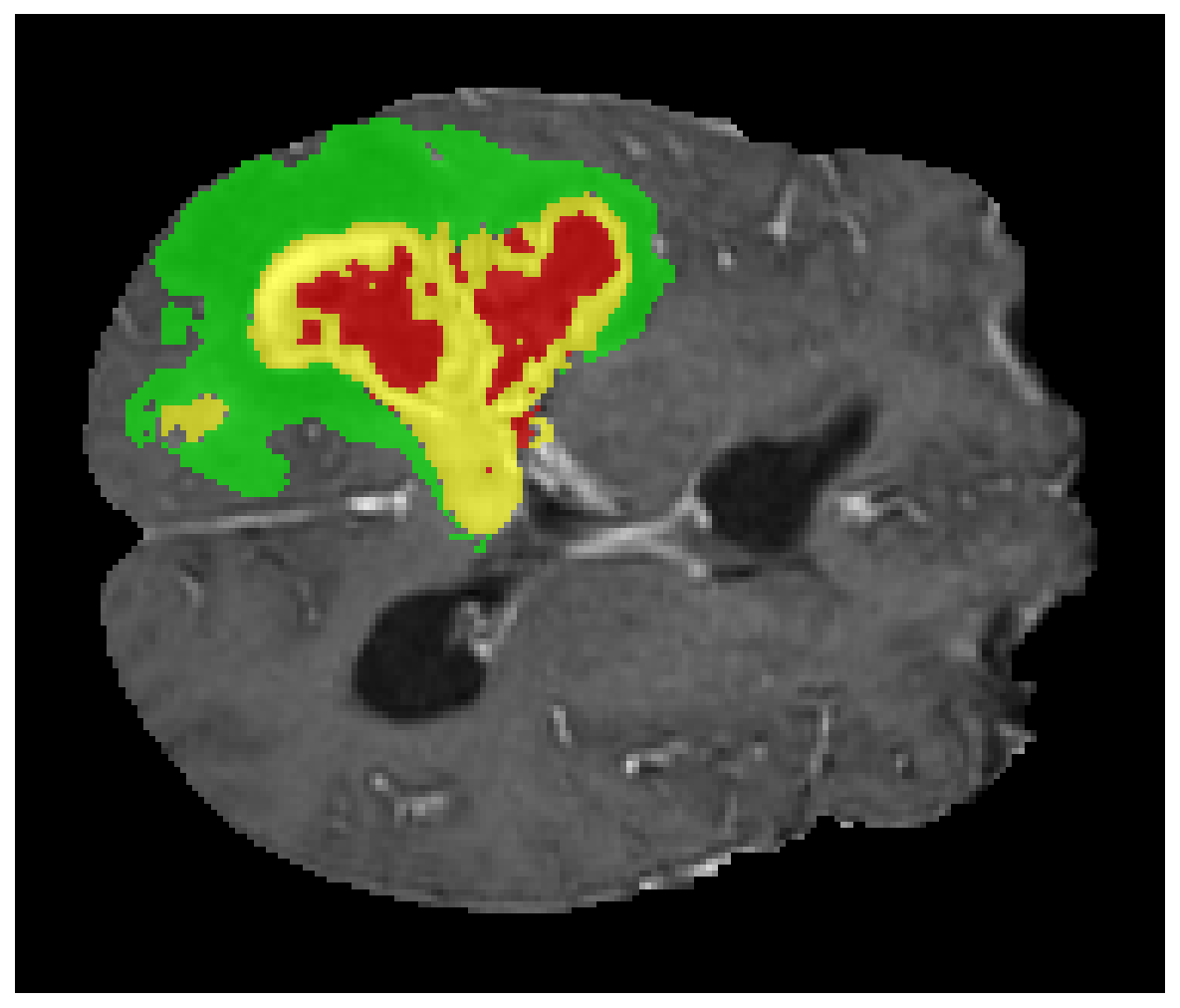}{88.96} &
\img{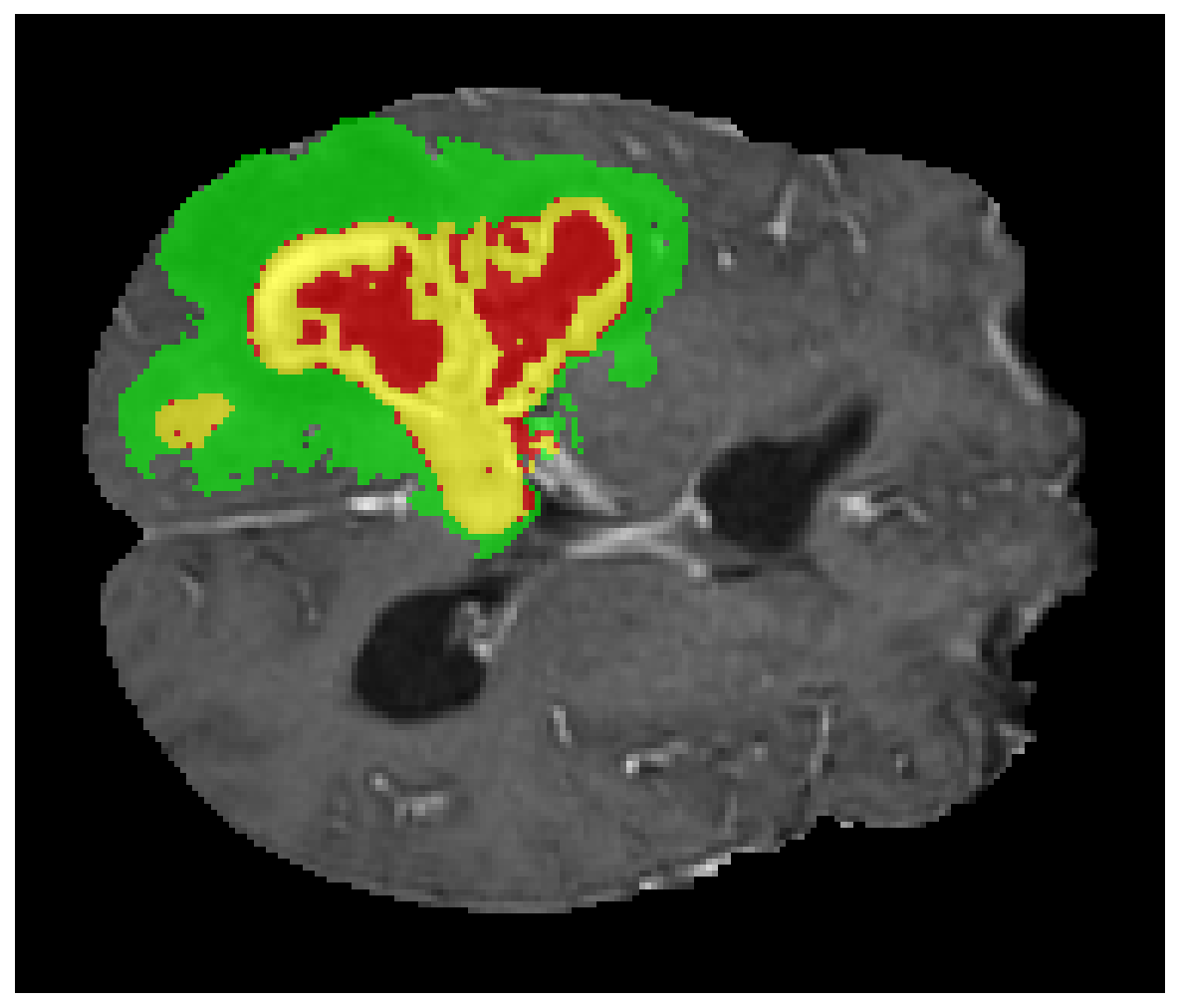}{88.99} &
\img{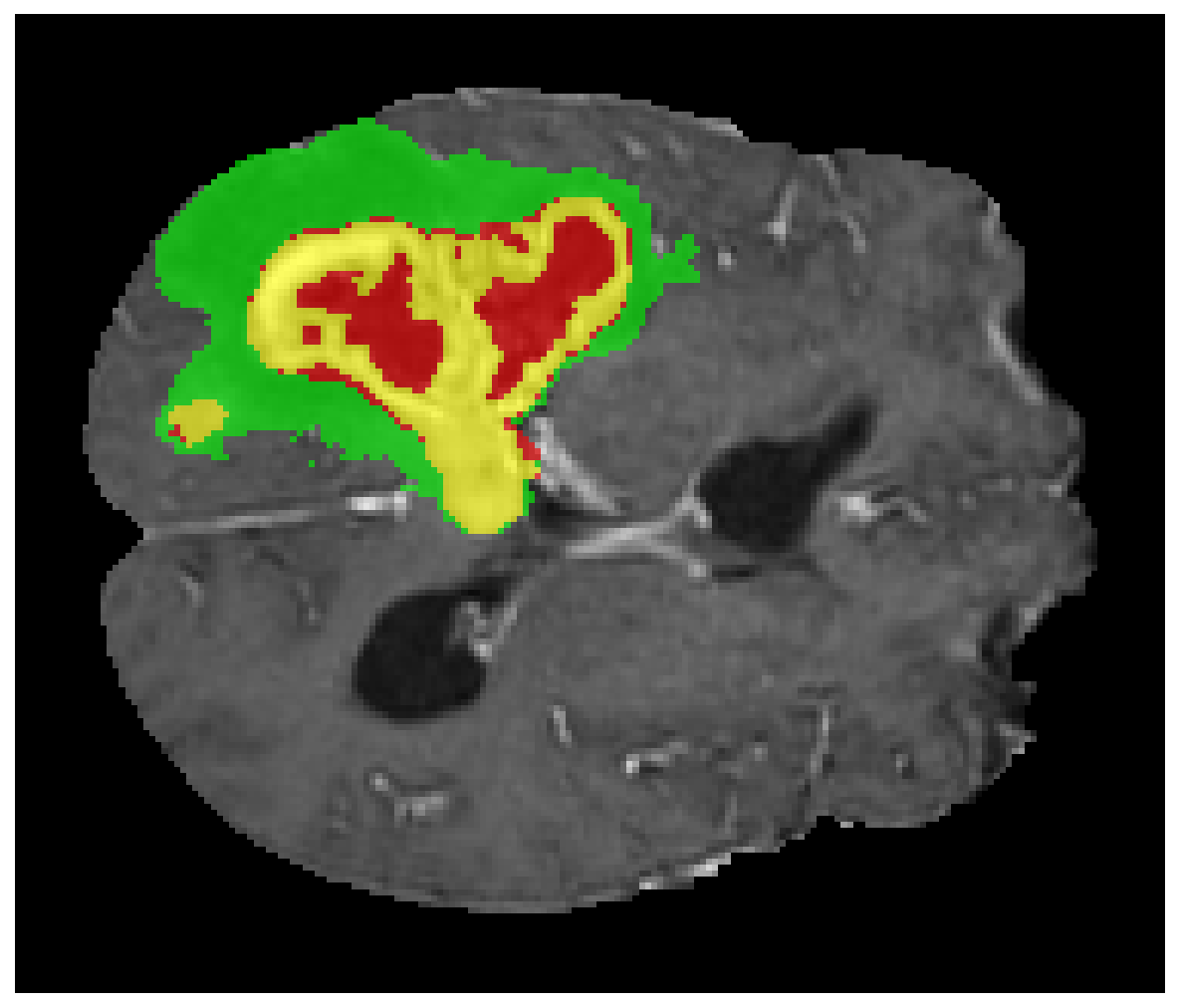}{90.89} &
\img{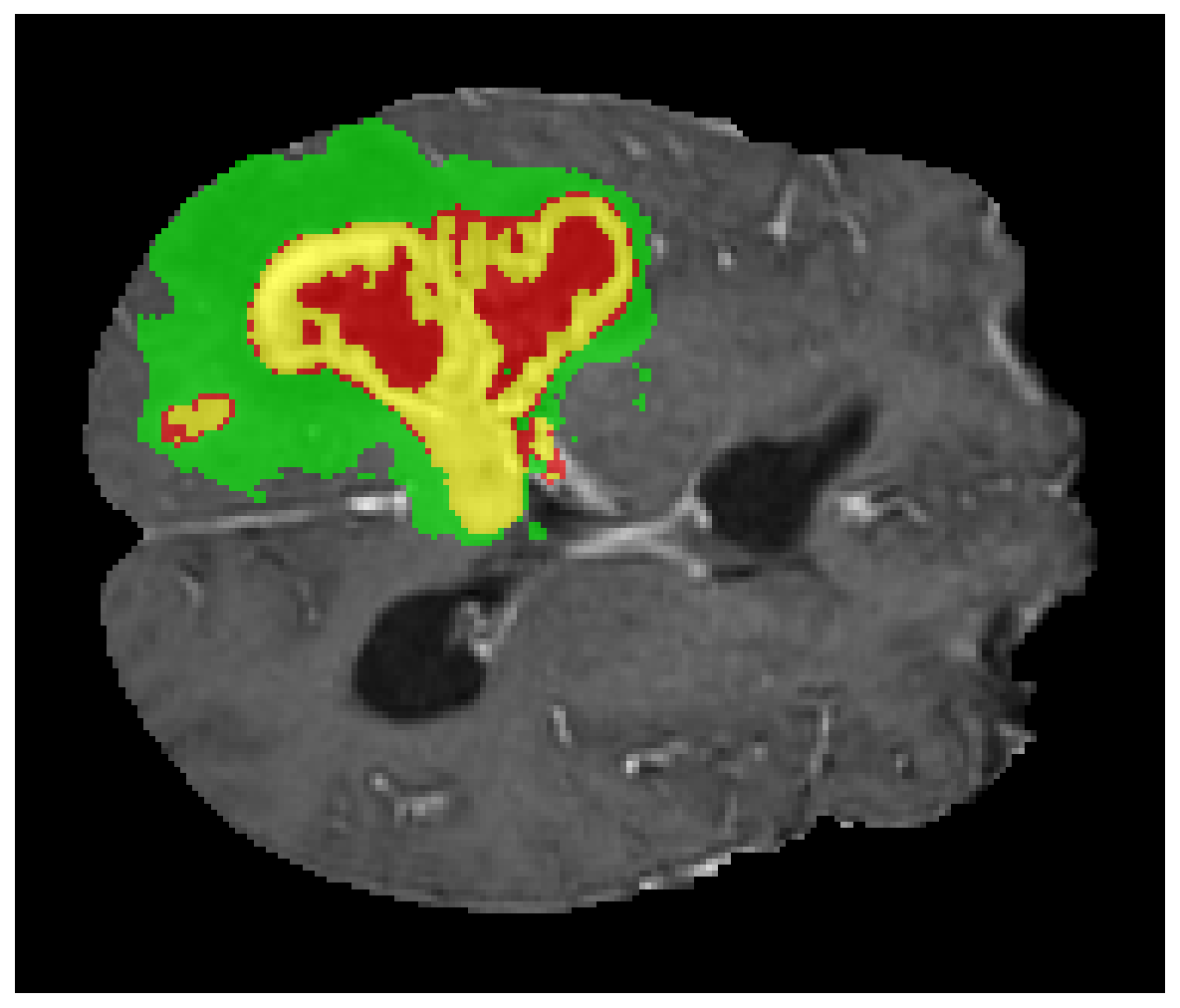}{89.63} &
\img{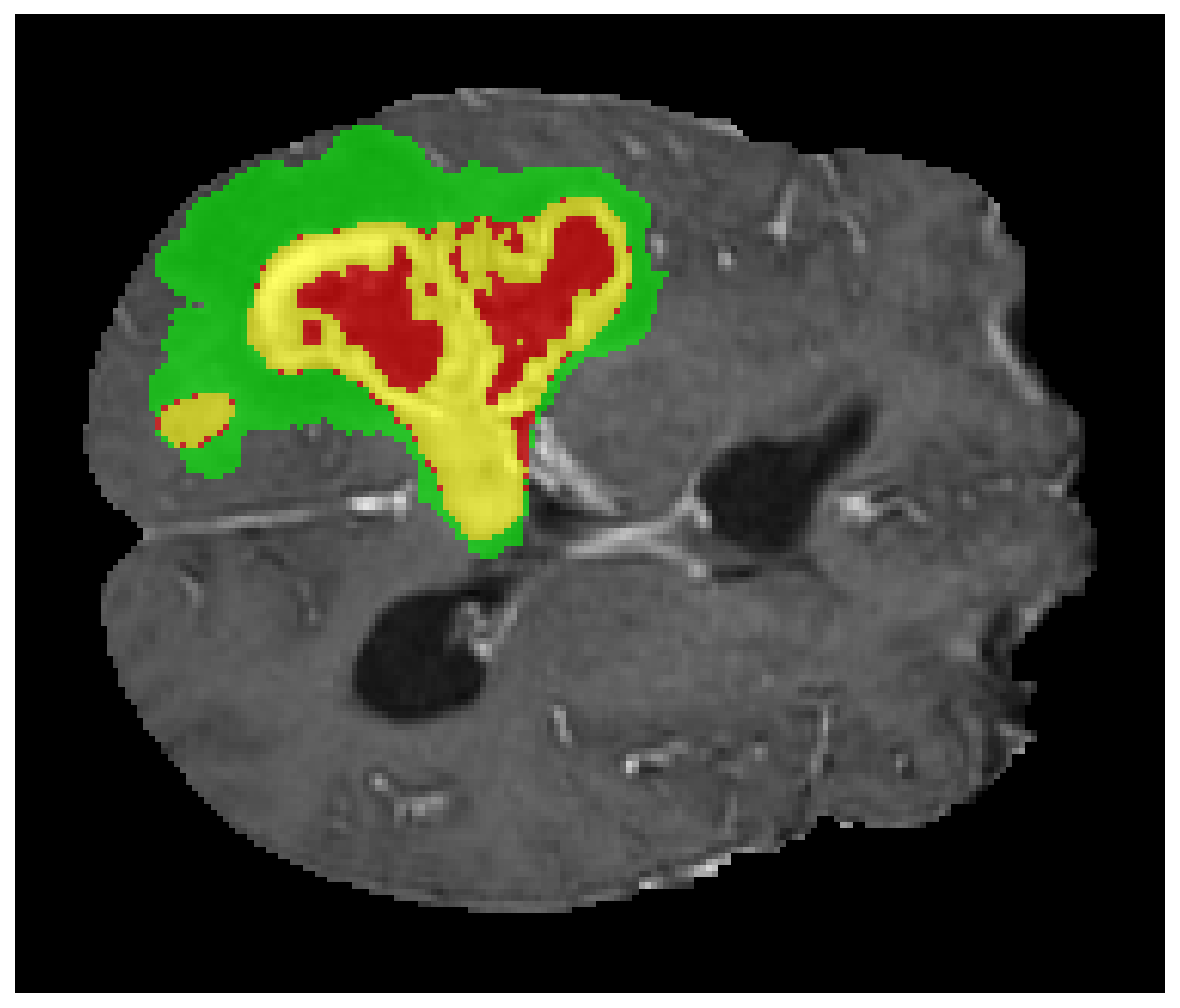}{92.14} \\

\modalityrow{T2 only} &
\img{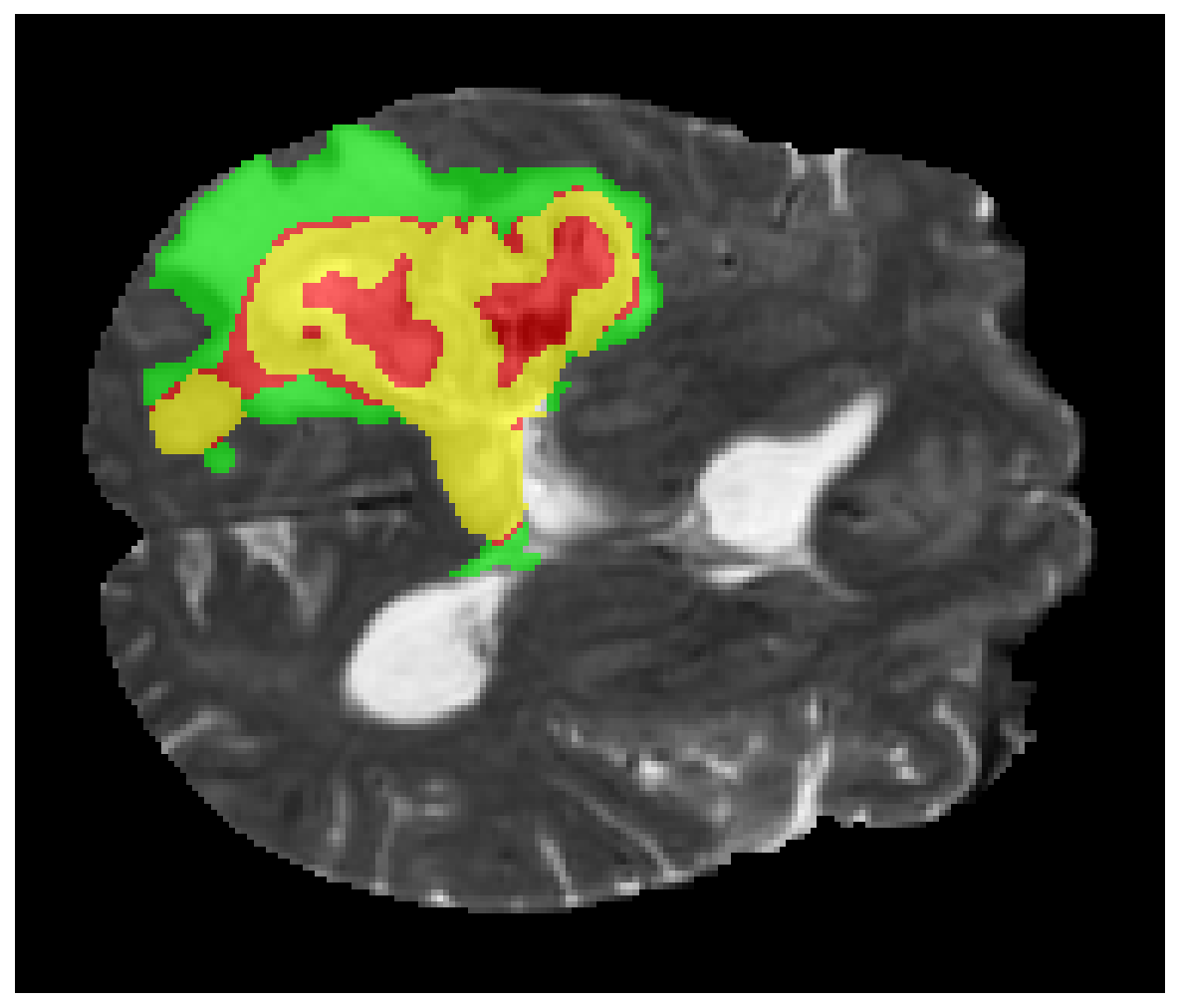}{} &
\img{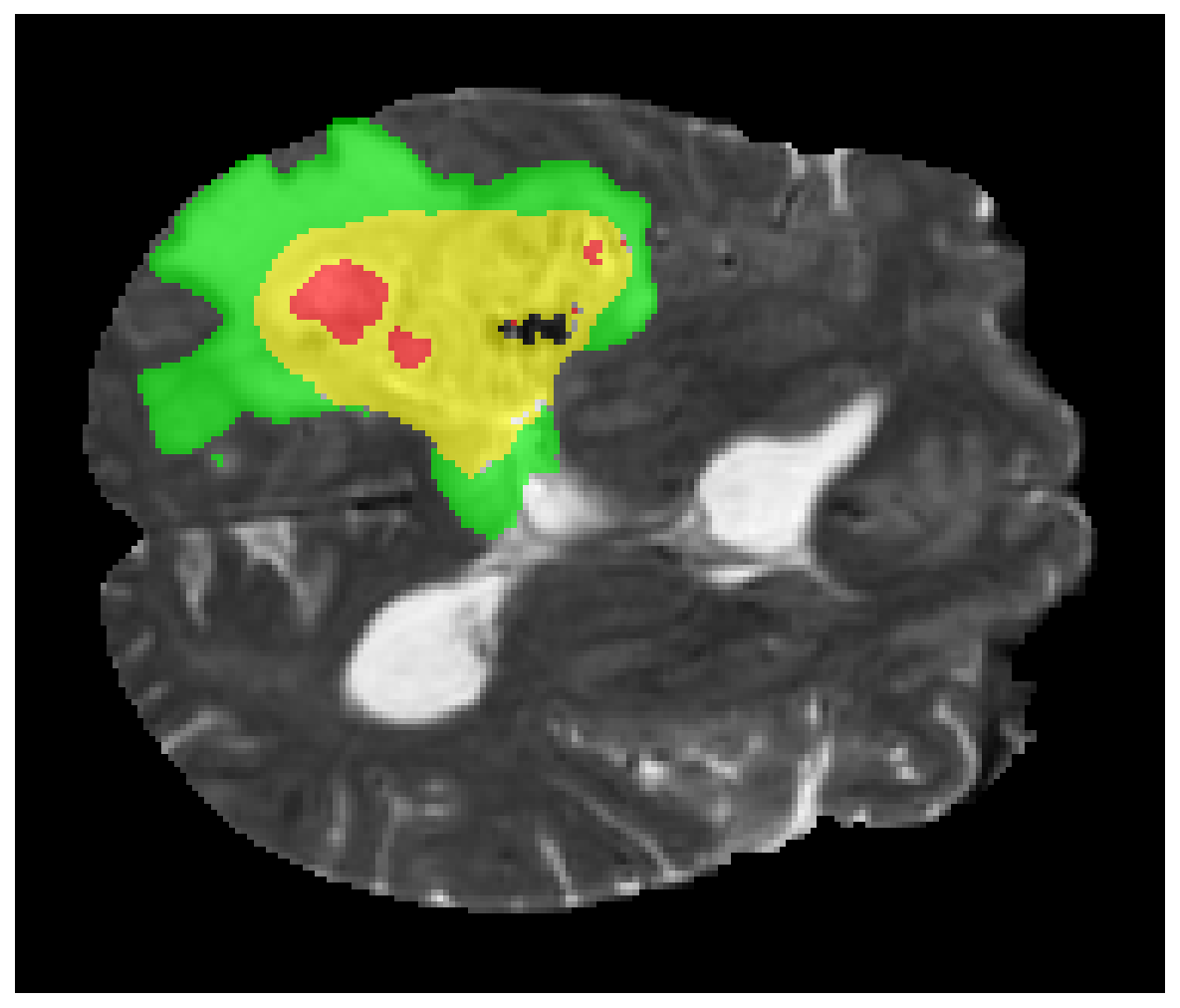}{82.73} &
\img{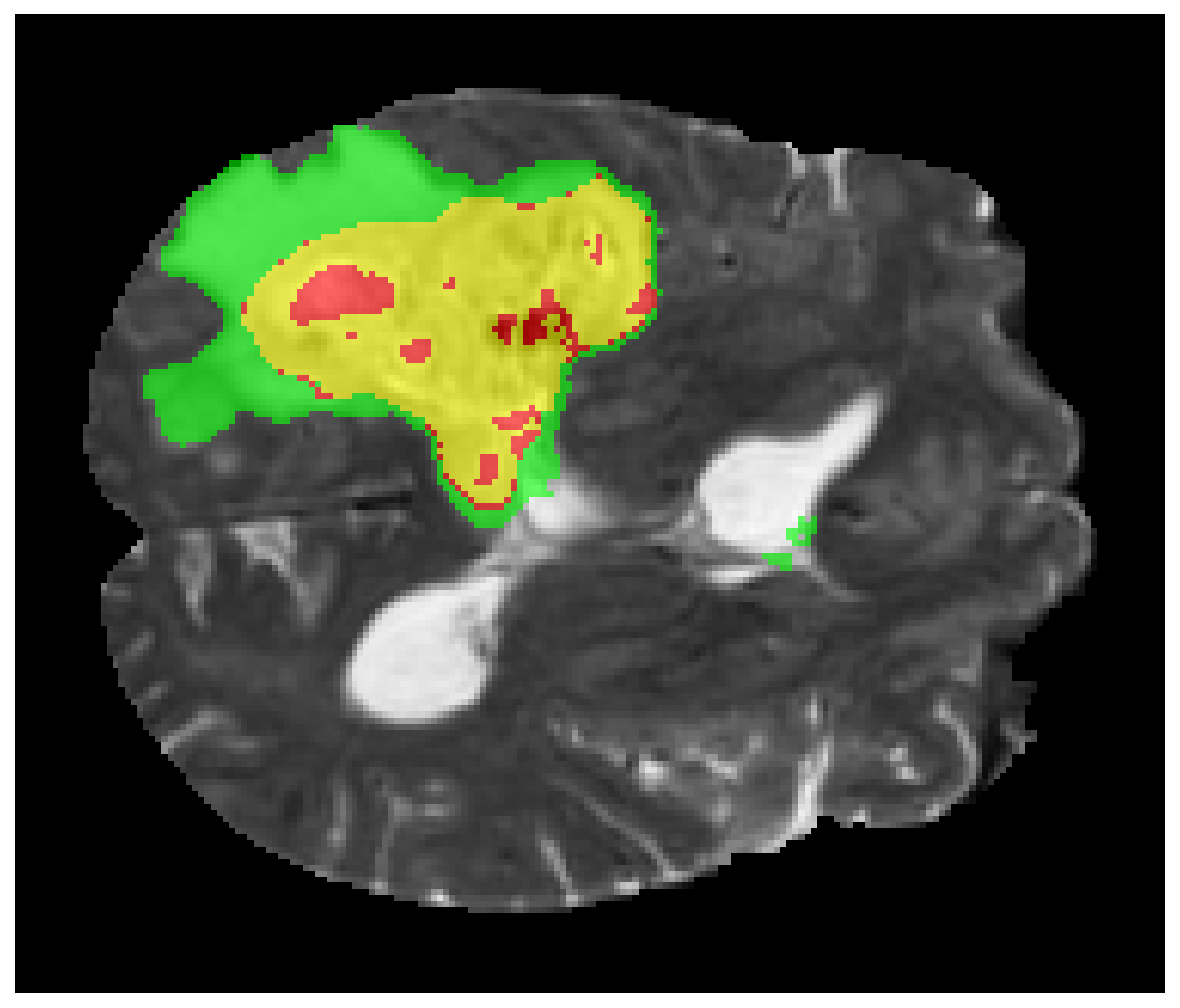}{84.09} &
\img{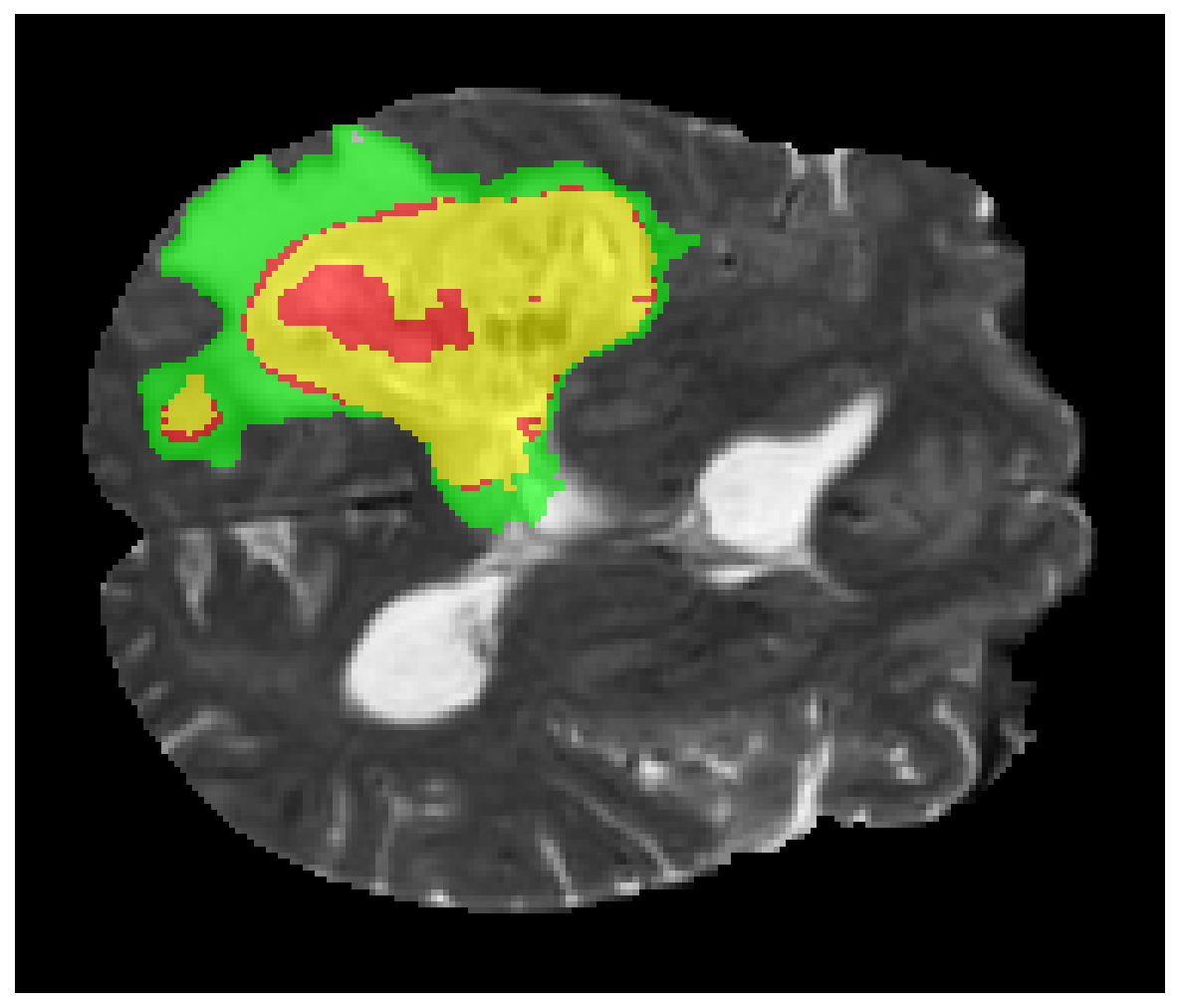}{85.15} &
\img{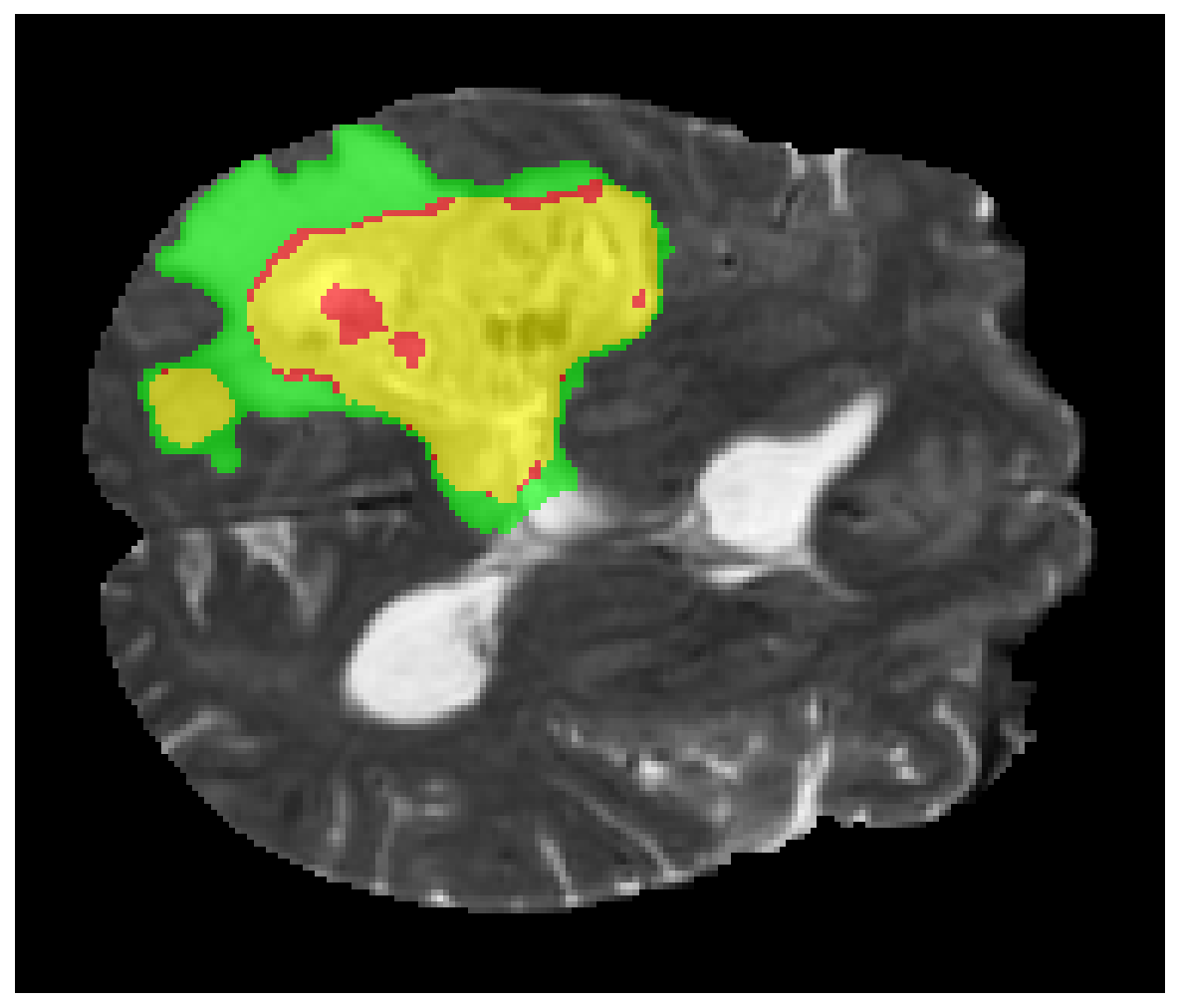}{86.04} &
\img{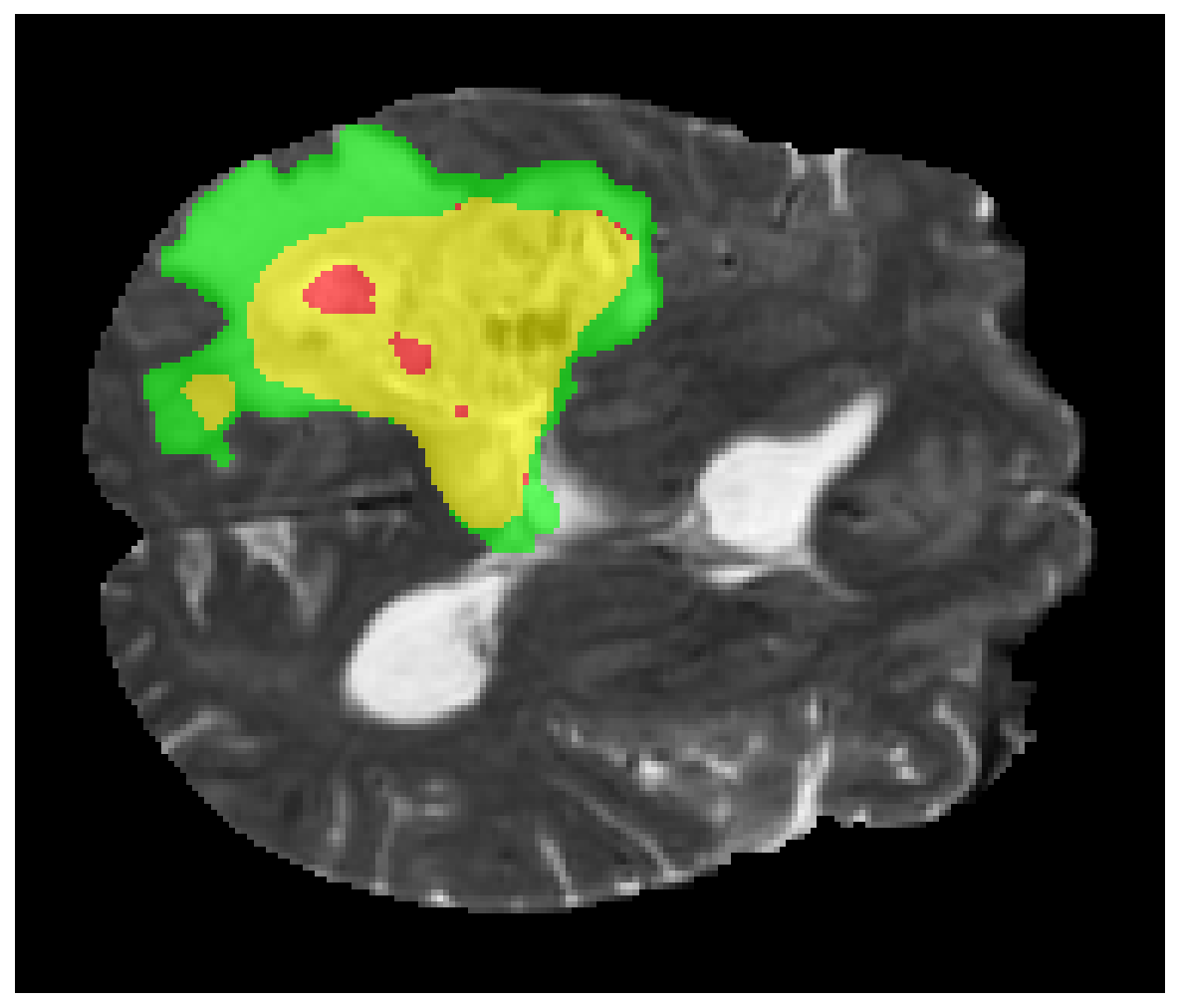}{87.01} \\

\modalityrow{All} &
\img{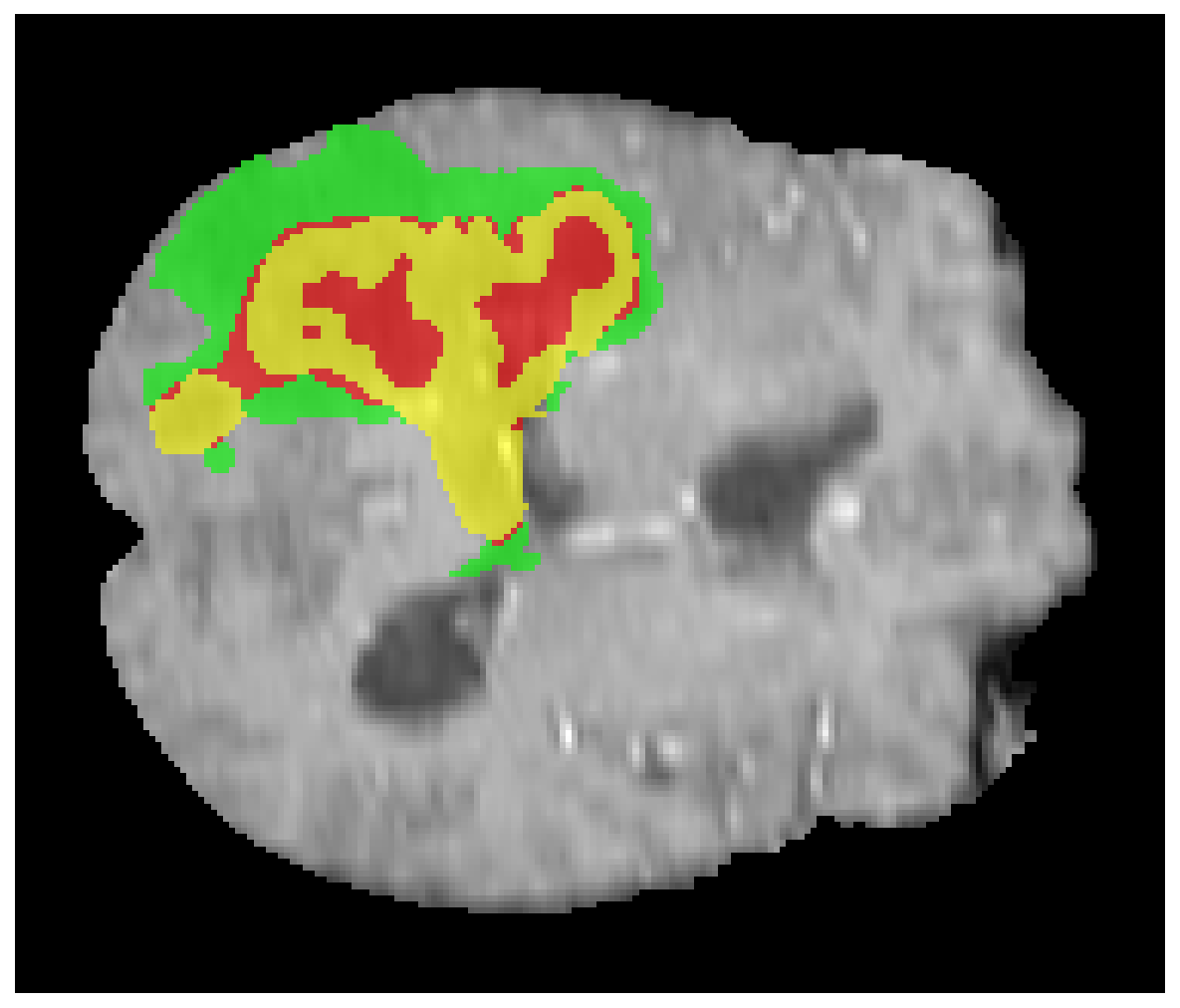}{} &
\img{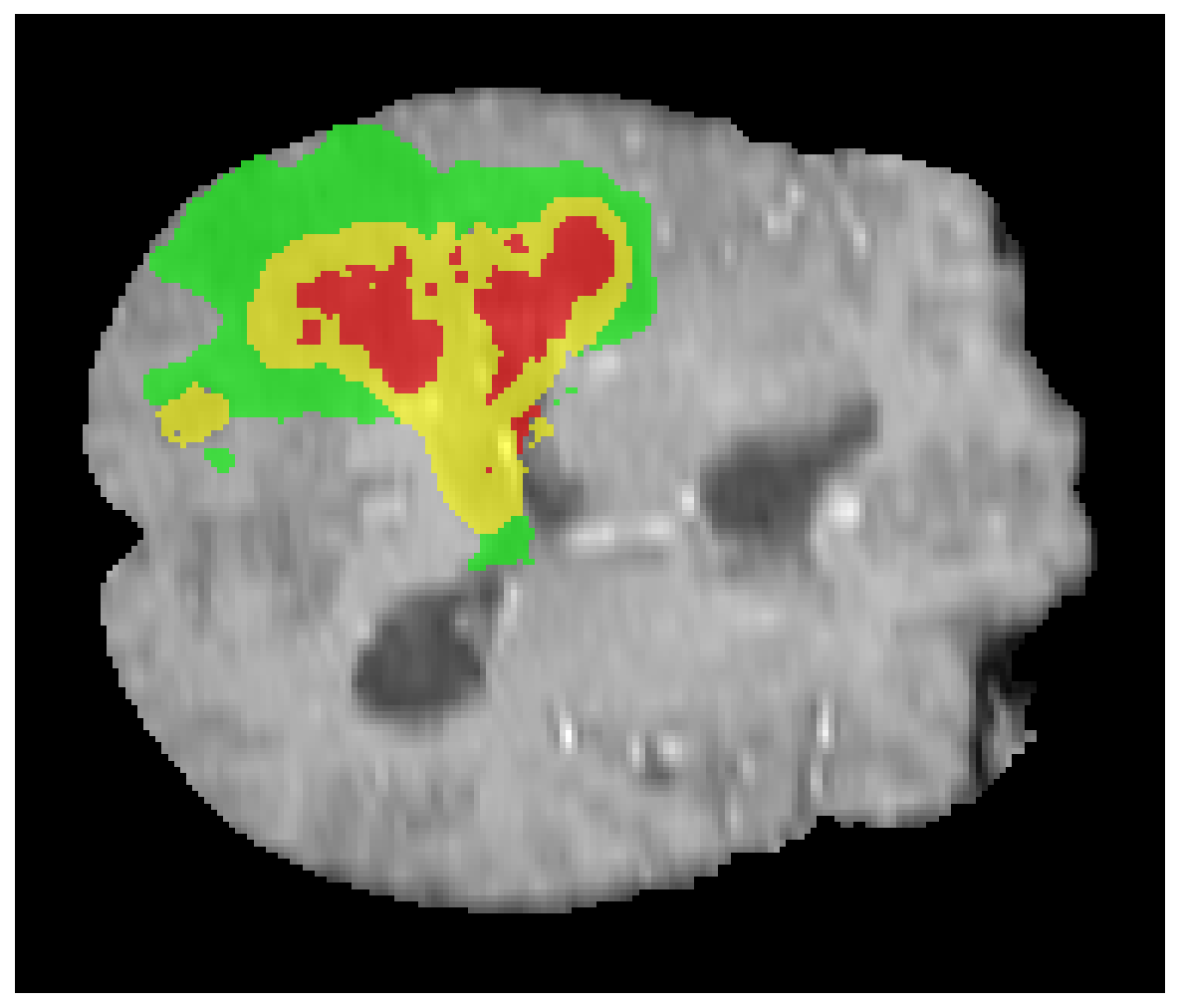}{92.16} &
\img{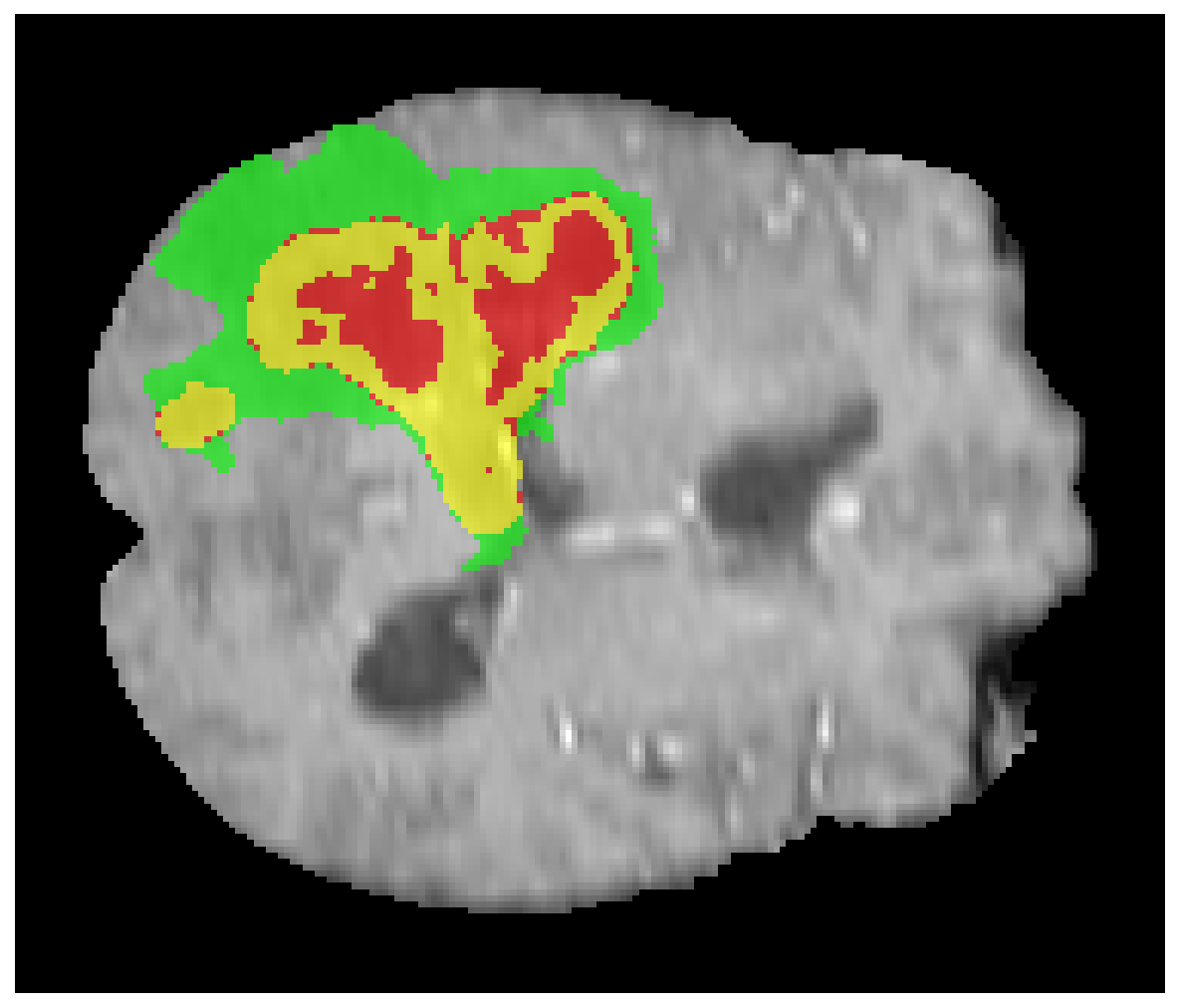}{93.15} &
\img{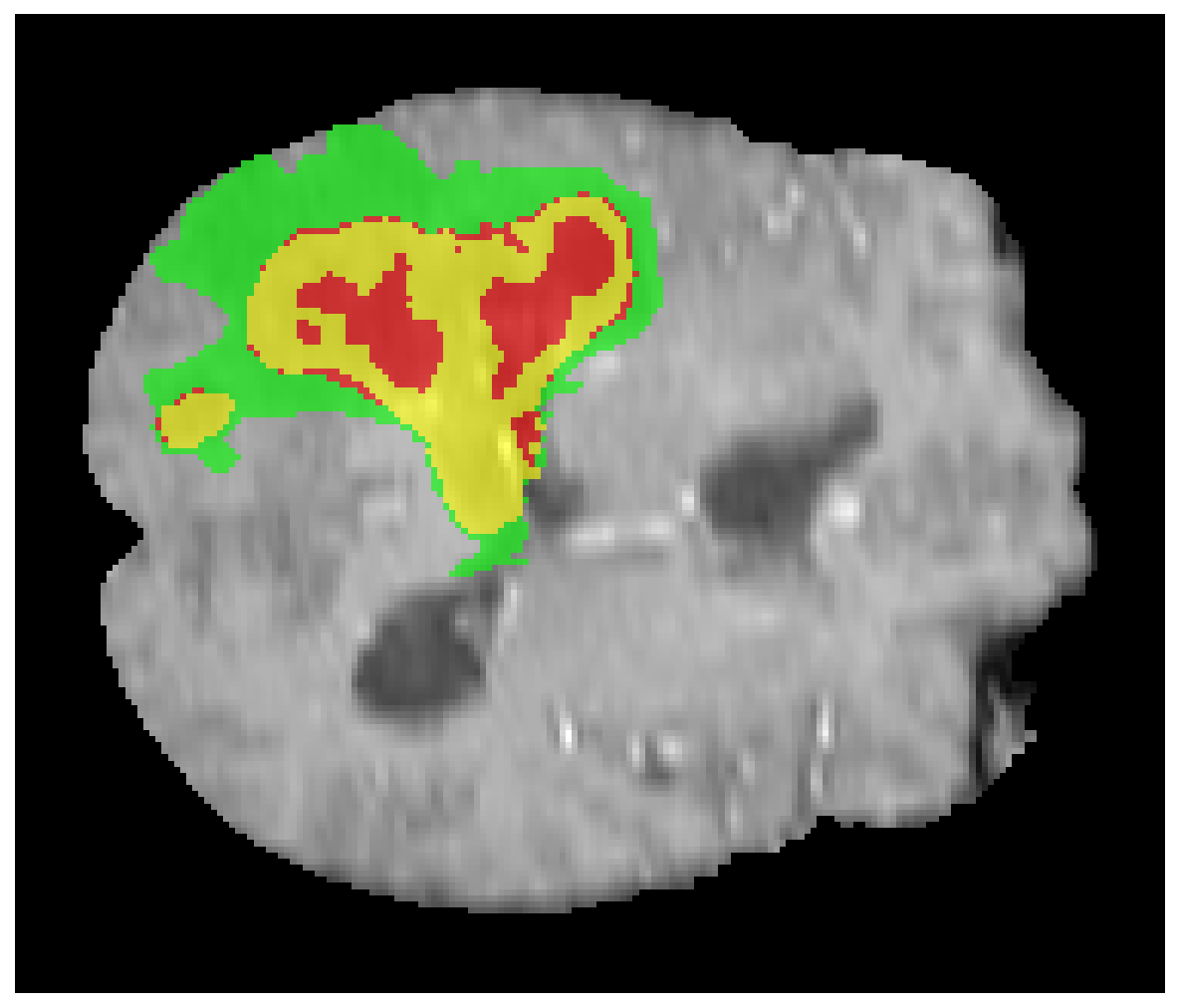}{93.50} &
\img{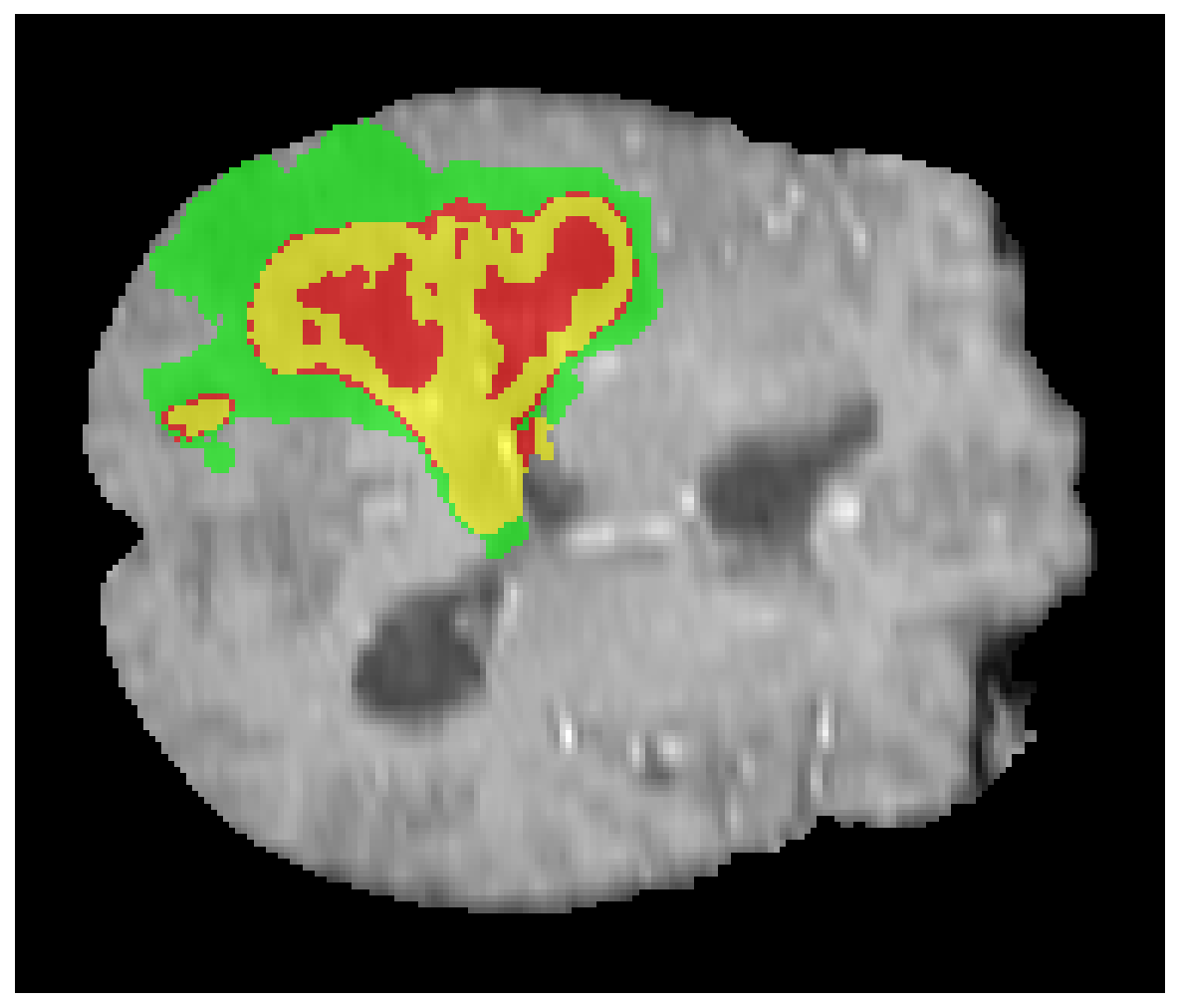}{92.23} &
\img{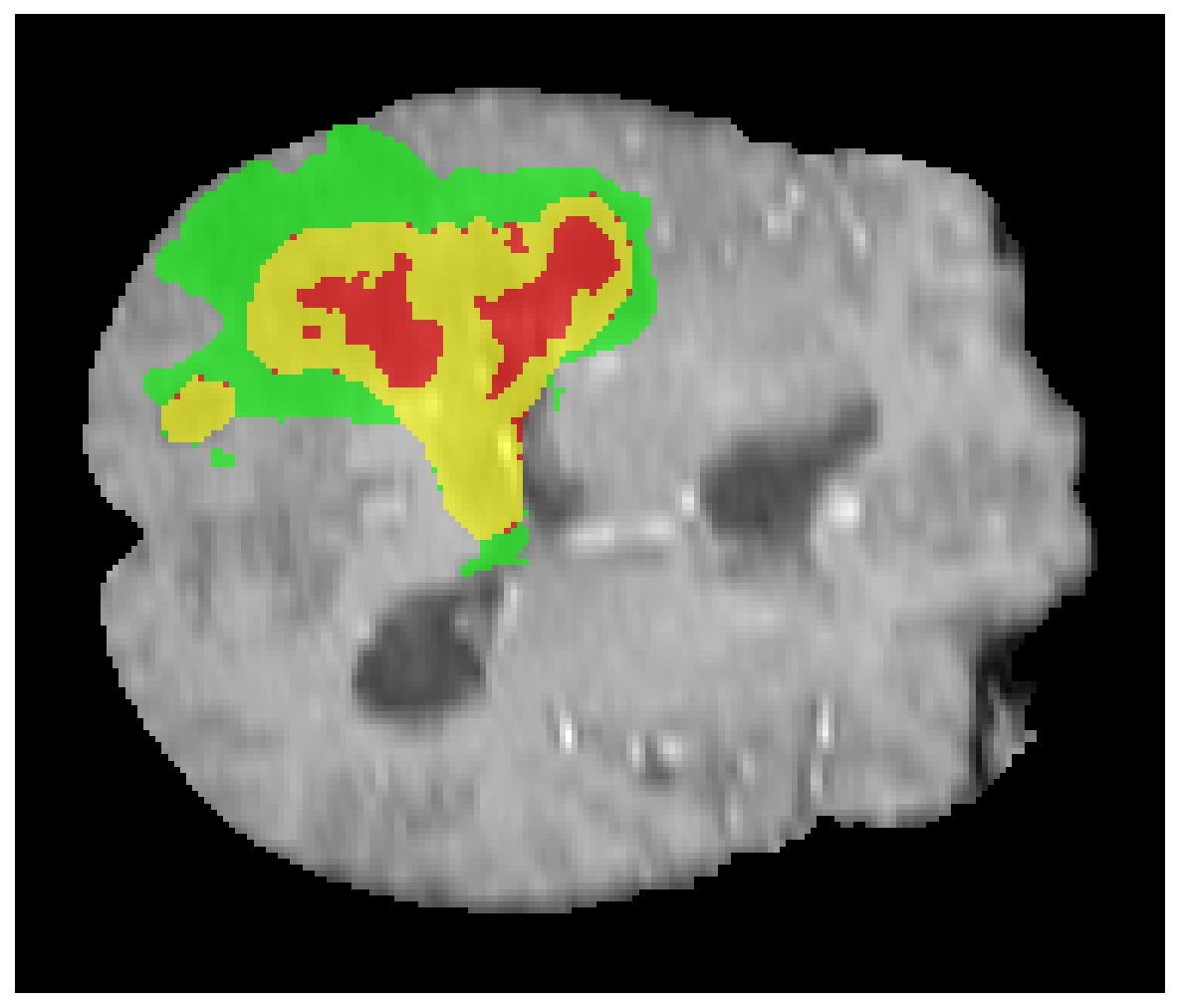}{94.39} \\

\end{tabular}}

\begin{tikzpicture}
    \fill[yellow] (0,0) rectangle (0.4,0.4);
    \node[right] at (0.4,0.2) {Enhancing Tumor};

    \fill[red] (4,0) rectangle (4.4,0.4);
    \node[right] at (4.4,0.2) {Necrotic/Non-Enhancing Tumor};

    \fill[green] (10,0) rectangle (10.4,0.4);
    \node[right] at (10.4,0.2) {Edema};
\end{tikzpicture}

\caption{Qualitative comparison on BraTS 2018. Ground truth (GT) and predictions from different methods under single-modality and all-modality settings are shown. Yellow indicates enhancing tumor, red indicates the necrotic and non-enhancing tumor regions, and green indicates edema. Dice scores are reported for each prediction.}
\label{fig:qualitative-brats}
\end{figure*}

\paragraph{Single-model approaches with modality dropout.}
A widely used strategy replaces missing modalities with zeros, often combined with sparsified training that randomly drops modalities so that one network can process arbitrary subsets at test time~\citep{pemberton_multi-class_2023}. While simple, this forces a single fixed parameter set to serve all $2^N-1$ combinations and inevitably compromises configurations that would benefit from different fusion behavior.

\paragraph{Knowledge transfer, generative, and dedicated models.}
Knowledge distillation~\citep{maheshwari_missing_2023,wang_learnable_2023} trains incomplete-modality students to mimic full-modality teachers, while SMIL~\citep{ma_smil_2021} casts the problem as Bayesian meta-learning. Generative approaches~\citep{chartsias_multimodal_2018,dar_image_2019} synthesize missing modalities but risk hallucinating clinically dangerous features and require multi-stage training. Training a separate dedicated model per non-empty subset maximizes specialization but scales exponentially in storage and compute~\citep{wu2024deepmultimodallearningmissing}; LARGO aims at retaining this specialization while replacing explicit storage by a jointly optimized low-rank parameterization.

\begin{revblock}
\paragraph{Recent state of the art.}
LS3M~\citep{zhang_ls3m_2025} reorders modality feature sequences with differentiable permutations and fuses them with state-space blocks; Hierarchical Gradient Alignment (HGA)~\citep{shi_hga_2026} aligns gradients across configurations to handle realistic, imbalanced incomplete-modality distributions; and UniME~\citep{song_unime_2026} pretrains a ViT uni-encoder with masked image modelling and adds modality-specific CNN encoders before fusion. All three act on features or on the optimizer and keep the model representation, whereas LARGO changes the parameterization of the model family and could be combined with them. We do not claim empirical superiority over them: UniME is evaluated on BraTS 2023/2024 and HGA uses a different, realistic-incompleteness protocol.\end{revblock}

\begin{figure*}[!t]
\centering

\resizebox{0.7\linewidth}{!}{%
\newcommand{\modelcol}[1]{\textbf{#1}}
\newcommand{\modalityrow}[1]{
\begin{tikzpicture}
\path (0,0)--(0,-2.55) node[midway,rotate=90] {\textbf{#1}};
\end{tikzpicture}
}

\newcommand{\img}[2]{%
\begin{tikzpicture}
    \node[inner sep=0pt] at (0,0) {\includegraphics[angle=-90,width=0.14\linewidth,trim={80 100 80 100}, clip]{#1}};
    \node[anchor=south east, font=\bfseries, text=black,fill=white,fill opacity=0.7,text opacity=1] at (0.5,-1.15) {#2};
\end{tikzpicture}
}

\begin{tabular}{c c c c c c}
 & \modelcol{mmFormer} & \modelcol{M\textsuperscript{3}AE} & \modelcol{ShaSpec} & \modelcol{SimMLM} & \modelcol{LARGO (Ours)} \\
 \vspace{0.1cm}\\

\modalityrow{DWI only} & 
\img{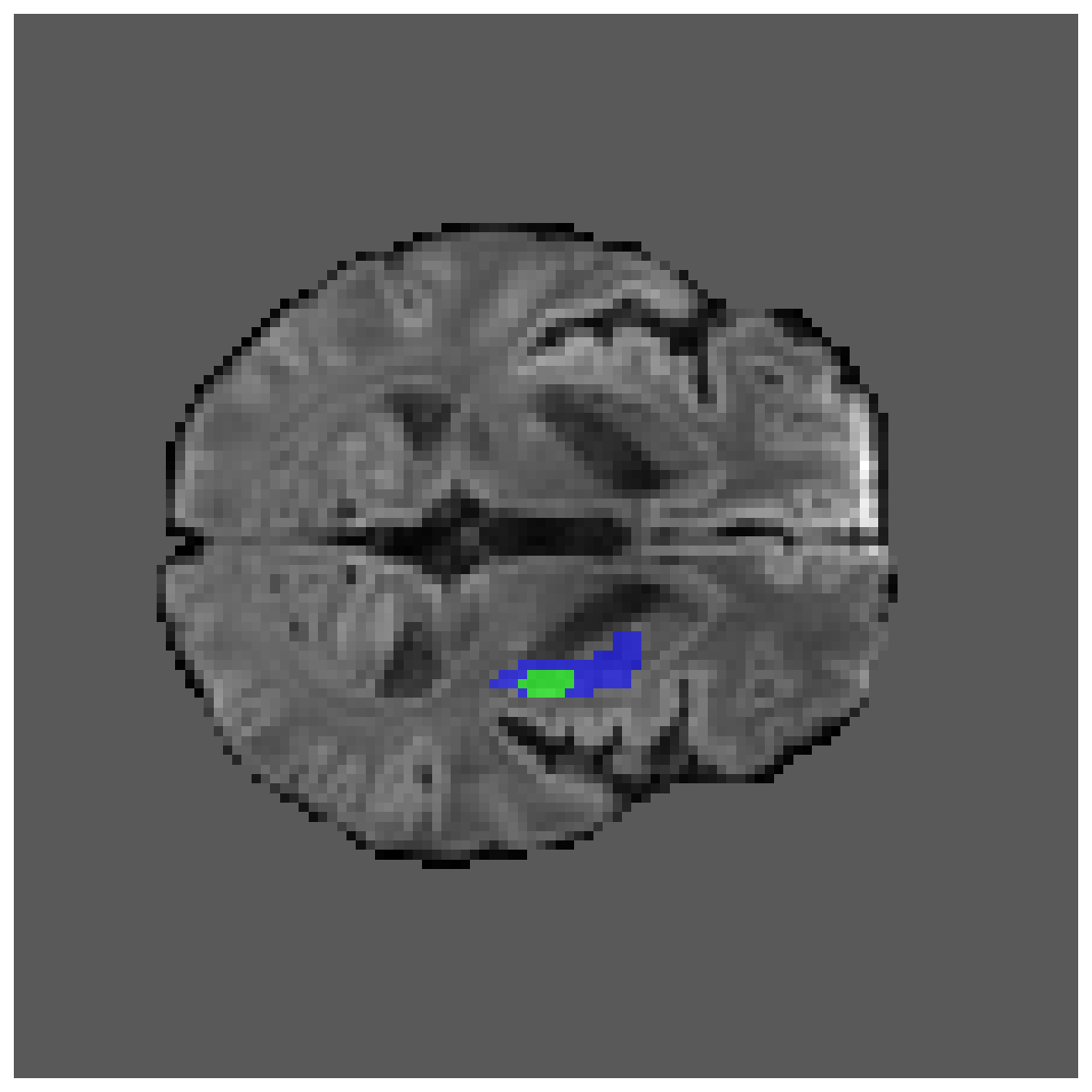}{39.47} &
\img{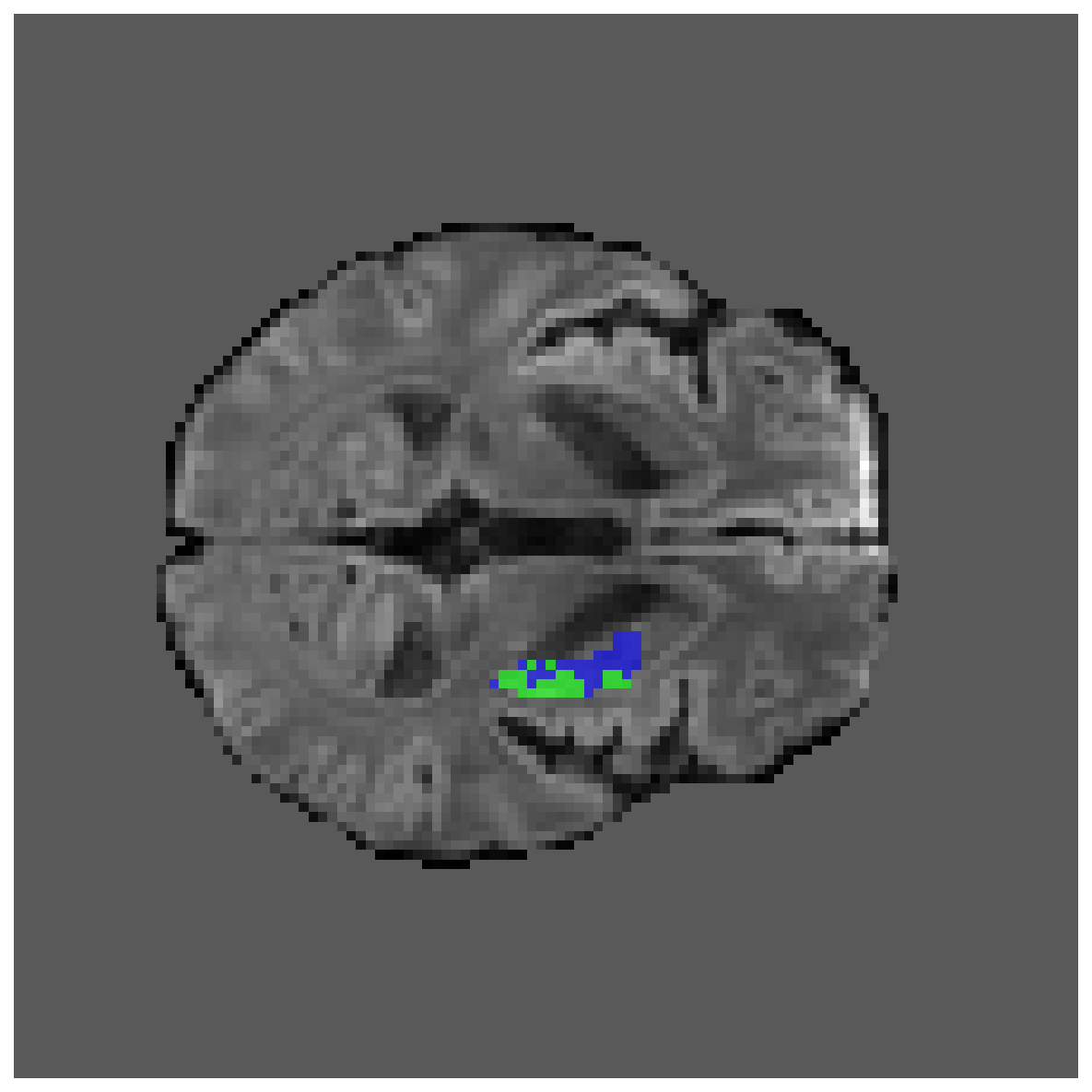}{61.36} &
\img{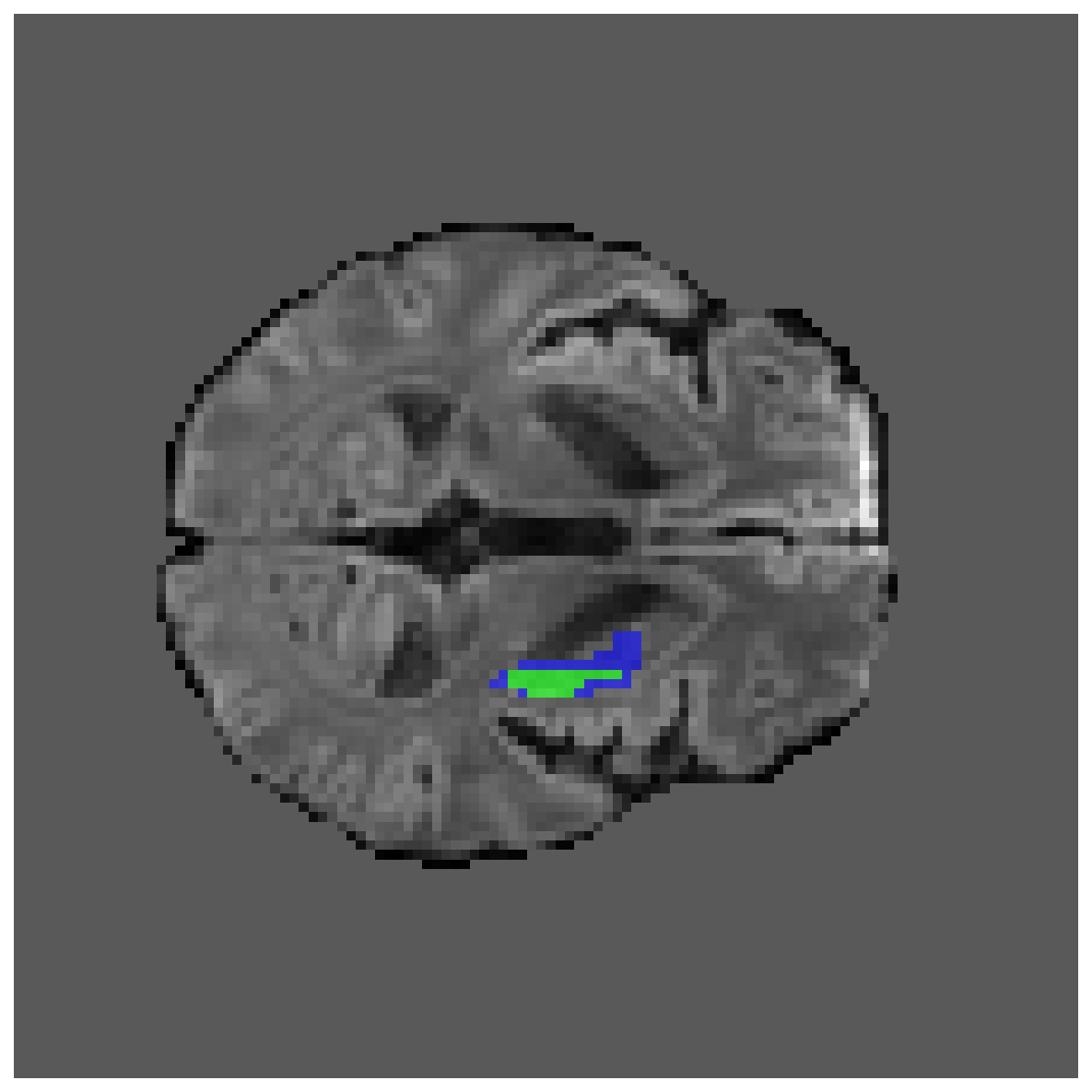}{58.14} &
\img{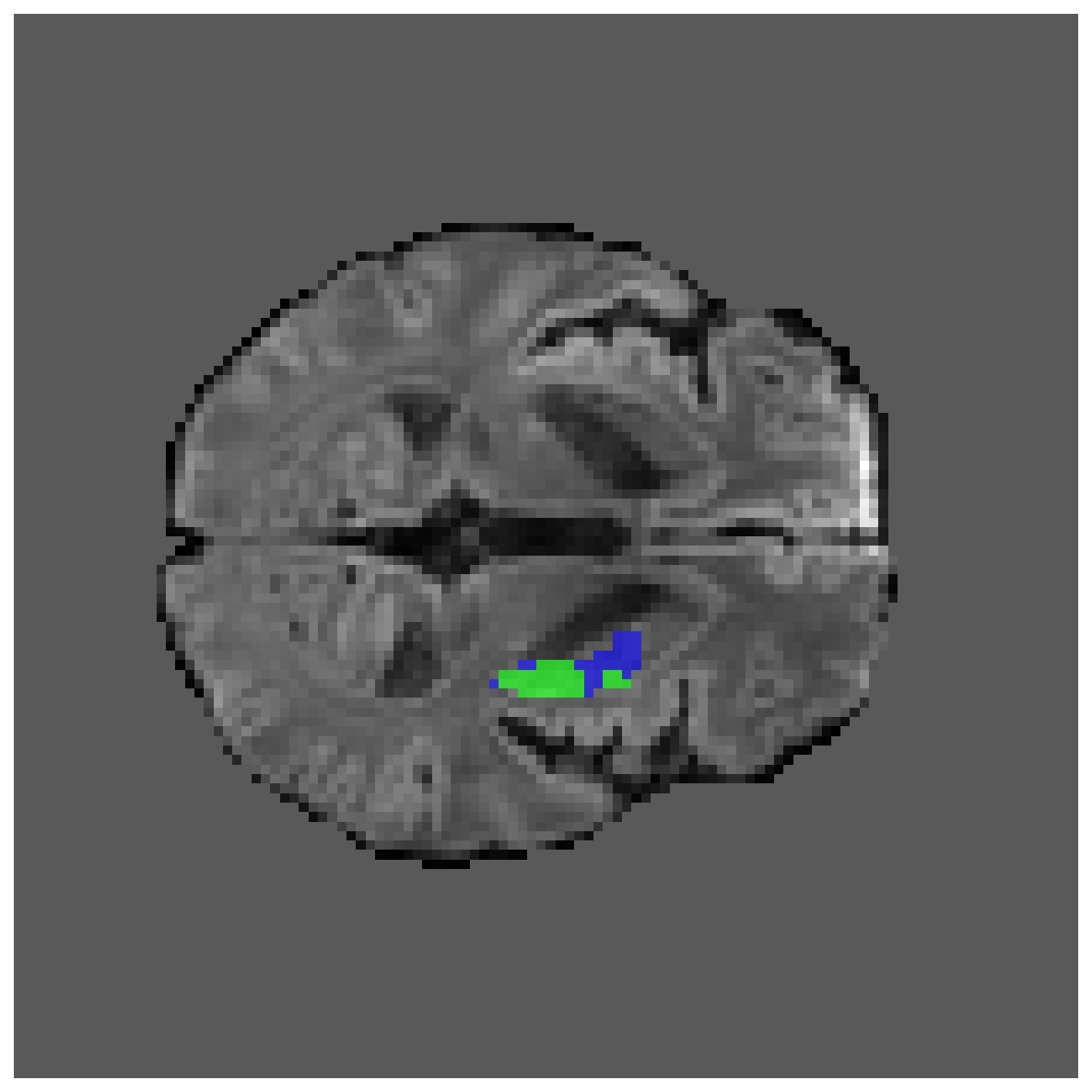}{71.58} &
\img{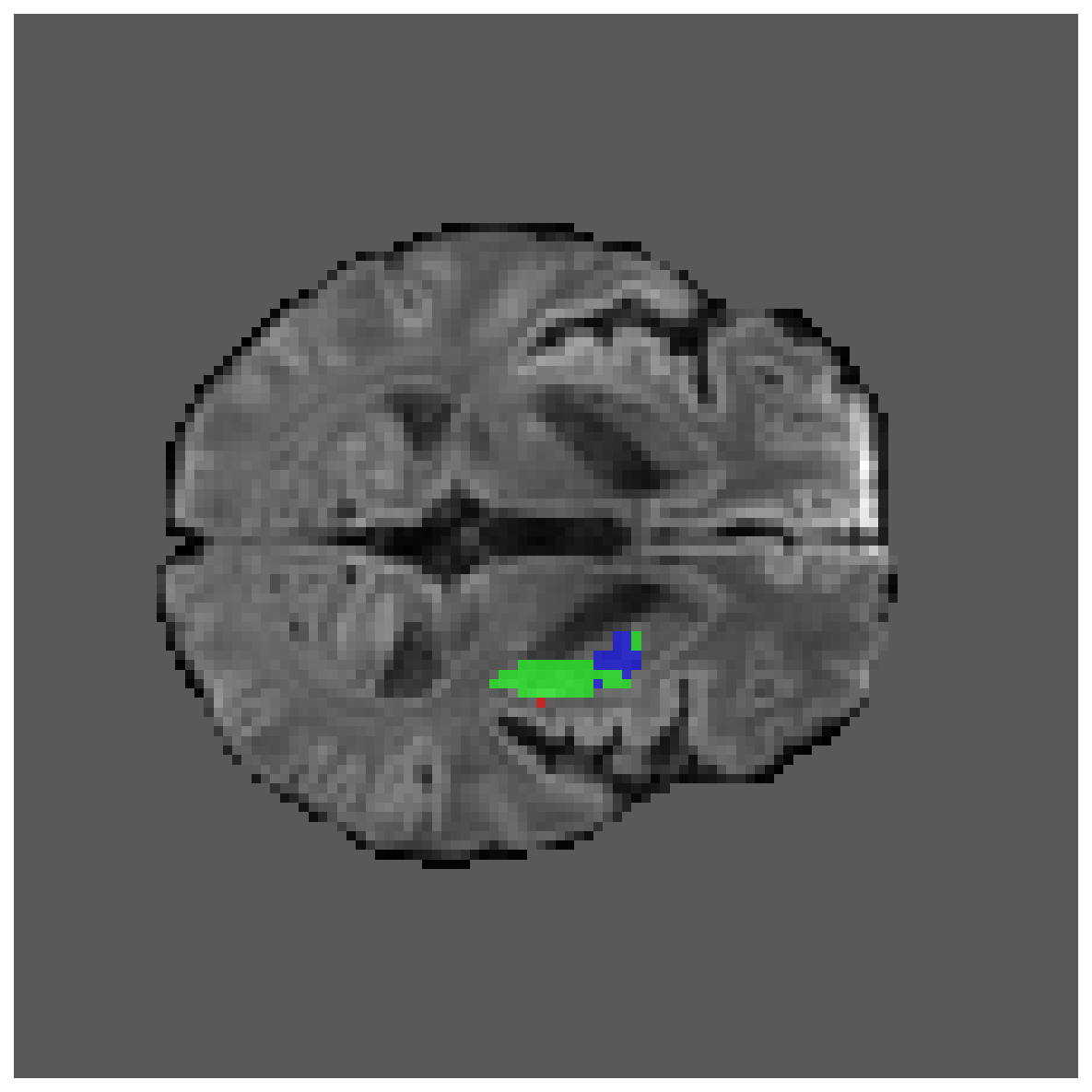}{84.11} \\

\modalityrow{ADC only} & 
\img{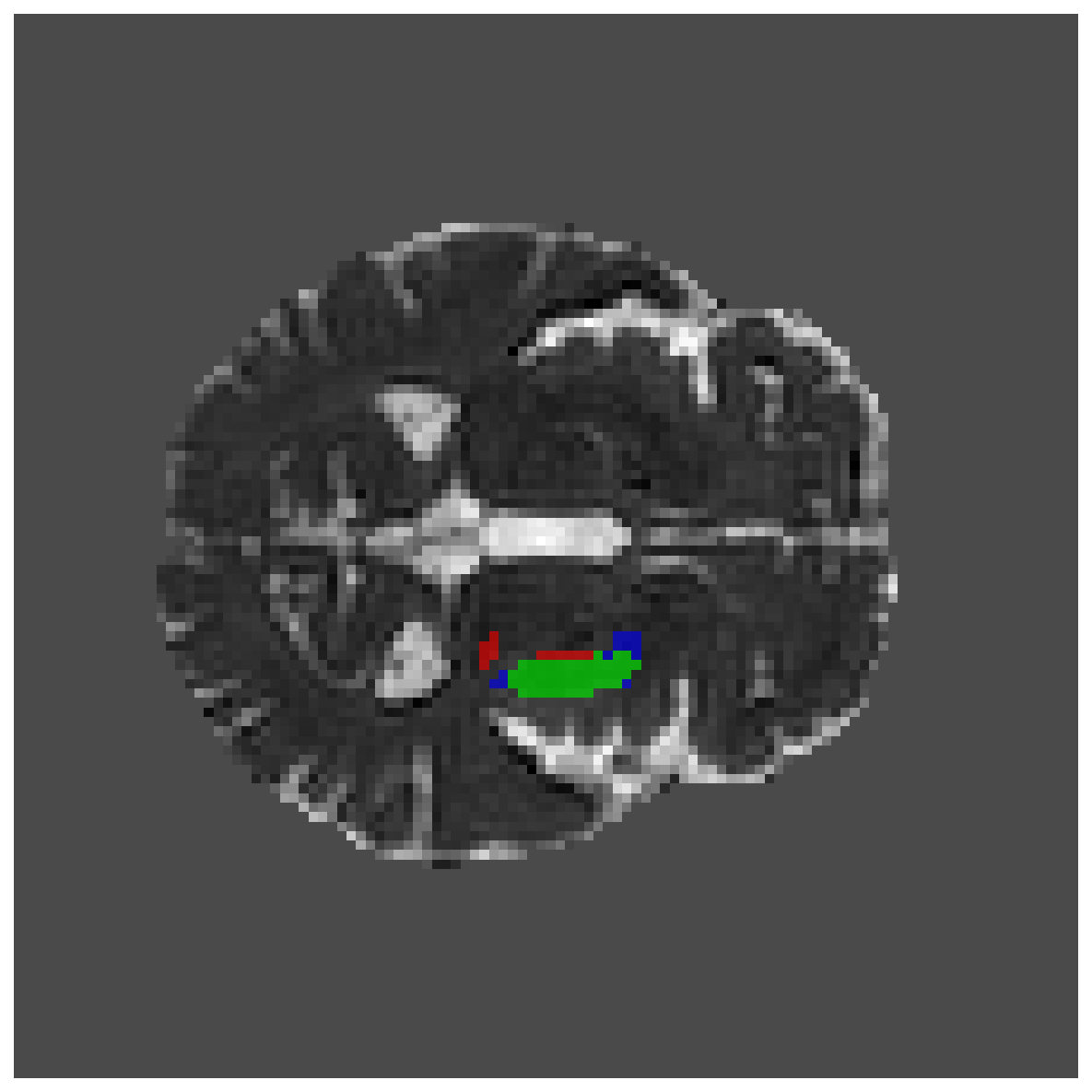}{80.33} &
\img{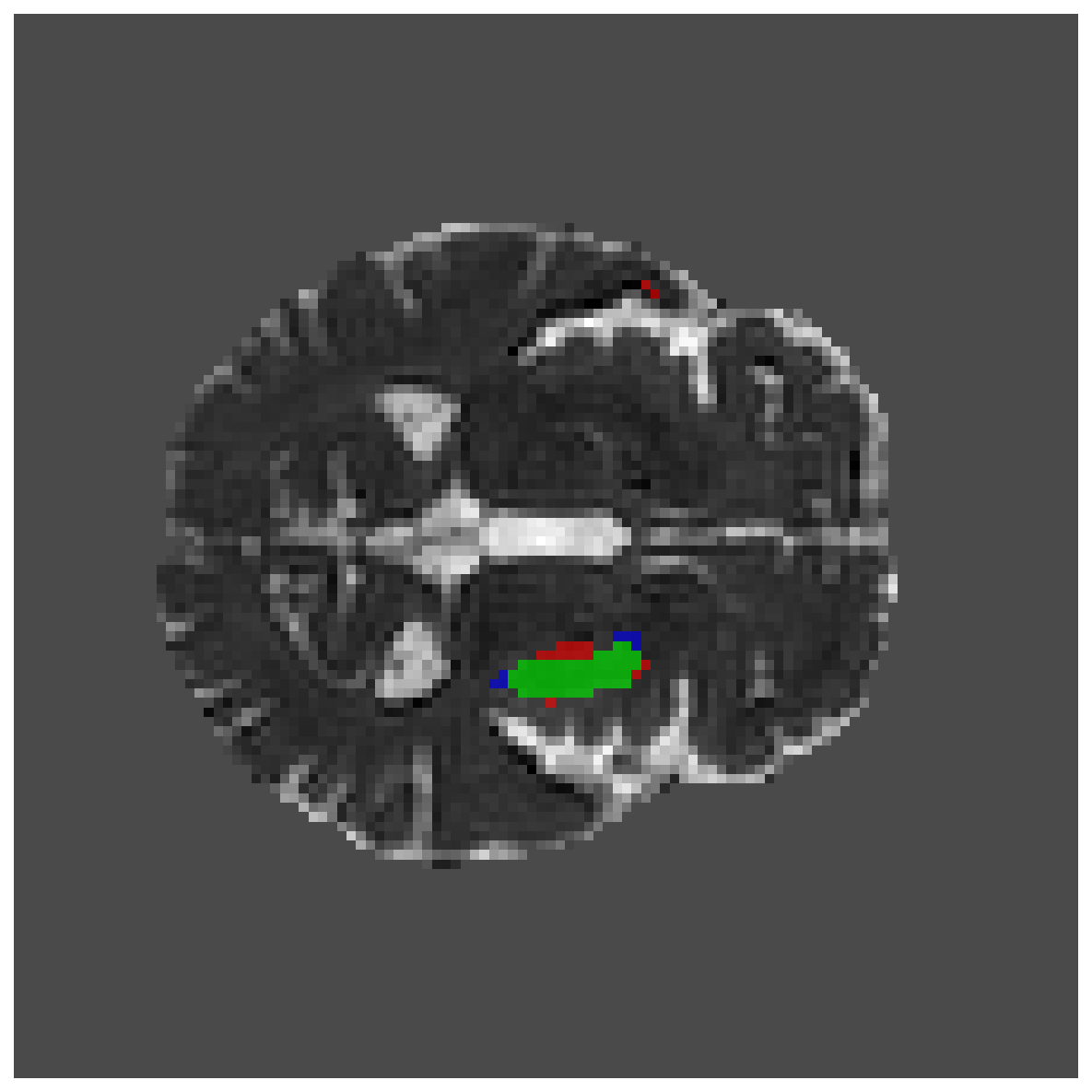}{83.08} &
\img{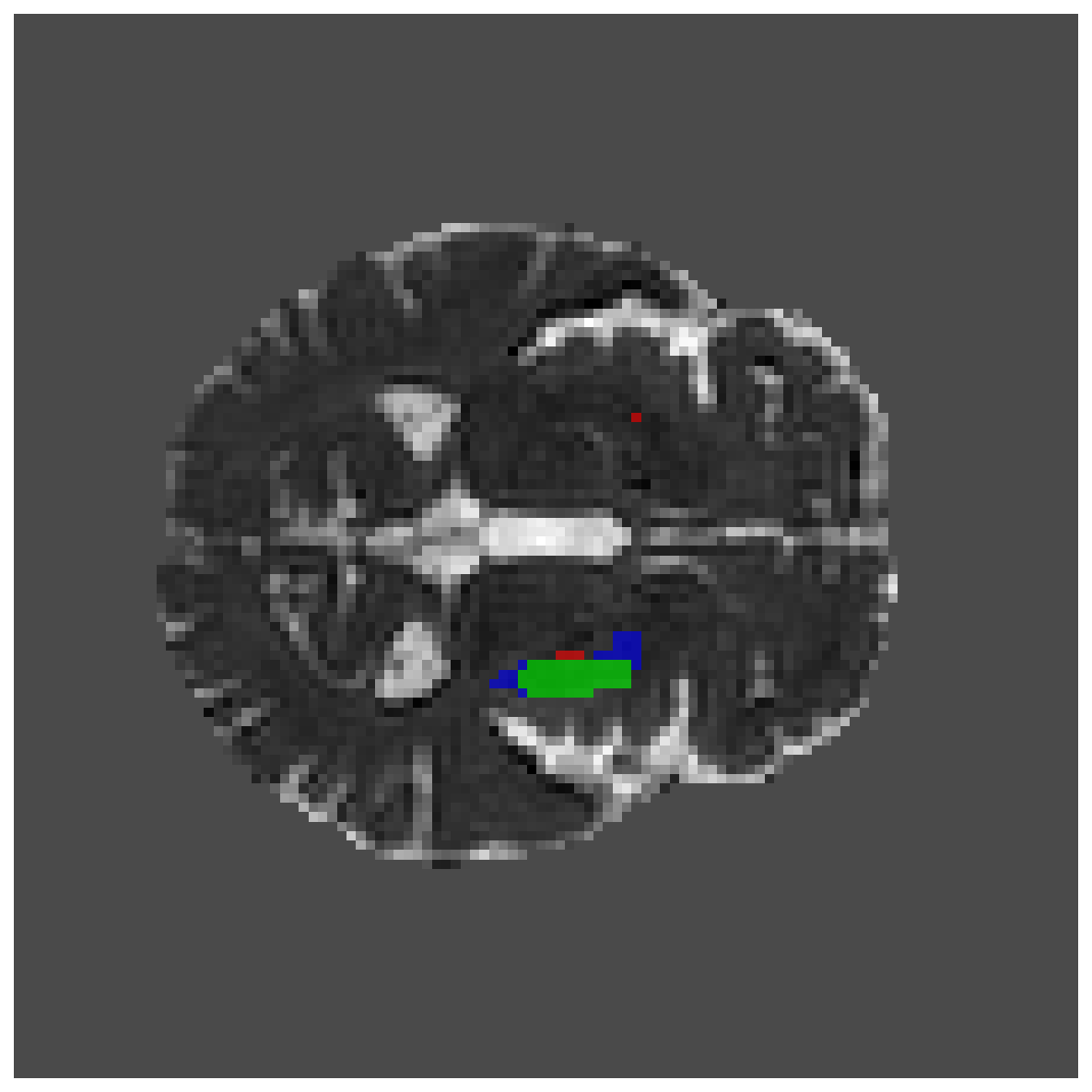}{78.50} &
\img{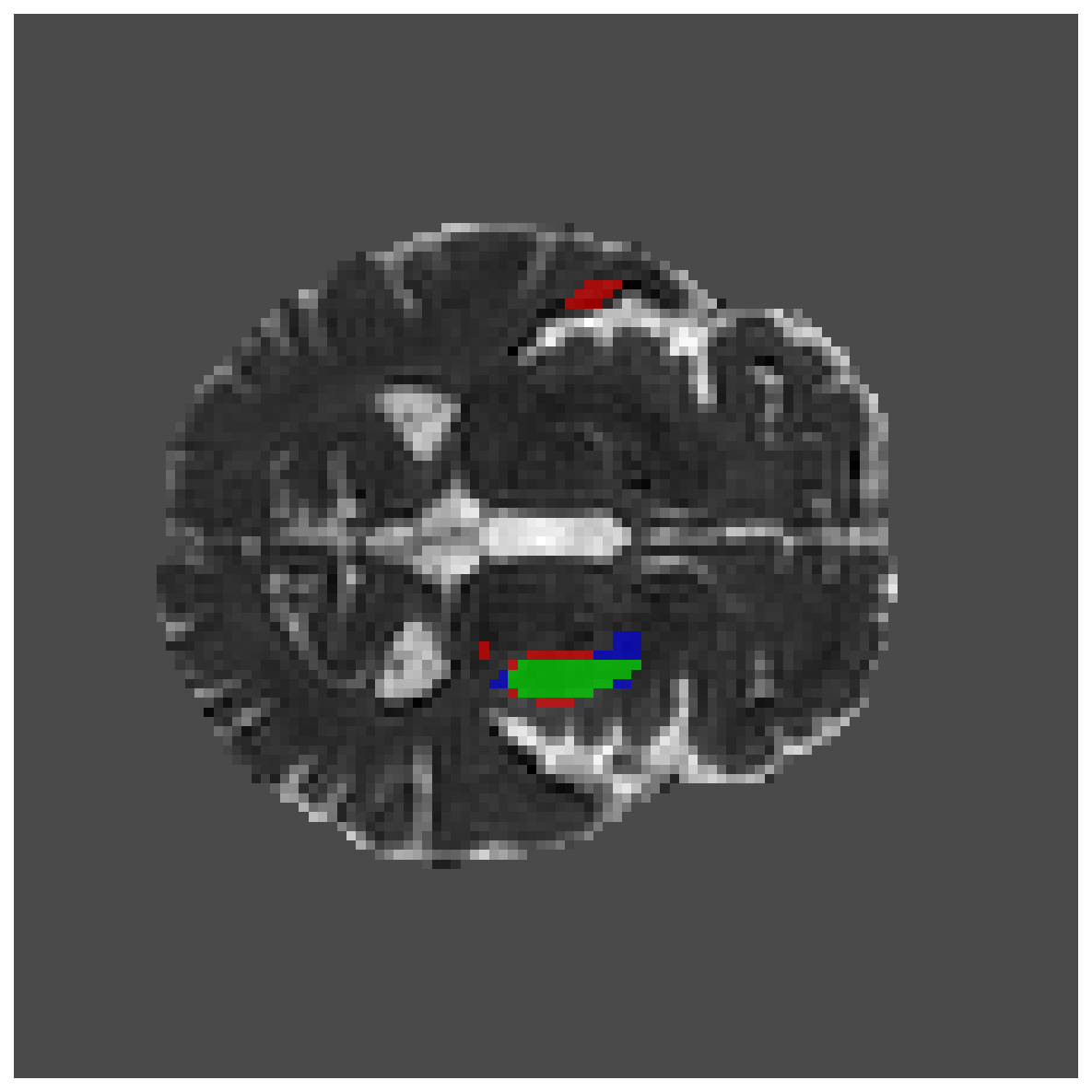}{67.67} &
\img{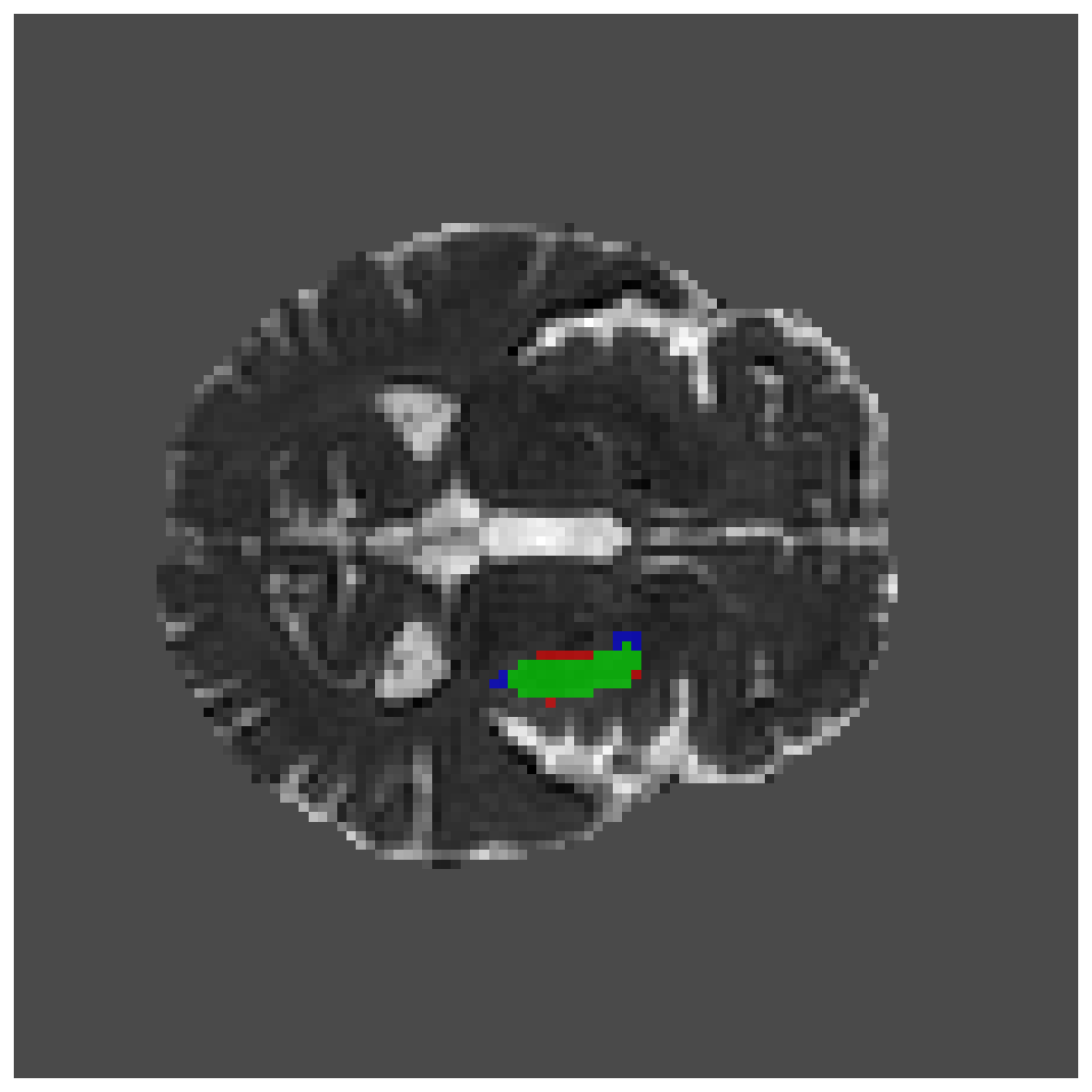}{86.89} \\

\modalityrow{FLAIR only} & 
\img{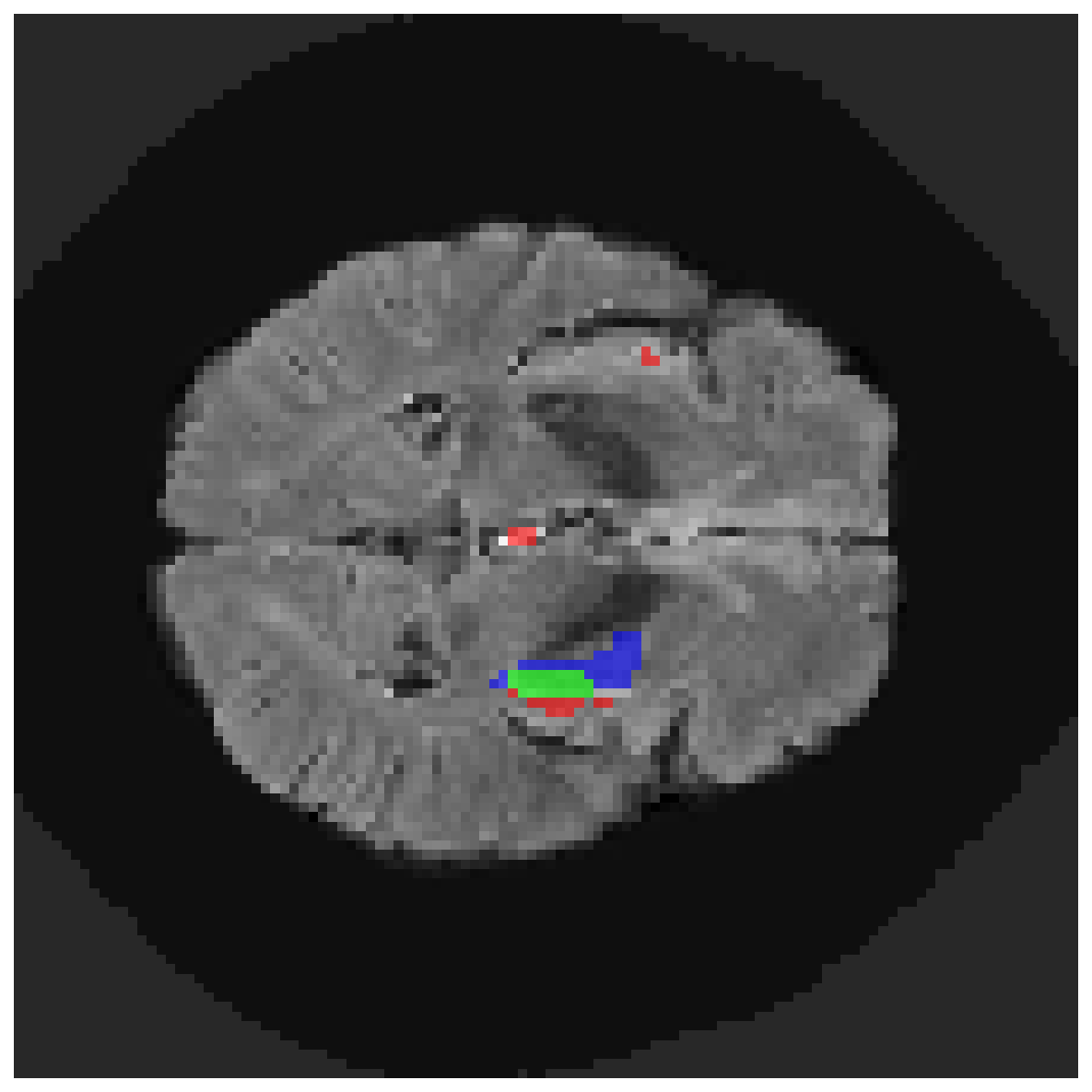}{46.73} &
\img{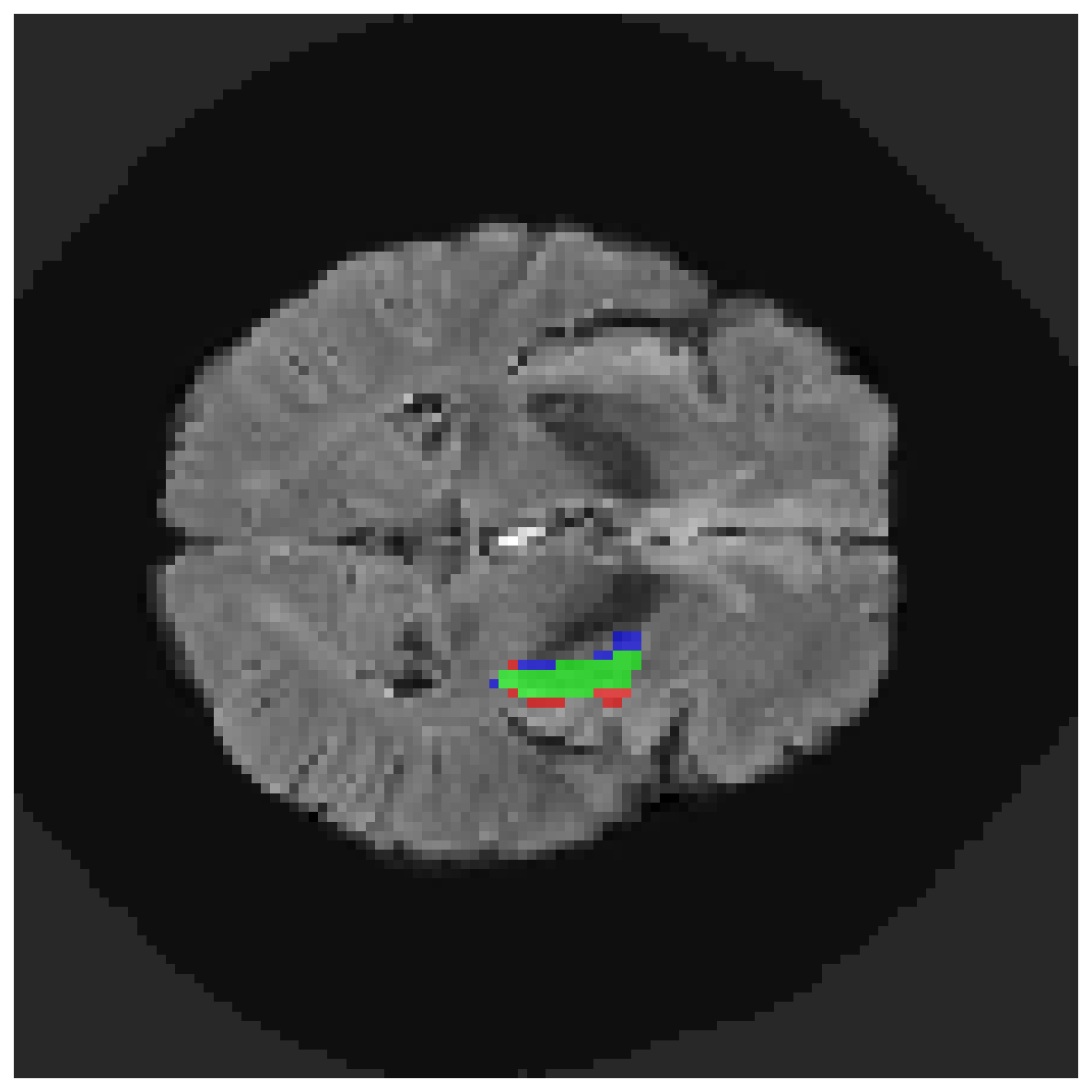}{79.34} &
\img{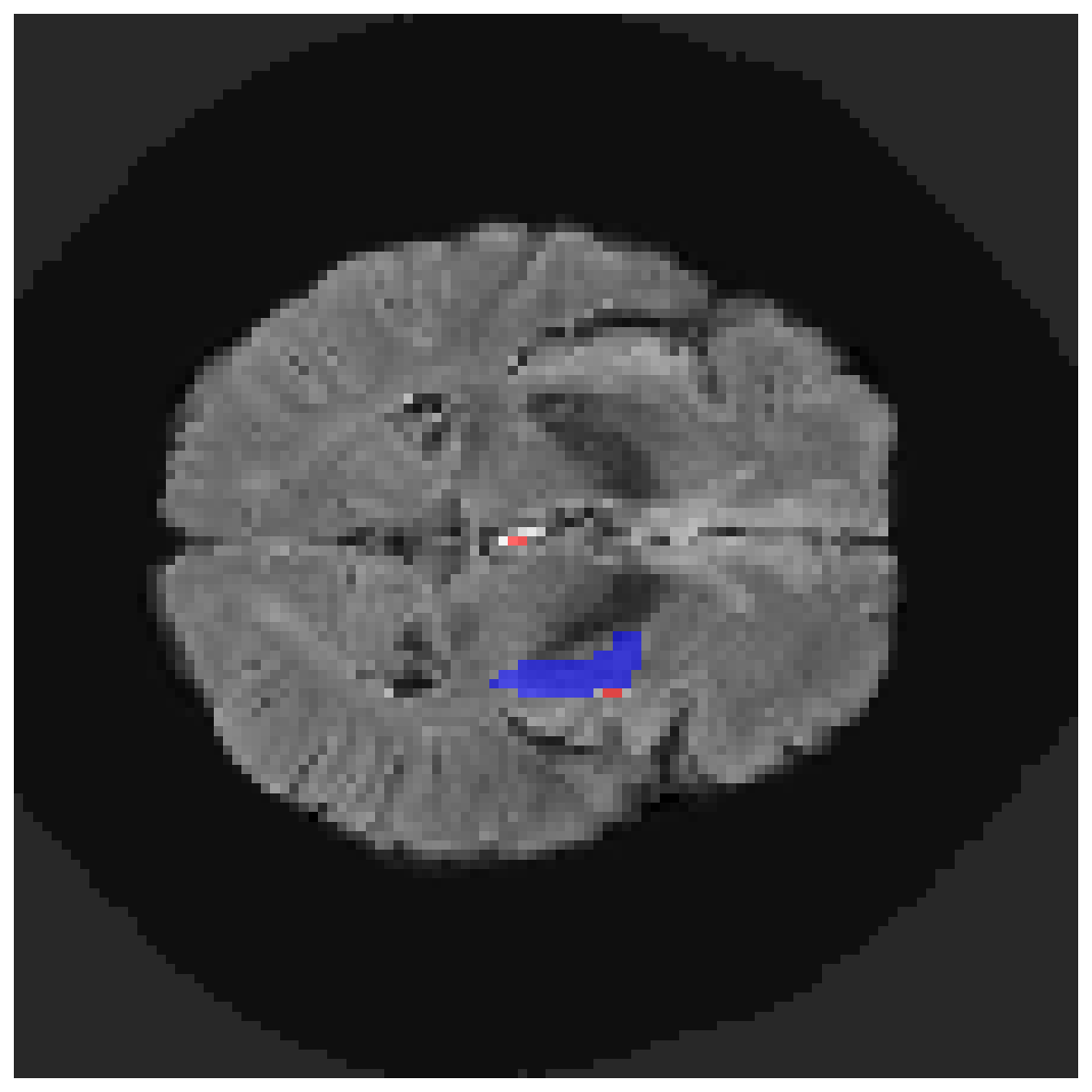}{0.00} &
\img{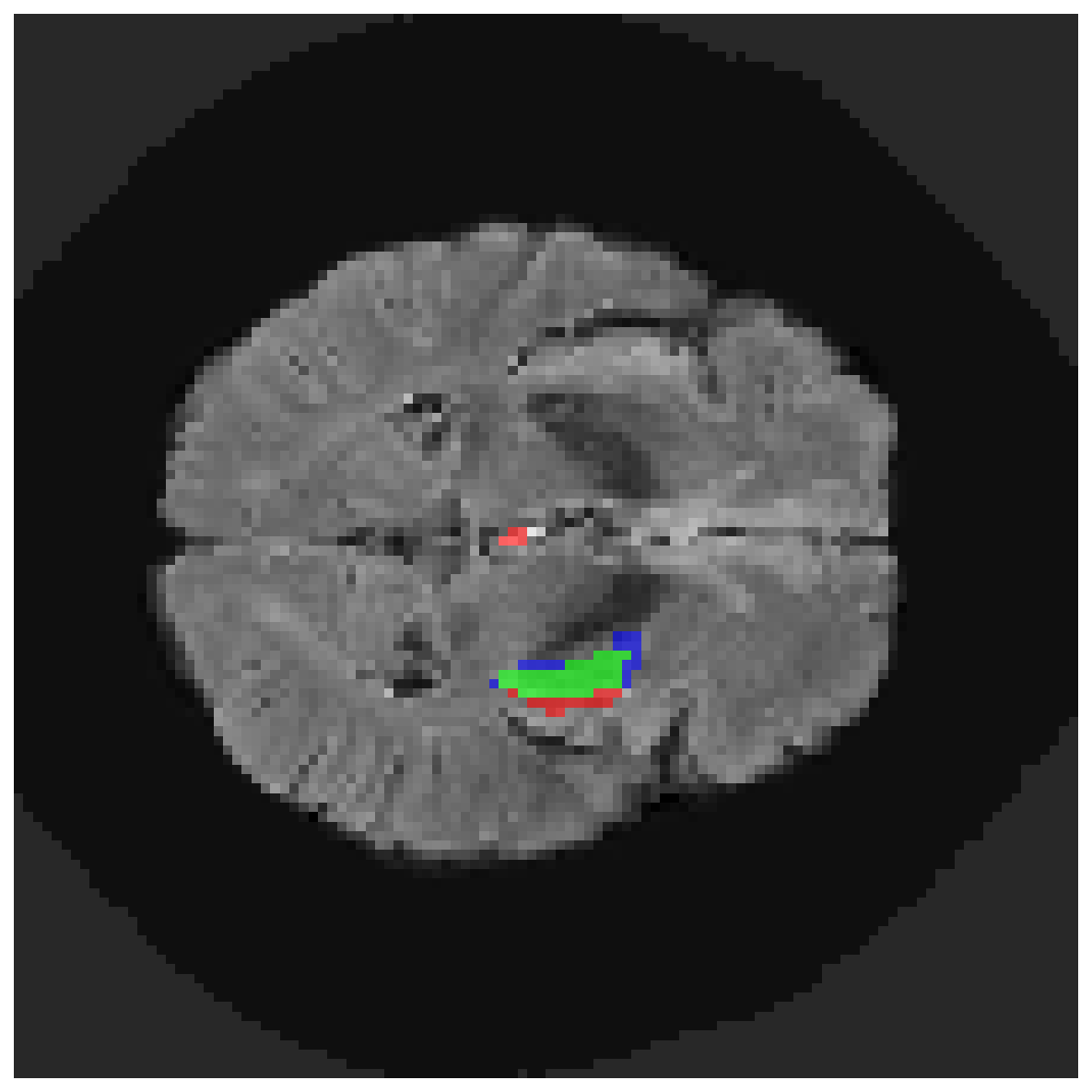}{70.40} &
\img{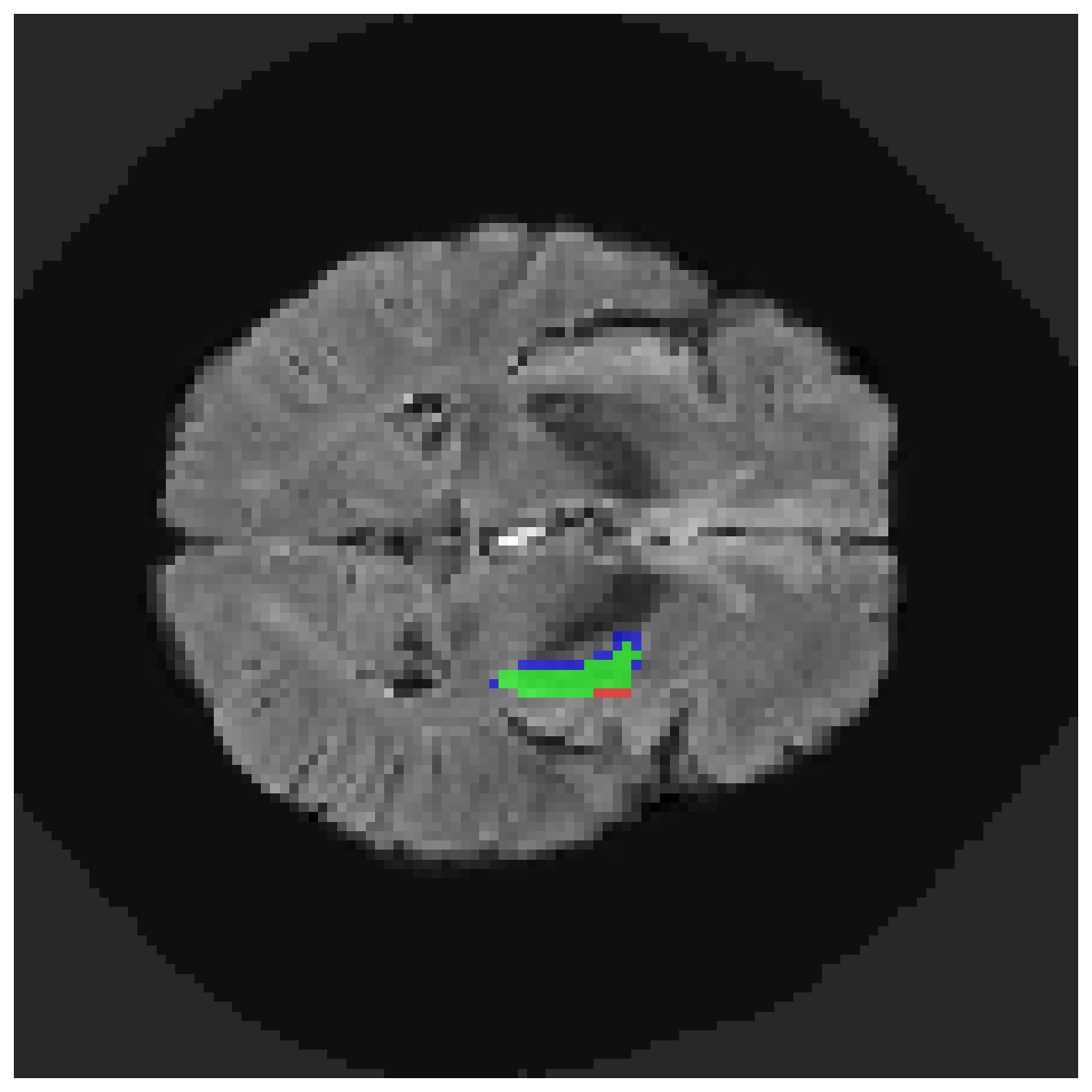}{81.82} \\

\modalityrow{All} & 
\img{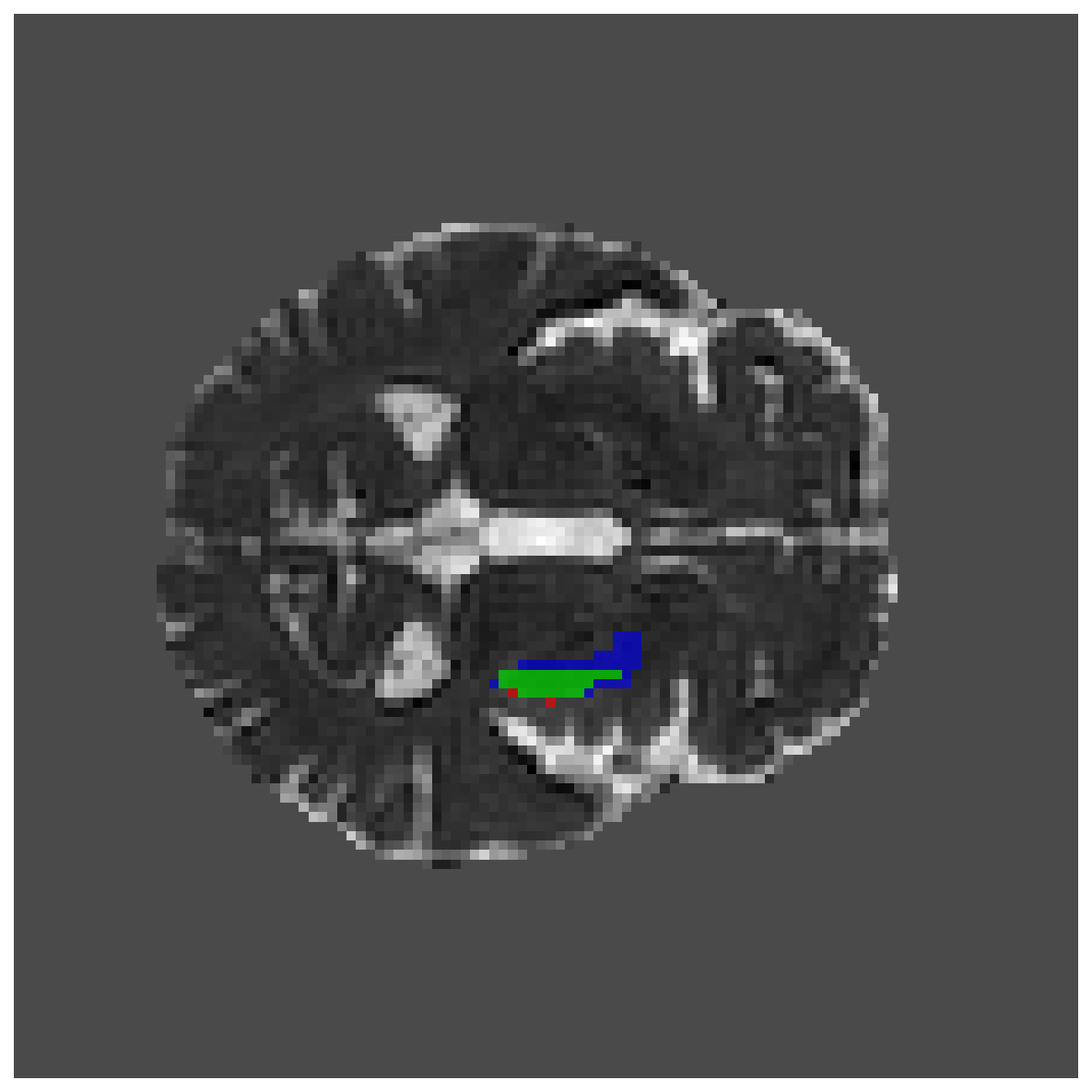}{64.52} &
\img{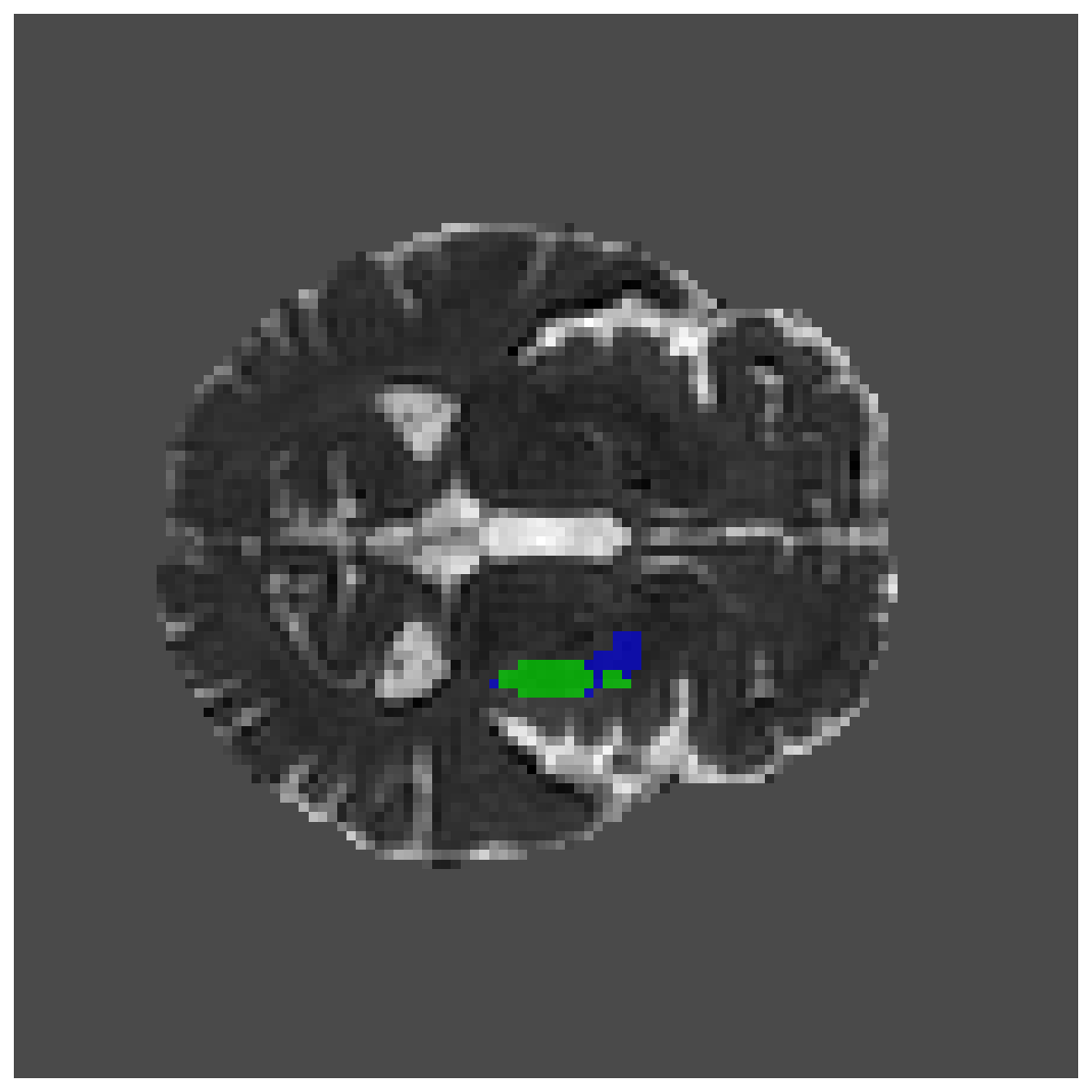}{78.00} &
\img{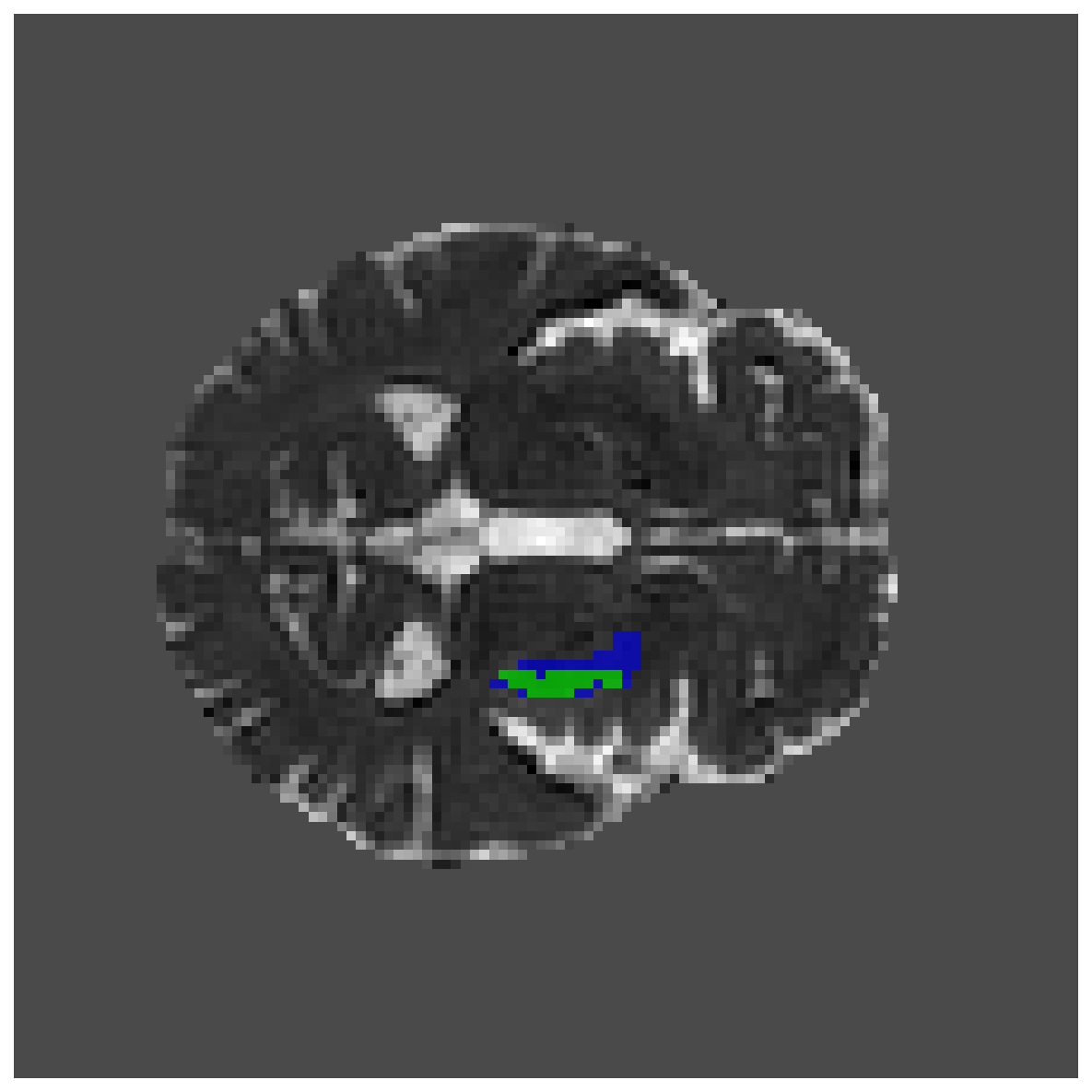}{62.92} &
\img{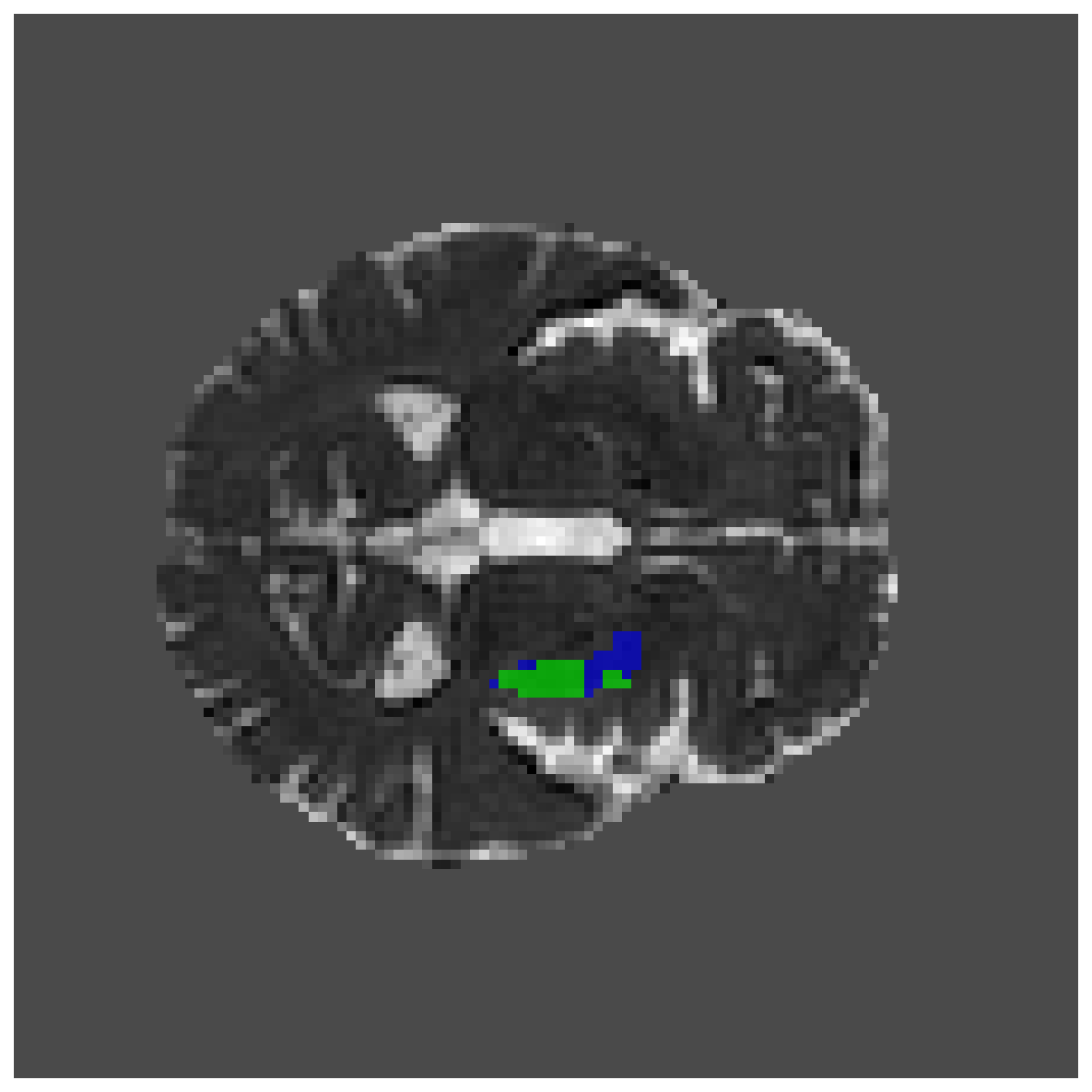}{72.92} &
\img{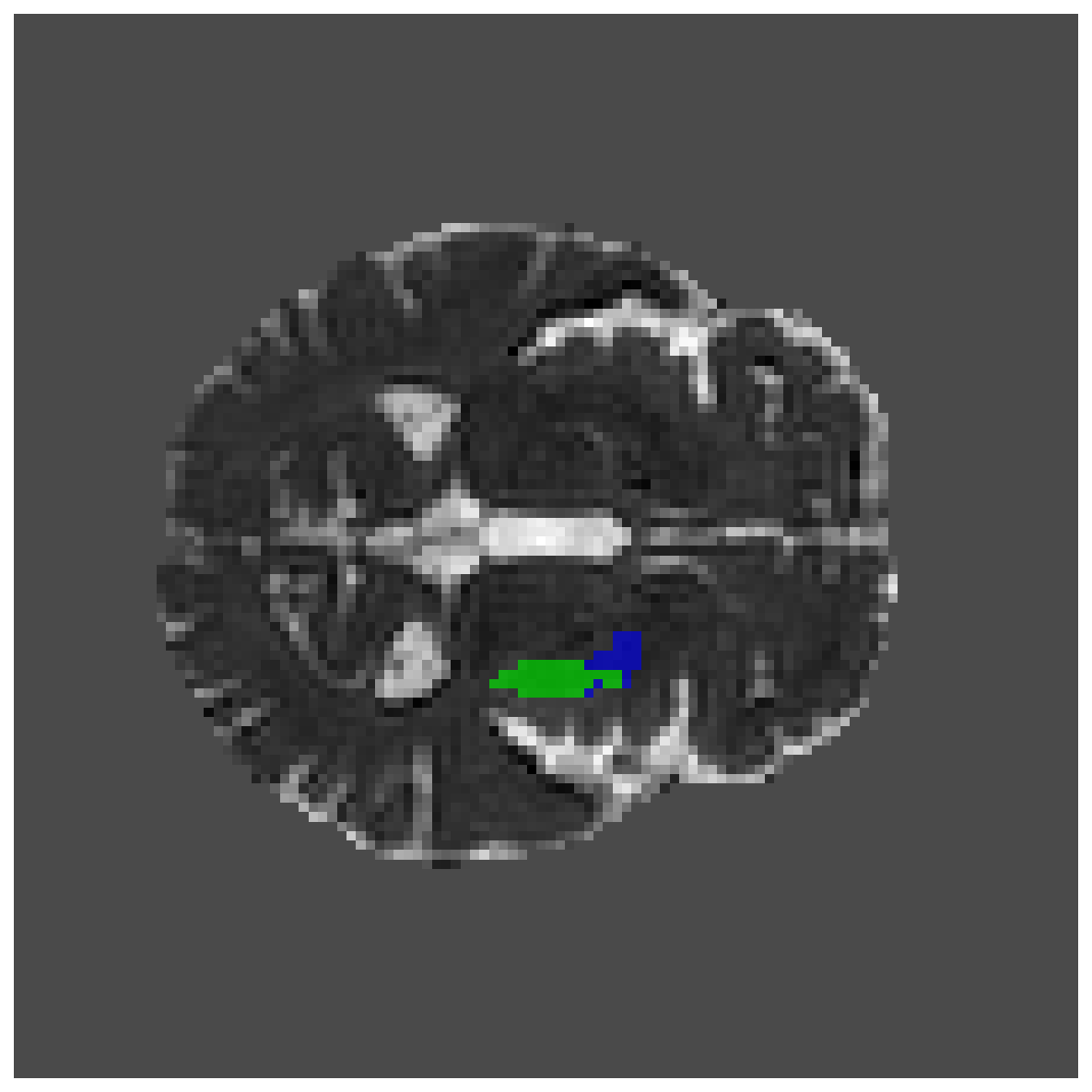}{79.21} \\

\end{tabular}
}

\begin{tikzpicture}
    \fill[green] (0,0) rectangle (0.4,0.4);
    \node[right] at (0.4,0.2) {True Positives};

    \fill[red] (5,0) rectangle (5.4,0.4);
    \node[right] at (5.4,0.2) {False Positives};

    \fill[blue] (10,0) rectangle (10.4,0.4);
    \node[right] at (10.4,0.2) {False Negatives};
\end{tikzpicture}
    
\caption{Qualitative comparison on the ISLES 2022 dataset. Predictions from different methods under single-modality and
all-modality settings are shown. Green denotes true positives, red false positives, and blue false negatives. Dice scores (\%) are reported for each example.}
\label{fig:qualitative-isles}
\end{figure*}

\paragraph{\rev{Brain tumor segmentation under missing modalities.}}
Many methods improve robustness by combining modality-specific encoders with feature aggregation. HeMIS~\citep{havaei_hemis_2016} processes each modality independently and aggregates feature statistics, UHVED~\citep{dorent_hetero-modal_2019} maps modalities into a shared variational latent space, mmFormer~\citep{zhang_mmformer_2022} uses transformer-based cross-modal attention, and ShaSpec~\citep{wang_multi-modal_2023} separates shared and modality-specific representations through dedicated branches. \rev{M³AE~\citep{liu2023m3ae} and SimMLM~\citep{li2025simmlm} likewise fuse in feature space; we compare against mmFormer, M³AE, ShaSpec, and SimMLM in \cref{sec:experiments}.} All of these operate in \emph{feature} space, whereas LARGO acts in \emph{weight} space.

\paragraph{Hypernetworks and dynamic weight generation.}
Hypernetworks generate or modulate target-network weights conditioned on an input, task, or architecture descriptor~\citep{ha_hypernetworks_2017,debrabandere_dynamic_2016,zhang_graph_2019,nava_meta-learning_2023}. LARGO is related in spirit, since it reconstructs subset-specific weights, but does not learn an auxiliary generator: it stores the entire missing-modality model family directly in a factorized tensor, indexing modality subsets through one factor matrix and reconstructing convolutional weights by tensor contraction.

\paragraph{Tensor decomposition for multimodal fusion.} Tensor factorization has also been used to model high-order cross-modal interactions, e.g.\ in tensor fusion networks, low-rank multimodal fusion, and multi-way multimodal attention~\citep{zadeh_tensor_2017,liu_efficient_2018,zhang_tdsf_2025}. These methods factorize \emph{feature} representations or fusion operators; LARGO instead factorizes \emph{model weights} across modality-subset-specific networks.

\paragraph{Tensor decomposition for network compression.}
Tensor decompositions~\citep{sidiropoulos_tensor_2017,kolda_tensor_2009} compress neural networks by approximating high-order weight tensors with low-rank factors. CP decomposition has been applied to convolutional layers to reduce parameters and computation~\citep{kim_compression_2015,lebedev_speeding_2014}, and \citet{ashtari2020low} compressed a single U-Net for brain tumor segmentation. These methods perform \emph{within-model} compression---factorizing the weights of one network. LARGO instead performs \emph{inter-model} compression: it introduces a model dimension corresponding to modality subsets and jointly compresses the entire family of $2^N-1$ models---to our knowledge, the first use of tensor decomposition to jointly parameterize the family of dedicated missing-modality models. \rev{Moreover, within-model compression decomposes one pretrained network and fine-tunes it, whereas LARGO learns all subset-specific networks jointly from scratch, so the shared factors must represent models receiving overlapping or disjoint modality subsets at once.}

\section{Method}
\label{sec:method}

\subsection{Problem formulation}

We define the base network $\tns{F}_{\vct{\theta}}$, with parameters $\vct{\theta}$, that takes all available modalities $\{\tns{X}^{(n)}\}_{n=1}^{N}$ as input. Ideally, we want to train different models on non-empty subsets of the available modalities, such as $\{\tns{F}_{{\vct{\theta}}^{(m)}}\}_{m=1}^{M}$, where $M=2^N-1$. 

Instead of training $M$ copies of $\tns{F}_{\vct{\theta}}$, we want to optimize this by applying a dimensionality constraint on the set of networks, such that the total number of parameters remains the same as for $\tns{F}_{\vct{\theta}}$. Therefore, our proposed hypernetwork LARGO is designed to approximate these models, i.e. $\tns{H}_{\vct{\varphi}}\approx\{\tns{F}_{{\vct{\theta}}^{(m)}}\}_{m=1}^{M}$, which is fundamentally different from the way existing approaches handle missing modalities.

\subsection{Low-rank hypernetwork architecture}

\subsubsection{Hypernetwork design}

We compress the networks by applying compression across the corresponding (transposed) convolutional layers.

\paragraph{Stem and head layers.} No compression is applied. We retain $M$ copies of this layer. For the stem layers, we adapt the number of input channels to the number of input modalities. This does not significantly affect the compression ratio, as these layers are very small due to the limited number of input and output channels. To obtain a compression ratio of $1/M$, we would have to set the rank to 1, which severely limits the expressive capacity of the model and leads to poor representations.

\paragraph{Inner layers.} We compress the parameters across all $M$ dedicated models using tensor decomposition. We represent the weight tensor for the $l$-th layer as $\tns{W} \in \R^{M \times C_{\text{in}} \times C_{\text{out}} \times K}$, where $C_{\text{out}}$ and $C_{\text{in}}$ are the output and input channels, and $K$ represents the flattened spatial kernel dimensions. We suppress the index $l$ in the remainder of this section for readability.

\subsubsection{Low-rank compression of the convolutional layers}
\paragraph{Convolutional layer.} In any convolutional neural network, the output neuron at location $(h,w)$ in the channel $c_\text{out}$ is computed as:

\begin{equation}
\boldsymbol{\mathcal{Y}}_{c_{\text{out}},h,w} = \left(\boldsymbol{\mathcal{X}}*\boldsymbol{\mathcal{V}}\right)_{c_{\text{out}},h,w}  = \left\langle \boldsymbol{\mathcal{S}}_{h,w},\boldsymbol{\mathcal{V}}^{(c_{\text{out}})}\right\rangle+\mathbf{b}_{c_{\text{out}}},
\end{equation}

where $\boldsymbol{\mathcal{V}}=\left\{\boldsymbol{\mathcal{V}}^{(c_{\text{out}})}\right\}_{c_{\text{out}}=1}^{C_\text{out}}$, $\boldsymbol{\mathcal{V}}^{(c_{\text{out}})}\in\R^{C_{\text{in}} \times K_h \times K_w}$ represents the tensor containing the kernel weights, with $\mathbf{b}\in \R^{C_{\text{out}}}$ the bias vector, and where $\boldsymbol{\mathcal{S}}_{h,w}$ is the sliding window operator on the input tensor $\boldsymbol{\mathcal{X}}$ defined as $\boldsymbol{\mathcal{S}}_{h,w} \triangleq \boldsymbol{\mathcal{X}}_{h-\Delta_h:h+\Delta_h,w-\Delta_w:w+\Delta_w}$, with $K_h=2\Delta_h+1$ and $K_w=2\Delta_w+1$.

\paragraph{Low-rank decomposition.} We decompose the weight tensor $\boldsymbol{\mathcal{W}}=\left\{\boldsymbol{\mathcal{V}}^{[m]}\right\}_{m=1}^M$ containing the kernel weights of the $M$ dedicated models, using Low-rank decomposition along the modes of the tensor ($M$, $C_{\text{in}}$, $C_{\text{out}}$, and $K$). For example, the Canonical Polyadic (CP) decomposition has the following structure:

\begin{align}
\boldsymbol{\mathcal{W}} &= \sum_{r=1}^{R} \vct{a}_{r} \circ \vct{b}_{r} \circ \vct{c}_{r} \circ \vct{d}_{r}
\label{eq:CPD-conv}
\end{align}

where $\vct{a}_{r} \in \R^{M}$, $\vct{b}_{r} \in \R^{C_{\text{in}}}$, $\vct{c}_{r} \in \R^{C_{\text{out}}}$, and $\vct{d}_{r} \in \R^{K}$ are the $r$-th columns of factor matrices $\boldsymbol{A}$, $\boldsymbol{B}$, $\boldsymbol{C}$, and $\boldsymbol{D}$ respectively.

This decomposition is visually represented as a tensor network diagram in \cref{fig:conv-reparameterization}\rev{, which also summarizes the three stages of the construction: stacking the dedicated kernels, factorizing the joint tensor, and reconstructing the kernel of the target subset}. In the supplementary material, we present an ablation study comparing CP and Tucker decomposition to assess the effect of the choice of decomposition.

\paragraph{Rank selection.} The rank $R$ is chosen to maintain the number of parameters per layer, thereby applying a compression of $1/M$. Each of the original convolutional layers has $C_{\text{in}}  C_{\text{out}}  K$ parameters, while the low rank convolutional layer has $(M + C_{\text{in}} +C_{\text{out}} + K) R$ parameters. Thus, we set the rank for the $1/M$ compression as follows:
\begin{equation}
R = \left\lfloor \frac{C_{\text{in}} C_{\text{out}} K}{M + C_{\text{in}} + C_{\text{out}} + K} \right\rfloor
\label{eq:rank}
\end{equation}

In the supplementary material, we present an ablation study on the rank $R$, comparing different values to assess the effect of varying compression ratios.

\paragraph{Bias.} We retain unique bias terms for all $M$ layers, resulting in $C_{\text{out}} \times M$ parameters. This has only a small effect on the total number of parameters.

\paragraph{Initialization.} The weight matrices $\mtx{B}$, $\mtx{C}$, and $\mtx{D}$ representing the input channels, output channels, and flattened spatial kernel dimensions, respectively, are initialized using Kaiming normal initialization. The weight matrix $\mtx{A}$, representing the number of models, is initialized with ones, and the bias weights with zeros. Note that weight matrices are trained from scratch, so we do not compute the CPD from pretrained weights.

\paragraph{Normalization.} To enforce numerical stability and improve gradient flow, we normalize the atoms corresponding to the input channels, output channels, and kernel dimensions. The atom containing $M$ is not normalized. This does not affect the low-rank representations, because the norms of these three atoms can be absorbed into the other atom.

\paragraph{Forward pass and modality selection.} During inference, given a specific modality combination $m$, the hypernetwork reconstructs the dedicated weights corresponding to $m$. For CP decomposition, the forward pass computes:
\begin{equation}
\tns{W}^{(m)} = \sum_{r=1}^{R} A_{m,r} \left( \vct{b}_{r} \circ \vct{c}_{r} \circ \vct{d}_r \right)
\end{equation}
where $A_{m,r}$ is a scalar value at position $m$ in the rank vector $\vct{a}_{r}$. This is equivalent to computing the whole joint kernel weights tensor $\tns{W}$, and then selecting the kernels for model $m$, but avoids the computational overhead of computing unused parameters. The construction of the hypernetwork is visually represented in \cref{fig:hypernetwork}.

\begin{table}[t]
\vspace{-15pt}
    \centering
    \caption{Dice scores (\%)~$\uparrow$ on BraTS 2018 (5-fold cross-validation; mmF: mmFormer, ShaS: ShaSpec, SimM: SimMLM). $\bullet$/$\circ$: modality present/absent. Best and second-best scores per tumor region in \textbf{bold} and \underline{underlined}.}
    \setlength{\tabcolsep}{2pt}%
    {\scriptsize\begin{tabular*}{\textwidth}{@{\extracolsep{\fill}}ccccccccccccccccccc@{}}
        \toprule
        \multirow{2}{*}[-0.4ex]{\textbf{Fl}} & \multirow{2}{*}[-0.4ex]{\textbf{T1}} & \multirow{2}{*}[-0.4ex]{\textbf{T1c}} & \multirow{2}{*}[-0.4ex]{\textbf{T2}} & \multicolumn{5}{c}{\textbf{Whole tumor (WT)}} & \multicolumn{5}{c}{\textbf{Tumor core (TC)}} & \multicolumn{5}{c}{\textbf{Enhancing tumor (ET)}}\\
        \cmidrule(lr){5-9}\cmidrule(lr){10-14}\cmidrule(lr){15-19}
         &  &  &  & mmF & M³AE & ShaS & SimM & \makecell{LARGO\\(Ours)} & mmF & M³AE & ShaS & SimM & \makecell{LARGO\\(Ours)} & mmF & M³AE & ShaS & SimM & \makecell{LARGO\\(Ours)} \\
        \midrule
    $\bullet$&$\circ$&$\circ$&$\circ$&
    86.71&87.28&87.50&\underline{88.20}&\textbf{88.85}&
    66.73&67.82&\underline{71.15}&68.27&\textbf{72.28}&
    35.97&42.09&\underline{43.27}&42.33&\textbf{43.74}\\
    $\circ$&$\bullet$&$\circ$&$\circ$&
    68.38&74.32&77.09&\underline{78.06}&\textbf{78.41}&
    60.20&64.54&\underline{70.15}&68.02&\textbf{70.51}&
    29.41&36.84&\underline{42.12}&40.02&\textbf{42.36}\\
    $\circ$&$\circ$&$\bullet$&$\circ$&
    74.95&74.49&\underline{77.32}&76.05&\textbf{79.13}&
    80.71&80.54&\underline{82.77}&81.97&\textbf{83.70}&
    73.87&74.34&75.07&\underline{75.88}&\textbf{76.21}\\
    $\circ$&$\circ$&$\circ$&$\bullet$&
    83.26&84.43&84.70&\underline{85.17}&\textbf{86.20}&
    66.31&68.44&\underline{71.88}&69.89&\textbf{72.44}&
    37.91&42.59&\underline{46.35}&44.76&\textbf{46.65}\\
    $\bullet$&$\bullet$&$\circ$&$\circ$&
    88.35&88.82&\underline{88.92}&88.53&\textbf{89.84}&
    71.80&71.51&\textbf{75.39}&70.19&\underline{75.11}&
    40.67&44.18&\underline{46.47}&43.14&\textbf{47.61}\\
    $\bullet$&$\circ$&$\bullet$&$\circ$&
    88.89&\underline{89.28}&89.04&88.70&\textbf{90.09}&
    82.81&\underline{84.06}&83.85&82.63&\textbf{84.56}&
    75.11&\textbf{76.31}&75.76&75.99&\underline{76.03}\\
    $\bullet$&$\circ$&$\circ$&$\bullet$&
    88.82&\underline{89.32}&89.12&89.30&\textbf{89.99}&
    71.40&71.62&\underline{74.50}&71.56&\textbf{74.99}&
    42.60&45.27&\underline{47.59}&46.04&\textbf{48.83}\\
    $\circ$&$\bullet$&$\bullet$&$\circ$&
    78.07&76.83&\underline{79.90}&79.35&\textbf{80.45}&
    82.19&82.28&\underline{83.69}&82.50&\textbf{84.10}&
    75.73&74.59&\underline{76.29}&75.89&\textbf{76.82}\\
    $\circ$&$\bullet$&$\circ$&$\bullet$&
    85.13&85.88&\underline{86.27}&85.58&\textbf{86.88}&
    70.61&71.03&\textbf{74.37}&71.30&\underline{73.85}&
    41.88&44.63&\underline{47.76}&45.66&\textbf{48.11}\\
    $\circ$&$\circ$&$\bullet$&$\bullet$&
    86.16&85.60&\underline{86.23}&85.83&\textbf{87.20}&
    83.47&83.87&\underline{84.15}&82.61&\textbf{84.80}&
    76.62&\underline{76.23}&75.93&76.01&\textbf{76.82}\\
    $\bullet$&$\bullet$&$\bullet$&$\circ$&
    89.32&\underline{89.39}&89.35&88.82&\textbf{90.22}&
    83.46&\textbf{84.41}&84.26&82.77&\underline{84.37}&
    76.41&76.51&\underline{77.05}&75.98&\textbf{77.35}\\
    $\bullet$&$\bullet$&$\circ$&$\bullet$&
    89.27&\underline{89.52}&\underline{89.52}&89.35&\textbf{90.14}&
    73.04&72.89&\underline{75.94}&72.45&\textbf{76.09}&
    45.57&46.00&\underline{48.80}&46.62&\textbf{49.22}\\
    $\bullet$&$\circ$&$\bullet$&$\bullet$&
    89.80&\underline{90.01}&89.73&89.46&\textbf{90.49}&
    83.60&\underline{84.25}&84.13&82.71&\textbf{84.32}&
    76.03&\underline{76.74}&75.57&76.03&\textbf{77.42}\\
    $\circ$&$\bullet$&$\bullet$&$\bullet$&
    86.51&85.80&\underline{86.66}&85.86&\textbf{87.47}&
    83.48&84.05&\underline{84.50}&82.78&\textbf{84.73}&
    76.75&76.48&\underline{76.82}&75.99&\textbf{77.09}\\
    $\bullet$&$\bullet$&$\bullet$&$\bullet$&
    89.93&\underline{90.09}&89.88&89.44&\textbf{90.44}&
    83.77&\textbf{84.69}&84.40&82.73&\underline{84.47}&
    76.13&\underline{76.59}&76.40&76.02&\textbf{78.18}\\
    \midrule
    \multicolumn{4}{c}{Average}&
    84.90&85.40&\underline{86.08}&85.85&\textbf{87.05}&
    76.24&77.07&\underline{79.01}&76.82&\textbf{79.35}&
    58.71&60.63&\underline{62.08}&61.09&\textbf{62.83}\\
        \bottomrule
    \end{tabular*}}
    \label{tab:brats2018-baselines}
\end{table}

\subsection{Training procedure}

\paragraph{Loss function.} We use the standard segmentation loss $\mathcal{L}_\text{seg} = \mathcal{L}_\text{Dice} + \mathcal{L}_\text{CE}$. We train the hypernetwork with random modality dropout combined with full-modality guidance: at each iteration, we sample a modality subset $m \sim \mathcal{U}(1,M)$ and minimize
\begin{equation}
\mathcal{L}_\text{total} = \tfrac{1}{2}\bigl(\mathcal{L}_\text{seg}(\tns{H}_{\vct{\varphi}}(\tns{X}_m;m),\tns{Y}) + \mathcal{L}_\text{seg}(\tns{H}_{\vct{\varphi}}(\tns{X};M),\tns{Y})\bigr),
\label{eq:loss-total}
\end{equation}
where $\tns{X}_m$ is the input with modality combination $m$, $\tns{X}$ is the full-modality input, and $\tns{Y}$ the target. The first term ensures the hypernetwork handles all $2^N-1$ missing-modality scenarios; the second term provides full-modality guidance throughout training.

\begin{wraptable}[12]{r}{0.58\textwidth}
    \vspace{-16pt}
    \centering
    \caption{Model complexity on BraTS 2018\rev{, measured on the same hardware with the full-modality input}. Best and second-best scores in \textbf{bold} and \underline{underlined}.}
    \setlength{\tabcolsep}{3pt}%
    {\scriptsize\begin{tabular*}{\linewidth}{@{\extracolsep{\fill}}lS[table-format=3.1]cccS[table-format=1.2]@{}}
        \toprule
        Model & {\makecell{\# Params\\(M)}} & \rev{\makecell{Time\\per epoch}} & \rev{GFLOPs} & \makecell{Avg Dice\\(\%) $\uparrow$} & {\makecell{Speed\\(vol/s) $\uparrow$}} \\
        \midrule
        mmFormer & 36.7 & \rev{2m\,30s} & \rev{138.8} & 73.28 $\pm$ 2.31 & 0.89 \\
        M³AE & 2.1 & \rev{1m\,07s} & \rev{333.7} & 74.37 $\pm$ 0.99 & \underline{1.02}\\
        ShaSpec & 187.7 & \rev{2m\,42s} & \rev{1411.9} & \underline{75.73} $\pm$ 0.89& 0.68 \\
        SimMLM & 7.8 & \rev{1m\,12s} & \rev{242.1} & 74.59 $\pm$ 0.80 & \textbf{1.14}\\
        \midrule
        LARGO (Ours) & 22.6 & \rev{1m\,51s} & \rev{1155.1} & \textbf{76.41} $\pm$ 0.78 & \textbf{1.14}\\
        \bottomrule
    \end{tabular*}}
    \label{tab:comparison-brats}
\end{wraptable}

\section{Experiments}
\label{sec:experiments}

\subsection{Experimental setup}

\paragraph{Baselines.} We compare our proposed hypernetwork LARGO against four state-of-the-art missing modality methods: mmFormer~\citep{zhang_mmformer_2022}, M³AE~\citep{liu2023m3ae}, ShaSpec~\citep{wang_multi-modal_2023}, and SimMLM~\citep{li2025simmlm}. To ensure fair comparison, all models are trained from scratch under identical conditions\rev{: the same folds, learning rate, batch size, scheduler, optimizer, preprocessing, augmentation, patch size, cross-validation protocol, and evaluation procedure, without method-specific hyperparameter tuning, so the reported differences isolate architectural contributions.}

\paragraph{Architecture.} We use nnUnet~\citep{isensee_nnu-net_2021} as base architecture for LARGO, which automatically configures the U-Net topology based on dataset properties. For avMNIST, we use a different architecture. Details of the model architecture are provided in the supplementary material.

\begin{revblock}
\paragraph{Statistical analysis.} Each patient contributes one paired observation (Dice averaged over all modality subsets and, for BraTS, tumor regions); we apply two-sided paired Wilcoxon signed-rank tests of LARGO against each baseline with Benjamini--Hochberg correction (\cref{tab:significance}, supplementary material). All patient-level comparisons remain significant, as do 57 of 60 per-subset comparisons on BraTS 2018 and 26 of 28 on ISLES 2022.
\end{revblock}

\subsection{BraTS 2018 dataset}

\paragraph{Dataset.} The BraTS 2018 dataset~\citep{brats1,brats2,brats3} contains 285 brain MRI cases with four modalities: FLAIR (fluid-attenuated inversion recovery), T1-weighted, T1-weighted with contrast enhancement (T1c), and T2-weighted. The task is to segment three tumor regions: whole tumor (WT), tumor core (TC), and enhancing tumor (ET). With four modalities, there are 15 possible missing-modality scenarios.

\paragraph{Training.} All models are trained for 750 epochs using stochastic gradient descent with Nesterov momentum (0.99), learning rate ($5\times 10^{-4}$), and weight decay ($10^{-5}$).

\begin{wraptable}[15]{r}{0.5\textwidth}
    \vspace{-16pt}
\centering
\caption{Dice scores (\%)~$\uparrow$ on ISLES 2022 (5-fold cross-validation). $\bullet$/$\circ$: modality present/absent. Best and second-best scores in \textbf{bold} and \underline{underlined}.}
    \setlength{\tabcolsep}{3pt}%
\adjustbox{max width=\linewidth}{\scriptsize\begin{tabular}{cccccccc}
    \toprule
    \multirow{3}{*}[-1ex]{\textbf{DWI}} &
    \multirow{3}{*}[-1ex]{\textbf{ADC}} &
    \multirow{3}{*}[-1ex]{\textbf{FLAIR}} &
    \multicolumn{5}{c}{\textbf{Stroke lesion}}\\
    \cmidrule(lr){4-8}
     &  & & mmFormer & M$^3$AE & ShaSpec & SimMLM & \makecell{LARGO\\(Ours)} \\
    \midrule
    $\bullet$ & $\circ$ & $\circ$ & 66.39 & 72.20 & 62.76 & \underline{74.11} & \textbf{74.23} \\
    $\circ$ & $\bullet$ & $\circ$ & \underline{40.03} & 37.79 & 24.47 & 39.92 & \textbf{42.49} \\
    $\circ$ & $\circ$ & $\bullet$ & 35.05 & \underline{38.60} & 32.63 & 38.28 & \textbf{41.09}  \\
    $\bullet$ & $\bullet$ & $\circ$ & 69.33 & 74.10 & 65.10 & \underline{74.39} & \textbf{74.67} \\
    $\bullet$ & $\circ$ & $\bullet$ & 67.12 & 72.88 & 64.50 & \underline{74.54} & \textbf{75.75} \\
    $\circ$ & $\bullet$ & $\bullet$ & 50.50 & \underline{50.82} & 41.54 & 44.51 & \textbf{53.70}  \\
    $\bullet$ & $\bullet$ & $\bullet$ & 69.76 & \underline{74.62} & 65.44 & 74.42 & \textbf{76.82} \\
    \midrule
    \multicolumn{3}{c}{Average} & 56.88 & \underline{60.14} & 50.92 & 60.02 & \textbf{62.68} \\
    \bottomrule
\end{tabular}}
\label{tab:isles-baselines}
\end{wraptable}

\paragraph{Quantitative results.} \Cref{tab:brats2018-baselines} presents comprehensive Dice scores across all 15 missing-modality combinations. LARGO consistently achieves the best performance, ranking first in all 15 scenarios of WT, 11 out of 15 for TC, and 14 out of 15 for ET, and second in all other cases. On average, LARGO achieves 87.05\% for WT, 79.35\% for TC, and 62.83\% for ET, outperforming all baselines. Notably, ShaSpec achieves competitive results but requires 8.3$\times$ more parameters (187.7M vs. 22.6M). \rev{The largest gains occur in the difficult single-modality cases (e.g., T1-only), indicating that the shared factors preserve modality-specific behavior rather than collapsing to one generic representation; all gains are significant at the patient level (\cref{tab:significance}).}

\begin{wraptable}[10]{r}{0.5\textwidth}
    \vspace{-16pt}
    \centering
    \caption{Model complexity on ISLES 2022. Best and second-best scores in \textbf{bold} and \underline{underlined}.}
    \setlength{\tabcolsep}{3pt}%
    \adjustbox{max width=\linewidth}{\scriptsize\begin{tabular}{l S[table-format=3.1] c S[table-format=1.2]}
        \toprule
        Model & {\# Params (M)} & {Avg Dice (\%) $\uparrow$} & {Speed (vol/s) $\uparrow$} \\
        \midrule
        mmFormer & 35.3 & 56.88 $\pm$ 4.00 & 1.28\\
        M³AE     &  21.7 & \underline{60.14} $\pm$ 3.13 & \textbf{4.76}\\
        ShaSpec  & 151.7 & 50.92 $\pm$ 2.51 & 1.56\\
        SimMLM   &  5.9 & 60.02 $\pm$ 1.99 & 3.56 \\
        \midrule
        LARGO (Ours)     & 22.4 & \textbf{62.67} $\pm$ 2.28 & \underline{3.88}\\
        \bottomrule
    \end{tabular}}
    \label{tab:comparison-isles}
\end{wraptable}

\Cref{tab:comparison-brats} compares model complexity and efficiency. LARGO achieves the highest average Dice score (76.41\%) while maintaining competitive inference speed (1.14 vol/s) and moderate parameter count (22.6M). \rev{LARGO is not the smallest or fastest model in absolute terms: mmFormer and M³AE use narrow convolutions and low-resolution transformer modules, whereas LARGO retains a full-width nnU-Net. Its compression claim is relative to the family it replaces: 15 dedicated nnU-Nets need about 338.6M parameters, whereas LARGO represents all of them with 22.60M, the budget of a single nnU-Net (22.57M), a $\sim$15$\times$ reduction. It also trains in a single stage, with an epoch time below mmFormer and ShaSpec.}

\paragraph{Qualitative results.} \Cref{fig:qualitative-brats} visualizes segmentation results for single-modality scenarios. Yellow indicates enhancing tumor, red indicates necrotic/non-enhancing tumor, and green indicates edema. LARGO produces more accurate segmentations with fewer false positives and false negatives compared to all baselines, particularly in challenging single-modality cases with limited information.

\subsection{ISLES 2022 dataset}

\paragraph{Dataset.} The ISLES 2022 dataset~\citep{de_la_rosa_deepisles_2025,hernandez_petzsche_isles_2022} contains 250 stroke MRI cases with three modalities: DWI (diffusion-weighted imaging), ADC (apparent diffusion coefficient), and FLAIR (fluid-attenuated inversion recovery). The task is to segment acute and sub-acute ischemic stroke lesions. With three modalities, there are 7 possible missing-modality scenarios.

\paragraph{Training.} All models are trained for 500 epochs using stochastic gradient descent with Nesterov momentum (0.99), learning rate ($5\times 10^{-4}$), and weight decay ($10^{-5}$).

\paragraph{Quantitative results.} \Cref{tab:isles-baselines} presents Dice scores across all 7 modality combinations. LARGO scores best in all 7 scenarios, with an average Dice score of 62.67\%, outperforming M³AE (60.14\%), SimMLM (60.02\%), mmFormer (56.88\%), and ShaSpec (50.92\%). The improvement is particularly pronounced in challenging scenarios with limited modalities, such as ADC-only (42.49\% vs. 40.03\%) and FLAIR-only (41.09\% vs. 38.60\%). \rev{All gains are significant at the patient level (\cref{tab:significance}).}

\begin{wraptable}[10]{r}{0.5\textwidth}
    \vspace{-16pt}
    \caption{Accuracy (\%)~$\uparrow$ on avMNIST (predefined train-test split). $\bullet$/$\circ$: modality present/absent. Best and second-best scores in \textbf{bold} and \underline{underlined}.}
    \centering
    \setlength{\tabcolsep}{3pt}%
    \adjustbox{max width=\linewidth}{\scriptsize\begin{tabular}{ccccccc}
        \toprule
        \multirow{3}{*}[-1ex]{\textbf{Image}} & \multirow{3}{*}[-1ex]{\textbf{Audio}} & \multicolumn{5}{c}{\textbf{Classification accuracy}}\\
        \cmidrule(lr){3-7}
         & & mmFormer & M³AE & ShaSpec & SimMLM & \makecell{LARGO\\(Ours)} \\
        \midrule
        $\bullet$ & $\circ$ & \underline{94.67} & 93.33 & 94.00 & {94.00} & \textbf{96.00} \\
        $\circ$ & $\bullet$ & \textbf{95.67} & \underline{95.00} & \underline{95.00} & 94.67 & \textbf{95.67} \\
        $\bullet$ & $\bullet$ & \textbf{99.33} & \textbf{99.33} & \textbf{99.33} & \textbf{99.33} & \textbf{99.33} \\
        \midrule
        \multicolumn{2}{c}{Average} & \underline{96.56} & 95.89 & 96.11 & 96.00 & \textbf{97.00} \\
        \bottomrule
    \end{tabular}}
    \label{tab:avmnist-baselines}
\end{wraptable}

\Cref{tab:comparison-isles} compares model complexity and efficiency. LARGO achieves the highest average Dice score (62.67\%) with 22.4M parameters and competitive inference speed (3.88 vol/s). While M³AE is faster (4.76 vol/s), it achieves lower accuracy (60.14\%). This demonstrates our method's effectiveness on stroke lesion segmentation across diverse missing-modality scenarios.

\paragraph{Qualitative results.} \Cref{fig:qualitative-isles} visualizes segmentation results for single-modality scenarios. True positives appear in green, false positives in red, and false negatives in blue. LARGO consistently produces more accurate lesion boundaries and reduces false predictions compared to all baselines, demonstrating robust performance even with severely limited input information.

\subsection{avMNIST dataset}

As a proof-of-concept, we evaluate LARGO on a non-medical classification task with heterogeneous modalities.

\begin{wraptable}[9]{r}{0.5\textwidth}
    \vspace{-16pt}
    \centering
    \caption{Model complexity on avMNIST. Best and second-best scores in \textbf{bold} and \underline{underlined}.}
    \setlength{\tabcolsep}{3pt}%
    {\scriptsize\begin{tabular*}{\linewidth}{@{\extracolsep{\fill}}lccc@{}}
        \toprule
        Model & \makecell{\# Params\\(k)} & \makecell{Avg Acc\\(\%) $\uparrow$} & \makecell{Speed\\(samples/s) $\uparrow$} \\
        \midrule
        mmFormer & 204 & \underline{96.56} & 1792 \\
        M³AE & 206 & 95.89 & \underline{1984} \\
        ShaSpec & 195 & 96.11 & \textbf{2112} \\
        SimMLM & 203 & 96.00 & 1120 \\
        \midrule
        LARGO (ours) & 202 & \textbf{97.00} & \underline{1984} \\
        \bottomrule
    \end{tabular*}}
    \label{tab:comparison-avmnist}
\end{wraptable}

\paragraph{Dataset.} The audiovisual MNIST (avMNIST dataset)\footnote{\href{https://github.com/Jakobovski/free-spoken-digit-dataset}{https://github.com/Jakobovski/free-spoken-digit-dataset}} provides 3000 samples (2100 train + 900 test) with 2 modalities: images and audio. The images are taken from the MNIST dataset, while audio samples are recorded from 6 different speakers (50 per digit).
\paragraph{Training.} We employ a two-stage training protocol to ensure fair comparison across all models. In the first stage, modality-specific encoders (ImageEncoder and AudioEncoder) are pretrained independently for 100 epochs on their respective datasets: MNIST for visual features and FSDD for audio features. During the second stage, these pretrained encoders are frozen and integrated with model-specific fusion architectures. All models are then trained end-to-end for 100 epochs using random modality dropout.

For a fair comparison, the fusion bottleneck is 64 features for all models, the frozen encoders (approximately 74k parameters) are shared, the baseline-specific fusion architectures have approximately the same number of parameters, and all models use a batch size of 64, Adam with learning rate 0.001, and the cross-entropy loss.

\paragraph{Quantitative results.} \Cref{tab:avmnist-baselines} presents classification accuracy scores across all 3 missing-modality combinations. LARGO achieves the best average performance: +0.44\% increase in accuracy compared to its closest competitors. This experiment shows that LARGO is flexible and can be extended to non-medical use cases, classification tasks, and heterogeneous modalities. \Cref{tab:comparison-avmnist} compares model complexity and efficiency: LARGO achieves the highest average accuracy (97\%) at the same parameter budget.

\section{Discussion of model complexity}
\label{sec:complexity}

\paragraph{Asymptotic complexity.} For each convolutional layer, dynamic weight reconstruction performs a tensor contraction of cost $O(M C_{\text{in}} C_{\text{out}} K R)$, while the subsequent convolution costs $O(C_{\text{in}} C_{\text{out}} K S)$, where $S$ is the spatial size of the feature map.

\begin{wraptable}[9]{r}{0.5\textwidth}
\centering
\vspace{-16pt}
\caption{Computational overhead of dynamic weight reconstruction in LARGO relative to a single nnU-Net backbone, measured on BraTS~2018 ($N=4$, $M=15$).}
\label{tab:complexity}
    \setlength{\tabcolsep}{3pt}%
{\scriptsize\begin{tabular*}{\linewidth}{@{\extracolsep{\fill}}lccc@{}}
    \toprule
    Metric & nnU-Net & LARGO (Ours) & $\Delta$ \\
    \midrule
    Parameters       & 22{,}574{,}563  & 22{,}596{,}253  & $+0.1\%$ \\
    GFLOPs           & 1{,}160.2       & 1{,}445.8       & $+24.6\%$ \\
    Peak VRAM (MB)   & 4{,}910.79      & 6{,}144.98      & $+25.1\%$ \\
    \bottomrule
\end{tabular*}}
\end{wraptable}

\paragraph{Stem and head layers.} Stem and head are kept uncompressed. Their parameters grow exponentially with $N$, but on BraTS 2018 they account for $28{,}128$ and $1{,}485$ parameters respectively---only $0.131\%$ of the total. The bulk of the budget lies in the inner convolutional layers, so the stem/head overhead remains modest as $N$ grows. LARGO is therefore practical for $N \lesssim 10$, which covers most standard multimodal settings. \rev{For larger $N$, the stem can be padded to a fixed modality dimension and factorized along the model mode as well; we leave this to future work.}

\paragraph{Empirical overhead.} \Cref{tab:complexity} compares LARGO to its underlying nnU-Net backbone on BraTS~2018. The total parameter count is essentially identical ($+0.1\%$): LARGO matches a single dedicated model in size while implicitly representing all $M=15$ subset-specific models. Dynamic reconstruction increases inference compute by $24.6\%$ in GFLOPs and peak training memory by $25.1\%$. This is acceptable since missing-modality models typically add operations (shared encoders, gating modules, etc.) on top of standard segmentation backbones. \rev{Inference cost varies by less than 1\% with the number of available modalities (\cref{tab:cost-vs-modalities}, supplementary material), since only the uncompressed stem depends on it.}

\begin{revblock}
\paragraph{Scaling with the number of modalities.} Subset selection is an $O(1)$ row lookup in $\mtx{A}$ (no learned router). With the rank recomputed from \cref{eq:rank}, going from $M=15$ to $M=255$ subsets raises the parameter count by only 3.9\% (stem/head share below 4\%), while GFLOPs decrease (\cref{tab:scaling}, supplementary material). The $1/M$ budget is a target rather than a constraint: even $R/4$ costs only about 1.6 Dice points (rank ablation, supplementary material). This is architectural evidence only; accuracy beyond four modalities remains to be validated.
\end{revblock}

\section{Conclusion}

We introduced LARGO, a principled hypernetwork framework for handling missing modalities that leverages CP decomposition to compress $2^N-1$ dedicated models into a single network by exploiting structural similarities in convolutional kernels across modality-specific models.

Experimental evaluation on the BraTS 2018, ISLES 2022, and avMNIST datasets demonstrates that LARGO consistently outperforms state-of-the-art baselines across diverse missing-modality scenarios. In future work, we will explore more expressive, nonlinear hypernetwork formulations to further improve the weight approximation.

\begin{revblock}
\paragraph{Limitations.} Our empirical validation covers only convolutional backbones and $N\le 4$ modalities. The CP factorization applies unchanged to matrix-valued weights (as in the linear layers of the avMNIST model), so transformer attention and MLP layers are covered algebraically but not validated; likewise, the scaling analysis in \cref{sec:complexity} does not establish that accuracy is preserved for $N>4$. Dynamic reconstruction adds roughly 25\% compute and memory, and the uncompressed stem and head still grow with $2^N-1$ unless factorized.
\end{revblock}
\begin{ack}

This work is funded by the Flanders AI Research Program, under grant no. 174B09119. The computational resources (Stevin Supercomputer Infrastructure) and services used in this work were provided by the VSC (Flemish Supercomputer Center), funded by Ghent University, FWO, and the Flemish Government -- department EWI. \rev{The authors declare no competing interests.}

\end{ack}

\bibliography{main}

@String(ICLR = {Int. Conf. Learn. Represent.})

@String(AAAI = {AAAI})

@String(ICLR  = {ICLR})

@article{isensee_nnu-net_2021,
	title = {{nnU}-Net: a self-configuring method for deep learning-based biomedical image segmentation},
	volume = {18},
	rights = {2020 The Author(s), under exclusive licence to Springer Nature America, Inc.},
	issn = {1548-7105},
	url = {https://www.nature.com/articles/s41592-020-01008-z},
	doi = {10.1038/s41592-020-01008-z},
	shorttitle = {{nnU}-Net},
	pages = {203--211},
	number = {2},
	journal = {Nature Methods},
	shortjournal = {Nat Methods},
	author = {Isensee, Fabian and Jaeger, Paul F. and Kohl, Simon A. A. and Petersen, Jens and Maier-Hein, Klaus H.},
	urldate = {2025-12-20},
	date = {2021-02},
	year = {2021},
	langid = {english},
	note = {Publisher: Nature Publishing Group},
}

@misc{wu2024deepmultimodallearningmissing,
      title={Deep Multimodal Learning with Missing Modality: A Survey}, 
      author={Renjie Wu and Hu Wang and Hsiang-Ting Chen and Gustavo Carneiro},
      year={2024},
      eprint={2409.07825},
      archivePrefix={arXiv},
      primaryClass={cs.CV},
      url={https://arxiv.org/abs/2409.07825}, 
}

@ARTICLE{azad2022medicalimagesegmentationmri,
  author={Azad, Reza and Dehghanmanshadi, Mohammad and Khosravi, Nika and Cohen-Adad, Julien and Merhof, Dorit},
  journal={Computational Visual Media}, 
  title={Addressing Missing Modality Challenges in {MRI} Images: A Comprehensive Review}, 
  year={2025},
  volume={11},
  number={2},
  pages={241-268},
  doi={10.26599/CVM.2025.9450399}}

@misc{brats1,
      title={The {RSNA-ASNR-MICCAI BraTS} 2021 Benchmark on Brain Tumor Segmentation and Radiogenomic Classification}, 
      author={Ujjwal Baid and Satyam Ghodasara and Suyash Mohan and Michel Bilello and Evan Calabrese and Errol Colak and others},
      year={2021},
      eprint={2107.02314},
      archivePrefix={arXiv},
      primaryClass={cs.CV},
      url={https://arxiv.org/abs/2107.02314}, 
}

@ARTICLE{brats2,
  author={Menze, Bjoern H. and Jakab, Andras and Bauer, Stefan and Kalpathy-Cramer, Jayashree and Farahani, Keyvan and Kirby, Justin and others},
  journal={IEEE Transactions on Medical Imaging}, 
  title={The Multimodal Brain Tumor Image Segmentation Benchmark ({BraTS})}, 
  year={2015},
  volume={34},
  number={10},
  pages={1993-2024},
  doi={10.1109/TMI.2014.2377694}}

@article{brats3,
	title = {Advancing {The} {Cancer} {Genome} {Atlas} glioma {MRI} collections with expert segmentation labels and radiomic features},
	volume = {4},
	copyright = {2017 The Author(s)},
	issn = {2052-4463},
	url = {https://www.nature.com/articles/sdata2017117},
	doi = {10.1038/sdata.2017.117},
	language = {en},
	number = {1},
	urldate = {2025-08-20},
	journal = {Scientific Data},
	author = {Bakas, Spyridon and Akbari, Hamed and Sotiras, Aristeidis and Bilello, Michel and Rozycki, Martin and Kirby, Justin S. and others},
	month = sep,
	year = {2017},
	note = {Publisher: Nature Publishing Group},
	pages = {170117},
}

@inproceedings{havaei_hemis_2016,
  title={{HeMIS}: Hetero-modal image segmentation},
  doi={10.1007/978-3-319-46723-8_54},
  author={Havaei, Mohammad and Guizard, Nicolas and Chapados, Nicolas and Bengio, Yoshua},
  booktitle={International conference on medical image computing and computer-assisted intervention},
  pages={469--477},
  year={2016},
  organization={Springer}
}

@inproceedings{liu2023m3ae,
  title={{M3AE}: Multimodal representation learning for brain tumor segmentation with missing modalities},
  author={Liu, Hong and Wei, Dong and Lu, Donghuan and Sun, Jinghan and Wang, Liansheng and Zheng, Yefeng},
  booktitle={Proceedings of the AAAI conference on artificial intelligence},
  volume={37},
  pages={1657--1665},
  year={2023}
}

@inproceedings{wang_multi-modal_2023,
  title={Multi-modal learning with missing modality via shared-specific feature modelling},
  author={Wang, Hu and Chen, Yuanhong and Ma, Congbo and Avery, Jodie and Hull, Louise and Carneiro, Gustavo},
  booktitle={Proceedings of the IEEE/CVF conference on computer vision and pattern recognition},
  pages={15878--15887},
  year={2023},
  doi={10.1109/CVPR52729.2023.01524}
}

@inproceedings{dorent_hetero-modal_2019,
	location = {Cham},
	title = {Hetero-Modal Variational Encoder-Decoder for Joint Modality Completion and Segmentation},
	isbn = {978-3-030-32245-8},
	doi = {10.1007/978-3-030-32245-8_9},
	pages = {74--82},
	booktitle = {Medical Image Computing and Computer Assisted Intervention – {MICCAI} 2019},
	publisher = {Springer International Publishing},
	author = {Dorent, Reuben and Joutard, Samuel and Modat, Marc and Ourselin, Sébastien and Vercauteren, Tom},
	editor = {Shen, Dinggang and Liu, Tianming and Peters, Terry M. and Staib, Lawrence H. and Essert, Caroline and Zhou, Sean and Yap, Pew-Thian and Khan, Ali},
	year = {2019},
	langid = {english},
}

@article{pemberton_multi-class_2023,
	title = {Multi-class glioma segmentation on real-world data with missing {MRI} sequences: comparison of three deep learning algorithms},
	volume = {13},
	copyright = {2023 The Author(s)},
	issn = {2045-2322},
	shorttitle = {Multi-class glioma segmentation on real-world data with missing {MRI} sequences},
	url = {https://www.nature.com/articles/s41598-023-44794-0},
	doi = {10.1038/s41598-023-44794-0},
	language = {en},
	number = {1},
	urldate = {2025-09-15},
	journal = {Scientific Reports},
	author = {Pemberton, Hugh G. and Wu, Jiaming and Kommers, Ivar and Müller, Domenique M. J. and Hu, Yipeng and Goodkin, Olivia and Vos, Sjoerd B. and Bisdas, Sotirios and Robe, Pierre A. and Ardon, Hilko and Bello, Lorenzo and Rossi, Marco and Sciortino, Tommaso and Nibali, Marco Conti and Berger, Mitchel S. and Hervey-Jumper, Shawn L. and Bouwknegt, Wim and Van den Brink, Wimar A. and Furtner, Julia and Han, Seunggu J. and Idema, Albert J. S. and Kiesel, Barbara and Widhalm, Georg and Kloet, Alfred and Wagemakers, Michiel and Zwinderman, Aeilko H. and Krieg, Sandro M. and Mandonnet, Emmanuel and Prados, Ferran and de Witt Hamer, Philip and Barkhof, Frederik and Eijgelaar, Roelant S.},
	month = nov,
	year = {2023},
	note = {Publisher: Nature Publishing Group},
	pages = {18911},
}

@INPROCEEDINGS{maheshwari_missing_2023,
  author={Maheshwari, Harsh and Liu, Yen-Cheng and Kira, Zsolt},
  booktitle={2024 IEEE/CVF Winter Conference on Applications of Computer Vision (WACV)}, 
  title={Missing Modality Robustness in Semi-Supervised Multi-Modal Semantic Segmentation}, 
  year={2024},
  volume={},
  number={},
  pages={1009-1019},
  doi={10.1109/WACV57701.2024.00106}}

@inproceedings{zhang_mmformer_2022,
  title={{mmFormer}: Multimodal medical transformer for incomplete multimodal learning of brain tumor segmentation},
  author={Zhang, Yao and He, Nanjun and Yang, Jiawei and Li, Yuexiang and Wei, Dong and Huang, Yawen and Zhang, Yang and He, Zhiqiang and Zheng, Yefeng},
  booktitle={International conference on medical image computing and computer-assisted intervention},
  pages={107--117},
  year={2022},
  organization={Springer}
}

@inproceedings{ashtari2020low,
  title={Low-rank convolutional networks for brain tumor segmentation},
  author={Ashtari, Pooya and Maes, Frederik and Van Huffel, Sabine},
  booktitle={International MICCAI Brainlesion Workshop},
  pages={470--480},
  year={2020},
  organization={Springer}
}

@inproceedings{ma_smil_2021,
  title={{SMIL}: Multimodal learning with severely missing modality},
  author={Ma, Mengmeng and Ren, Jian and Zhao, Long and Tulyakov, Sergey and Wu, Cathy and Peng, Xi},
  booktitle={Proceedings of the AAAI conference on artificial intelligence},
  volume={35},
  pages={2302--2310},
  year={2021}
}

@article{chartsias_multimodal_2018,
  title={Multimodal {MR} synthesis via modality-invariant latent representation},
  author={Chartsias, Agisilaos and Joyce, Thomas and Dharmakumar, Rohan and Tsaftaris, Sotirios A},
  journal={IEEE transactions on medical imaging},
  volume={37},
  number={3},
  pages={803--814},
  year={2018}
}

@article{dar_image_2019,
  title={Image synthesis in multi-contrast {MRI} with conditional generative adversarial networks},
  author={Dar, Salman UH and Yurt, Mahmut and Karacan, Levent and Erdem, Aykut and Erdem, Erkut and {\c{C}}ukur, Tolga},
  journal={IEEE transactions on medical imaging},
  volume={38},
  number={10},
  pages={2375--2388},
  year={2019}
}

@inproceedings{kim_compression_2015,
  title={Compression of deep convolutional neural networks for fast and low power mobile applications},
  author={Kim, Yong-Deok and Park, Eunhyeok and Yoo, Sungjoo and Choi, Taelim and Yang, Lu and Shin, Dongjun},
  booktitle={International Conference on Learning Representations},
  year={2015}
}

@inproceedings{lebedev_speeding_2014,
  title={Speeding-up convolutional neural networks using fine-tuned {CP}-decomposition},
  author={Lebedev, V and Ganin, Y and Rakhuba, M and Oseledets, I and Lempitsky, V},
  booktitle={3rd International Conference on Learning Representations, ICLR 2015-Conference Track Proceedings},
  year={2015}
}

@article{sidiropoulos_tensor_2017,
	title = {Tensor Decomposition for Signal Processing and Machine Learning},
	volume = {65},
	issn = {1053-587X, 1941-0476},
	url = {http://arxiv.org/abs/1607.01668},
	doi = {10.1109/TSP.2017.2690524},
	pages = {3551--3582},
	number = {13},
	journal = {{IEEE} Transactions on Signal Processing},
	shortjournal = {{IEEE} Trans. Signal Process.},
	author = {Sidiropoulos, Nicholas D. and De Lathauwer, Lieven  and Fu, Xiao and Huang, Kejun and Papalexakis, Evangelos E. and Faloutsos, Christos},
	urldate = {2025-09-18},
	date = {2017-07-01},
    year = {2017},
	eprinttype = {arxiv},
	eprint = {1607.01668 [stat]},
}

@article{kolda_tensor_2009,
	title = {Tensor Decompositions and Applications},
	volume = {51},
	url = {https://doi.org/10.1137/07070111X},
	doi = {10.1137/07070111X},
	pages = {455--500},
	number = {3},
	journal = {{SIAM} Review},
	author = {Kolda, Tamara G. and Bader, Brett W.},
	year = {2009},
}

@article{baltruvsaitis2018multimodal,
  title={Multimodal machine learning: A survey and taxonomy},
  author={Baltru{\v{s}}aitis, Tadas and Ahuja, Chaitanya and Morency, Louis-Philippe},
  journal={IEEE transactions on pattern analysis and machine intelligence},
  volume={41},
  number={2},
  pages={423--443},
  year={2018},
  publisher={IEEE}
}

@article{bayoudh2022survey,
  title={A survey on deep multimodal learning for computer vision: advances, trends, applications, and datasets},
  author={Bayoudh, Khaled and Knani, Raja and Hamdaoui, Fay{\c{c}}al and Mtibaa, Abdellatif},
  journal={The Visual Computer},
  volume={38},
  number={8},
  pages={2939--2970},
  year={2022},
  publisher={Springer}
}

@inproceedings{wang_learnable_2023,
	address = {Cham},
	title = {Learnable {Cross}-modal {Knowledge} {Distillation} for {Multi}-modal {Learning} with {Missing} {Modality}},
	isbn = {978-3-031-43901-8},
	doi = {10.1007/978-3-031-43901-8_21},
	language = {en},
	booktitle = {Medical {Image} {Computing} and {Computer} {Assisted} {Intervention} – {MICCAI} 2023},
	publisher = {Springer Nature Switzerland},
	author = {Wang, Hu and Ma, Congbo and Zhang, Jianpeng and Zhang, Yuan and Avery, Jodie and Hull, Louise and Carneiro, Gustavo},
	editor = {Greenspan, Hayit and Madabhushi, Anant and Mousavi, Parvin and Salcudean, Septimiu and Duncan, James and Syeda-Mahmood, Tanveer and Taylor, Russell},
	year = {2023},
	pages = {216--226},
}

@article{de_la_rosa_deepisles_2025,
	title = {{DeepISLES}: a clinically validated ischemic stroke segmentation model from the {ISLES}'22 challenge},
	volume = {16},
	issn = {2041-1723},
	doi = {10.1038/s41467-025-62373-x},
	shorttitle = {{DeepISLES}},
	pages = {7357},
	number = {1},
	journal = {Nature Communications},
	shortjournal = {Nat Commun},
	author = {de la Rosa, Ezequiel and Reyes, Mauricio and Liew, Sook-Lei and Hutton, Alexandre and Wiest, Roland and Kaesmacher, Johannes and Hanning, Uta and Hakim, Arsany and Zubal, Richard and Valenzuela, Waldo and Robben, David and Sima, Diana M. and Anania, Vincenzo and Brys, Arne and Meakin, James A. and Mickan, Anne and Broocks, Gabriel and Heitkamp, Christian and Gao, Shengbo and Liang, Kongming and Zhang, Ziji and Rahman Siddiquee, Md Mahfuzur and Myronenko, Andriy and Ashtari, Pooya and Van Huffel, Sabine and Jeong, Hyunsu and Yoon, Chiho and Kim, Chulhong and Huo, Jiayu and Ourselin, Sebastien and Sparks, Rachel and Clèrigues, Albert and Oliver, Arnau and Lladó, Xavier and Chalcroft, Liam and Pappas, Ioannis and Bertels, Jeroen and Heylen, Ewout and Moreau, Juliette and Hatami, Nima and Frindel, Carole and Qayyum, Abdul and Mazher, Moona and Puig, Domenec and Lin, Shao-Chieh and Juan, Chun-Jung and Hu, Tianxi and Boone, Lyndon and Goubran, Maged and Liu, Yi-Jui and Wegener, Susanne and Kofler, Florian and Ezhov, Ivan and Shit, Suprosanna and Hernandez Petzsche, Moritz R. and Müller, Michael and Menze, Bjoern and Kirschke, Jan S. and Wiestler, Benedikt},
	date = {2025-08-09},
    year = {2025},
	pmid = {40783484},
	pmcid = {PMC12335569},
}

@article{hernandez_petzsche_isles_2022,
	title = {{ISLES} 2022: A multi-center magnetic resonance imaging stroke lesion segmentation dataset},
	volume = {9},
	rights = {2022 The Author(s)},
	issn = {2052-4463},
	url = {https://www.nature.com/articles/s41597-022-01875-5},
	doi = {10.1038/s41597-022-01875-5},
	shorttitle = {{ISLES} 2022},
	pages = {762},
	number = {1},
	journal = {Scientific Data},
	shortjournal = {Sci Data},
	author = {Hernandez Petzsche, Moritz R. and de la Rosa, Ezequiel and Hanning, Uta and Wiest, Roland and Valenzuela, Waldo and Reyes, Mauricio and Meyer, Maria and Liew, Sook-Lei and Kofler, Florian and Ezhov, Ivan and Robben, David and Hutton, Alexandre and Friedrich, Tassilo and Zarth, Teresa and Bürkle, Johannes and Baran, The Anh and Menze, Björn and Broocks, Gabriel and Meyer, Lukas and Zimmer, Claus and Boeckh-Behrens, Tobias and Berndt, Maria and Ikenberg, Benno and Wiestler, Benedikt and Kirschke, Jan S.},
	urldate = {2025-11-06},
	date = {2022-12-10},
    year = {2022},
	langid = {english},
	note = {Publisher: Nature Publishing Group},
}

@inproceedings{li2025simmlm,
  title={{SimMLM}: A simple framework for multi-modal learning with missing modality},
  author={Li, Sijie and Chen, Chen and Han, Jungong},
  booktitle={Proceedings of the IEEE/CVF International Conference on Computer Vision},
  pages={24068--24077},
  year={2025}
}

@article{ha_hypernetworks_2017,
  title={Hypernetworks},
  author={Ha, David and Dai, Andrew and Le, Quoc V},
  journal={arXiv preprint arXiv:1609.09106},
  year={2016}
}

@article{debrabandere_dynamic_2016,
  title={Dynamic filter networks},
  author={Jia, Xu and De Brabandere, Bert and Tuytelaars, Tinne and Gool, Luc V},
  journal={Advances in neural information processing systems},
  volume={29},
  year={2016}
}

@article{zhang_graph_2019,
  title={Graph hypernetworks for neural architecture search},
  author={Zhang, Chris and Ren, Mengye and Urtasun, Raquel},
  journal={arXiv preprint arXiv:1810.05749},
  year={2018}
}

@article{nava_meta-learning_2023,
  title={Meta-learning via classifier (-free) diffusion guidance},
  author={Nava, Elvis and Kobayashi, Seijin and Yin, Yifei and Katzschmann, Robert K and Grewe, Benjamin F},
  journal={arXiv preprint arXiv:2210.08942},
  year={2022}
}

@inproceedings{zadeh_tensor_2017,
  title={Tensor fusion network for multimodal sentiment analysis},
  author={Zadeh, Amir and Chen, Minghai and Poria, Soujanya and Cambria, Erik and Morency, Louis-Philippe},
  booktitle={Proceedings of the 2017 conference on empirical methods in natural language processing},
  pages={1103--1114},
  year={2017}
}

@inproceedings{liu_efficient_2018,
  title={Efficient low-rank multimodal fusion with modality-specific factors},
  author={Liu, Zhun and Shen, Ying and Lakshminarasimhan, Varun Bharadhwaj and Liang, Paul Pu and Zadeh, AmirAli Bagher and Morency, Louis-Philippe},
  booktitle={Proceedings of the 56th Annual Meeting of the Association for Computational Linguistics (Volume 1: Long Papers)},
  pages={2247--2256},
  year={2018}
}

@article{zhang_tdsf_2025,
  title={TDSF-Net: Tensor decomposition-based subspace fusion network for multimodal medical image classification},
  author={Zhang, Yi and Xu, Guoxia and Zhao, Meng and Wang, Hao and Shi, Fan and Chen, Shengyong},
  journal={IEEE Transactions on Neural Networks and Learning Systems},
  year={2025},
  publisher={IEEE}
}

@inproceedings{zhang_ls3m_2025,
  title={Incomplete multi-modal brain tumor segmentation via learnable sorting state space model},
  author={Zhang, Zheyu and Lu, Yayuan and Ma, Feipeng and Zhang, Yueyi and Yue, Huanjing and Sun, Xiaoyan},
  booktitle={Proceedings of the IEEE/CVF Conference on Computer Vision and Pattern Recognition},
  year={2025},
  doi={10.1109/CVPR52734.2025.02420}
}

@article{shi_hga_2026,
  title={Addressing imbalanced modal incompleteness in realistic multi-modal medical image segmentation via hierarchical gradient alignment},
  author={Shi, Junjie and Sun, Zhaobin and Yu, Li and Yang, Xin and Yan, Zengqiang},
  journal={IEEE Transactions on Medical Imaging},
  volume={45},
  number={6},
  pages={3066--3081},
  year={2026},
  doi={10.1109/TMI.2026.3667605}
}

@inproceedings{song_unime_2026,
  title={Uni-encoder meets multi-encoders: Representation before fusion for brain tumor segmentation with missing modalities},
  author={Song, Peibo and Xue, Xiaotian and Zhang, Jinshuo and Wang, Zihao and Liu, Jinhua and Fu, Shujun and Bao, Fangxun and Yeo, Si Yong},
  booktitle={Proceedings of the IEEE/CVF Conference on Computer Vision and Pattern Recognition},
  year={2026}
}
\bibliographystyle{plainnat}

\clearpage
\appendix

\section{Backbone architecture}
\label{sec:backbone-arch}

\begin{wrapfigure}[34]{r}{0.5\textwidth}
    \vspace{-16pt}
    \centering
    \begin{tikzpicture}[
        block/.style={draw, rounded corners=2pt, minimum width=2.1cm, minimum height=0.36cm, inner sep=1pt, font=\scriptsize},
        lbl/.style={font=\scriptsize, inner sep=1pt},
        arrow/.style={->, thick, >=Stealth}
    ]
        \foreach \i/\y/\name/\c in {1/0/Stem block/orange!20, 2/-1.7/Downsample/red!20, 3/-3.4/Downsample/red!20, 4/-5.1/Downsample/red!20}{
            \node[block, fill=\c, anchor=north west] at (0,\y) {\name};
            \node[block, fill=gray!20, anchor=north west] at (0,\y-0.44) {Conv block \i};
        }
        \node[block, fill=red!20, anchor=north west] at (2.3,-6.8) {Downsample};
        \node[block, fill=pink!30, anchor=north west] at (2.3,-7.24) {Bridge block};
        \foreach \i/\y in {4/0, 3/-1.7, 2/-3.4, 1/-5.1}{
            \node[block, fill=gray!20, anchor=north west] at (4.7,\y) {Decoder block \i};
            \node[block, fill=green!20, anchor=north west] at (4.7,\y-0.44) {Concatenate};
            \node[block, fill=cyan!20, anchor=north west] at (4.7,\y-0.88) {Upsample};
        }
        \node[block, fill=purple!20, anchor=south west] at (4.7,0.5) {Head block};
        \draw[arrow] (1.05,1.0) -- (1.05,0) node[lbl, midway, right] {$4\times128^3$};
        \draw[arrow] (5.75,0) -- (5.75,0.5);
        \draw[arrow] (5.75,0.86) -- (5.75,1.5) node[lbl, midway, right] {$3\times128^3$};
        \draw[arrow] (1.05,-0.8) -- (1.05,-1.7) node[lbl, midway, right] {$32\times128^3$};
        \draw[arrow] (1.05,-2.5) -- (1.05,-3.4) node[lbl, midway, right] {$64\times64^3$};
        \draw[arrow] (1.05,-4.2) -- (1.05,-5.1) node[lbl, midway, right] {$128\times32^3$};
        \draw[arrow, rounded corners] (1.05,-5.9) -- (1.05,-6.55) -- (3.35,-6.55) -- (3.35,-6.8);
        \node[lbl, anchor=south west] at (1.15,-6.5) {$256\times16^3$};
        \draw[arrow, rounded corners] (3.35,-7.6) -- (3.35,-7.85) -- (5.75,-7.85) -- (5.75,-6.34);
        \node[lbl, anchor=west] at (5.85,-7.1) {$512\times8^3$};
        \draw[arrow] (5.75,-5.1) -- (5.75,-4.64);
        \draw[arrow] (5.75,-3.4) -- (5.75,-2.94);
        \draw[arrow] (5.75,-1.7) -- (5.75,-1.24);
        \foreach \y/\d in {-0.62/128, -2.32/64, -4.02/32, -5.72/16}{
            \draw[arrow] (2.1,\y) -- (4.7,\y);
        }
        \node[lbl, above] at (3.4,-0.62) {$32\times128^3$};
        \node[lbl, above] at (3.4,-2.32) {$64\times64^3$};
        \node[lbl, above] at (3.4,-4.02) {$128\times32^3$};
        \node[lbl, above] at (3.4,-5.72) {$256\times16^3$};
        \draw[rounded corners=3pt, thick] (0,-8.2) rectangle (6.8,-10.05);
        \foreach \y/\c/\n/\d in {-8.5/orange!20/Stem/Conv $3^3$, -8.95/red!20/Downsample/{Conv $2^3$, str.\ 2}, -9.4/gray!20/Conv block/Conv $3^3$}{
            \node[draw, rounded corners=2pt, fill=\c, minimum width=1.5cm, minimum height=0.3cm, inner sep=1pt, font=\scriptsize] at (0.95,\y) {\n};
            \node[lbl, anchor=west] at (1.8,\y) {\d};
        }
        \foreach \y/\c/\n/\d in {-8.5/purple!20/Head/Conv $1^3$, -8.95/cyan!20/Upsample/Transp.\ conv $2^3$, -9.4/green!20/Concatenate/Skip connection}{
            \node[draw, rounded corners=2pt, fill=\c, minimum width=1.5cm, minimum height=0.3cm, inner sep=1pt, font=\scriptsize] at (4.2,\y) {\n};
            \node[lbl, anchor=west] at (5.05,\y) {\d};
        }
        \node[lbl, anchor=west] at (0.15,-9.82) {All conv layers use instance normalization and leaky ReLU.};
    \end{tikzpicture}
    \caption{Configuration of the nnU-Net architecture used for the BraTS 2018 dataset. The encoder path (left) progressively downsamples the input through conv blocks with increasing channels. Skip connections transfer encoder features to the decoder path (right), which upsamples back to the original resolution. Numbers indicate channels$\times$spatial resolution.}
    \label{fig:brats-architecture}
\end{wrapfigure}
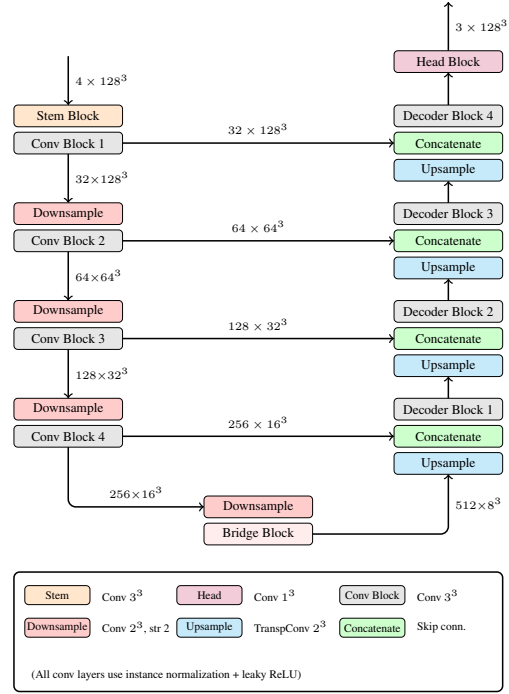

This section describes the base network architectures used in the BraTS 2018, ISLES 2022, and avMNIST experiments. Both architectures employ low-rank tensor decomposition in their convolutional and linear layers to enable parameter-efficient handling of missing modalities.

\subsection{3D segmentation network architecture}

For the BraTS 2018 and ISLES 2022 datasets, we employ a U-Net architecture based on the nnU-Net framework~\citep{isensee_nnu-net_2021}. The network follows the standard nnU-Net configuration with a symmetric encoder-decoder structure.

\paragraph {BraTS 2018 dataset.} The encoder consists of four downsampling stages with progressively increasing channel dimensions (32, 64, 128, 256, 512) and decreasing spatial resolutions ($128^3 \rightarrow 64^3 \rightarrow 32^3 \rightarrow 16^3 \rightarrow 8^3$). Each encoder block contains two 3D convolutions ($3\times3\times3$ kernels) with instance normalization and leaky ReLU activations, followed by strided convolution for downsampling.

The decoder mirrors the encoder structure with four upsampling stages using transposed convolutions. The final output layer produces a single-channel segmentation mask at the original $128^3$ resolution.

The network architecture is visually represented in \cref{fig:brats-architecture}.

\paragraph{ISLES 2022 dataset.}For the ISLES 2022 dataset, the network has one fewer downsample and upsample layer. The downsampling layers progressively increase the channel dimensions (64, 128, 256, 512) and decrease the spatial resolution ($64^3 \rightarrow 32^3 \rightarrow 16^3 \rightarrow 8^3$).

\subsection{avMNIST network architecture}

For the avMNIST digit classification task, we employ a fusion-based architecture that processes two modalities (image and audio) and handles $2^2 - 1 = 3$ modality combinations (image-only, audio-only, and both).

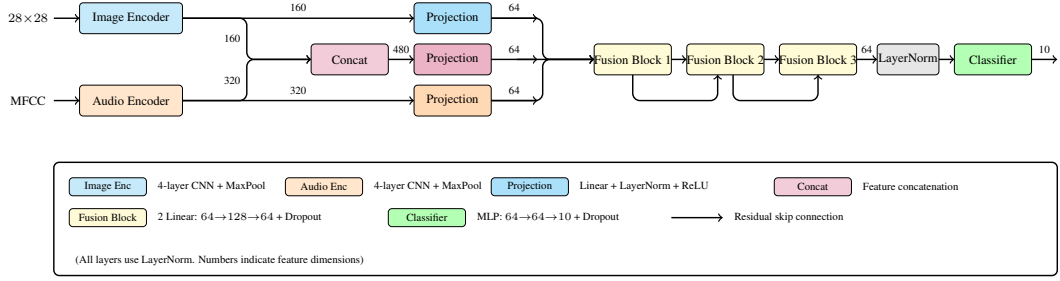
\begin{figure}
    \centering
    \begin{tikzpicture}[
        block/.style={draw, rounded corners=2pt, minimum width=1.9cm, minimum height=0.42cm, inner sep=1pt, font=\scriptsize},
        lbl/.style={font=\scriptsize, inner sep=1pt},
        arrow/.style={->, thick, >=Stealth}
    ]
        \node[lbl] at (0,0.8) {$28\times28$};
        \node[lbl] at (0,-0.8) {MFCC};
        \node[block, fill=cyan!20] (ie) at (1.9,0.8) {Image encoder};
        \node[block, fill=orange!20] (ae) at (1.9,-0.8) {Audio encoder};
        \draw[arrow] (0.5,0.8) -- (ie.west);
        \draw[arrow] (0.5,-0.8) -- (ae.west);
        \node[block, fill=purple!20, minimum width=1.4cm] (cat) at (5.2,0) {Concat};
        \node[block, fill=cyan!30] (pi) at (8.0,0.8) {Projection};
        \node[block, fill=purple!30] (pc) at (8.0,0) {Projection};
        \node[block, fill=orange!30] (pa) at (8.0,-0.8) {Projection};
        \draw[arrow] (ie.east) -- (pi.west) node[lbl, pos=0.6, above] {160};
        \draw[arrow] (ae.east) -- (pa.west) node[lbl, pos=0.6, below] {320};
        \draw[arrow, rounded corners] (3.6,0.8) -- (3.6,0) -- (cat.west);
        \draw[arrow, rounded corners] (3.6,-0.8) -- (3.6,0) -- (cat.west);
        \draw[arrow] (cat.east) -- (pc.west) node[lbl, midway, above] {480};
        \draw[arrow, rounded corners] (pi.east) -- (9.5,0.8) -- (9.5,0) -- (10.0,0);
        \draw[arrow, rounded corners] (pa.east) -- (9.5,-0.8) -- (9.5,0) -- (10.0,0);
        \draw[arrow] (pc.east) -- (10.0,0);
        \draw[arrow, rounded corners] (10.0,0) -- (10.4,0) -- (10.4,-1.45) -- (0.4,-1.45) -- (0.4,-2.4) -- (0.9,-2.4);
        \node[lbl, anchor=west] at (10.5,-0.75) {64};
        \node[block, fill=yellow!20, anchor=west] (f1) at (0.9,-2.4) {Fusion block 1};
        \node[block, fill=yellow!20, anchor=west] (f2) at (3.2,-2.4) {Fusion block 2};
        \node[block, fill=yellow!20, anchor=west] (f3) at (5.5,-2.4) {Fusion block 3};
        \node[block, fill=gray!20, minimum width=1.5cm, anchor=west] (ln) at (7.8,-2.4) {LayerNorm};
        \node[block, fill=green!30, minimum width=1.5cm, anchor=west] (cl) at (9.7,-2.4) {Classifier};
        \draw[arrow] (f1.east) -- (f2.west);
        \draw[arrow] (f2.east) -- (f3.west);
        \draw[arrow] (f3.east) -- (ln.west);
        \draw[arrow] (ln.east) -- (cl.west);
        \draw[arrow] (cl.east) -- ++(0.6,0) node[lbl, midway, above] {10};
        \draw[arrow, rounded corners=4pt] (f1.south) -- ++(0,-0.35) -| (f2.south);
        \draw[arrow, rounded corners=4pt] (f2.south) -- ++(0,-0.35) -| (f3.south);
        \draw[rounded corners=3pt, thick] (0,-3.3) rectangle (12.2,-4.8);
        \foreach \x/\y/\c/\n/\d in {0.95/-3.6/cyan!20/Image enc./4-layer CNN + max pooling, 0.95/-4.05/orange!20/Audio enc./4-layer CNN + max pooling, 0.95/-4.5/cyan!30/Projection/Linear + LayerNorm + ReLU, 6.6/-3.6/purple!20/Concat/Feature concatenation, 6.6/-4.05/yellow!20/Fusion block/2 linear ($64{\to}128{\to}64$) + dropout, 6.6/-4.5/green!30/Classifier/MLP ($64{\to}64{\to}10$) + dropout}{
            \node[draw, rounded corners=2pt, fill=\c, minimum width=1.6cm, minimum height=0.3cm, inner sep=1pt, font=\scriptsize] at (\x,\y) {\n};
            \node[lbl, anchor=west] at (\x+0.9,\y) {\d};
        }
    \end{tikzpicture}
    \caption{Configuration of the fusion architecture used for the avMNIST dataset. The image and audio encoders independently extract features. Features from available modalities are concatenated and projected on a 64-dimensional feature vector. Three fusion blocks with residual connections process the combined features, followed by layer normalization and a classifier producing digit predictions.}
    \label{fig:avmnist-architecture}
\end{figure}

The architecture follows a modality-specific encoding followed by fusion paradigm, illustrated in \cref{fig:avmnist-architecture}.

\paragraph{Modality-specific encoders.} Separate convolutional encoders process each modality independently. The image encoder processes $28 \times 28$ grayscale MNIST images through a 4-layer CNN with progressively increasing channels (16, 32, 64, 128), producing a 160-dimensional feature vector. The audio encoder processes MFCC spectrograms through a similar 4-layer CNN architecture with adapted dimensions, producing a 320-dimensional feature vector.

\paragraph{Projection and fusion.} The inputs from the available modalities are concatenated and projected by a unique projection block on a common 64-dimensional feature vector.

\paragraph{Deep fusion network.} The projected features pass through three fusion blocks with residual connections. Each block consists of two linear transformations ($64 \rightarrow 128 \rightarrow 64$) with layer normalization, ReLU activations, and dropout for regularization. A final layer normalization is applied before classification.

\paragraph{Classification head.} The classification head produces final predictions for the different digit classes. Each head consists of a two-layer MLP with dropout, allowing the network to learn combination-specific decision boundaries.

\paragraph{CPD compression of linear layers.} For linear layers, the joint weight tensor can be represented as $\tns{W}\in\R^{M\times C_\text{in}\times C_\text{out}}$. The CPD decomposition is then similar to \cref{eq:CPD-conv}, discarding the matrix $\mtx{D}$. The rank selection is equivalent to \cref{eq:rank}, discarding $K$.

\begin{wraptable}[11]{r}{0.56\textwidth}
    \vspace{-16pt}
    \caption{Dice scores (\%)~$\uparrow$ on ISLES 2022 (5-fold cross-validation) for different values of the rank $R$ and for the dedicated models (Ded.). $\bullet$/$\circ$: modality present/absent.}
    \centering
    \setlength{\tabcolsep}{2.5pt}%
    {\scriptsize\begin{tabular*}{\linewidth}{@{\extracolsep{\fill}}ccccccccc@{}}
        \toprule
        \multirow{2}{*}{\textbf{DWI}} & \multirow{2}{*}{\textbf{ADC}} & \multirow{2}{*}{\textbf{FLAIR}} & \multicolumn{6}{c}{\textbf{Stroke Lesion}}\\
         &  & & $R/4$ & $R/2$ & $R$ & $2R$ & $7R$ & Ded. \\
        \midrule
        $\bullet$ & $\circ$ & $\circ$ & 73.52 & 74.79 & 74.23 & 74.96 & 74.79 & 74.22 \\
        $\circ$ & $\bullet$ & $\circ$ & 39.74 & 41.12 & 42.49 & 43.06 & 43.95 & 45.08 \\
        $\circ$ & $\circ$ & $\bullet$ & 38.35 & 39.27 & 41.09 & 41.91 & 41.73 & 41.35 \\
        $\bullet$ & $\bullet$ & $\circ$  & 74.58 & 75.23 & 74.67 & 75.32 & 76.10 & 75.11 \\
        $\bullet$ & $\circ$ & $\bullet$ & 73.59 & 74.57 & 75.75 & 75.49 & 75.63 & 74.73 \\
        $\circ$ & $\bullet$ & $\bullet$ & 51.80 & 52.80 & 53.70 & 53.92 & 54.83 & 54.90 \\
        $\bullet$ & $\bullet$ & $\bullet$ &  75.92 & 76.85 & 76.82 & 77.14 & 76.45 & 74.89 \\
        \midrule
        \multicolumn{3}{c}{Average} & 61.07 & 62.09 & 62.68 & 63.11 & 63.36 & 62.90 \\
        \bottomrule
    \end{tabular*}}
    \label{tab:ablation-rank}
\end{wraptable}

\section{Ablation studies}

This section analyzes the impact of compression-related hyperparameters on model performance. We examine rank selection and compare different tensor decomposition methods on the ISLES 2022 dataset using 5-fold cross-validation.

\subsection{Impact of rank on performance}
\label{sec:ablation-rank}

The rank $R$ directly controls the compression ratio in our decomposition. As defined in \cref{sec:method}, we set $R$ to achieve a compression ratio of $1/M$ relative to dedicated models, where $M$ is the number of modality combinations. Lower ranks reduce parameters but limit model capacity, while higher ranks increase parameters.

\cref{fig:ablation-rank} and \cref{tab:ablation-rank} present results for five rank settings: $R/4$, $R/2$, $R$ (our default), $2R$, and $7R$. We also include dedicated models for comparison, which have approximately the same parameter count as the $7R$ configuration. The dedicated models were trained as a hypernetwork, with separate models and the same training procedure. The average Dice score increases from 61.07\% at $R/4$ to 63.36\% at $7R$, while the Hausdorff\textsuperscript{95} distance decreases from 11.51mm to 10.0mm. This trend indicates that higher compression (lower rank) reduces accuracy, as expected.

At $7R$, LARGO achieves 63.36\% Dice score compared to 62.90\% for dedicated models with similar parameter counts. The default rank $R$ achieves 62.68\% Dice score with $1/M$ compression relative to dedicated models.

\begin{wrapfigure}[21]{r}{0.5\textwidth}
    \vspace{-16pt}
    \centering
    \resizebox{\linewidth}{!}{
    \begin{tikzpicture}
\begin{axis}[
    title={},
    xlabel={Rank},
    ylabel={Average Dice Score (\%)},
    ylabel style={
        color=blue
    },
    xmin=0, xmax=8.1,
    ymin=60, ymax=64,
    xtick={0,0.25,0.5,1,2,7,8},
    xticklabels={0,$\frac{R}{4}$,$\frac{R}{2}$,$R$,$2R$,$7R$,Ded.},
    ytick={60,60.5,61,61.5,62,62.5,63,63.5,64},
    yticklabel style={color=blue},
    legend pos=north west,
    ymajorgrids=true,
    grid style=dashed,
]

\addplot[
    color=blue,
    mark=o,
    ]
    coordinates {
    (0.25,61.07)(0.5,62.09)(1,62.68)(2,63.11)(7,63.36)
    };
    \addplot[
    color=blue,
    mark=o,
    dashed,
    mark options={solid}
    ]
    coordinates {
    (7,63.36)(8,62.90)
    };
    
\end{axis}
\begin{axis}[
    xmin=0, xmax=8.1,
    ymin=10, ymax=12,
    axis y line*=right,
    axis x line=none,
    ylabel near ticks, yticklabel pos=right,
    ylabel={Average Hausdorff\textsuperscript{95} Distance (mm)},
    ylabel style={
        anchor=south,
        rotate=180,
        color=red
    },
    ytick={10,10.5,11,11.5,12},
    yticklabel style={color=red},
    grid style=dashed,
]

\addplot[
    color=red,
    mark=x,
] coordinates {
    (0.25,11.51)(0.5,11.18)(1,10.91)(2,10.4)(7,10)
};
\addplot[
    color=red,
    mark=x,
    dashed,
    mark options={solid}
] coordinates {
    (7,10)(8,11.03)
};

\end{axis}
\end{tikzpicture}}
    \caption{Ablation study on the rank $R$ for the ISLES 2022 dataset using 5-fold cross-validation. We present the values $R/4$, $R/2$, $R$, $2R$, and $7R$, with $R$ as in \cref{eq:rank}, as well as the performance of the dedicated models (Ded.), which is approximately equivalent in terms of number of parameters to the $7R$ case. The points for the average Dice score ($\uparrow$) and Hausdorff\textsuperscript{95} distance ($\downarrow$) are represented in blue and red, respectively.}
    \label{fig:ablation-rank}
\end{wrapfigure}

\subsection{Comparison of tensor decompositions}

We compare the canonical polyadic decomposition (CPD) used in our method against the Tucker decomposition, another common tensor decomposition approach. The Tucker decomposition, illustrated in \cref{fig:conv-reparameterization-tucker}, uses a core tensor and factor matrices along each mode. The modes $M$ and $K$ are not compressed, since these are relatively small compared to $C_\text{in}$ and $C_\text{out}$. The modes $C_\text{in}$ and $C_\text{out}$ share the same rank $R$. We set the rank for the $1/M$ compression as follows:

\begin{equation}
R=\left\lfloor
\frac{
- \left( C_{\text{in}} + C_{\text{out}} \right)
+ \sqrt{\Delta
}
}{
2 M K
}
\right\rfloor,
\end{equation}

with discriminant $
    \Delta
=
\left( C_{\text{in}} + C_{\text{out}} \right)^2
- 4 M K \left( M^2 + K^2 - K\, C_{\text{in}} C_{\text{out}} \right)\geq 0$.

\cref{tab:ablation-tensor-decomp} compares CPD and Tucker decompositions with matched parameter budgets. CPD achieves an average Dice score of 62.68\% compared to 61.97\% for Tucker. The difference is larger for single-modality scenarios: FLAIR-only shows 41.09\% vs 37.80\% (3.29 percentage points), and ADC+FLAIR shows 53.70\% vs 52.35\% (1.35 points). For the complete three-modality case, both methods perform similarly (76.82\% vs 77.07\%).

\begin{figure}[h]
    \centering
    \begin{minipage}[c]{0.48\textwidth}
        \centering
        \resizebox{\linewidth}{!}{\begin{tikzpicture}
            \fill[fill=red!10, rounded corners] (0,0) rectangle (3,-3.5);
            \node[anchor=north,yshift=-5pt] at (1.5,0) {\bf \color{red!80} Models};
            \fill[fill=blue!10, rounded corners](4,0) rectangle (7, -3.5);
            \node[anchor=north,yshift=-5pt] at (5.5,0) {\bf \color{blue!80} Kernels};
            \begin{scope}[shift={(0,1)}]
            \draw (2, -3) -- (1,-3);
            \draw (2,-3) -- (3.5,-3);
            \draw (5, -2) .. controls (4, -2) .. (3.5,-3);
            \draw (5, -3) -- (3.5,-3);
            \draw (5, -4) .. controls (4,-4) .. (3.5,-3);

            \draw (5, -2) -- (6,-2);
            \draw (5, -3) -- (6,-3);
            \draw (5, -4) -- (6,-4);

            \draw[fill=black] (2.75, -3) circle (0.05) node[above] {$m$};
            \draw[fill=black] (4.25, -3) circle (0.05) node[above] {$r$};
            \draw[fill=black] (4.25, -2.05) circle (0.05) node[above] {$r$};
            \draw[fill=black] (4.25, -3.95) circle (0.05) node[above] {$k$};

            \draw (3.5,-3)[fill=purple!30] circle (0.3);
            \draw (2,-3)[fill=red!30] circle (0.3);
            \draw (5,-2)[fill=blue!30] circle (0.3);
            \draw (5,-3)[fill=blue!30] circle (0.3);
            \draw (5,-4)[fill=blue!30] circle (0.3);
            \node at (2, -3) {$\boldsymbol{A}$};
            \node at (0.5, -3) {$m$};
            \node at (3.5, -3) {$\boldsymbol{\mathcal{G}}$};

            \node at (5, -2) {$\boldsymbol{B}$};
            \node at (5, -3) {$\boldsymbol{C}$};
            \node at (5, -4) {$\boldsymbol{D}$};
            \node at (6.5, -2) {$c_\text{in}$};
            \node at (6.5, -3) {$c_\text{out}$};
            \node at (6.5, -4) {$k$};
            \end{scope}
        \end{tikzpicture}}
        \caption{Tensor diagram of the Tucker reparameterization of the convolutional and transposed convolutional layers.}
        \label{fig:conv-reparameterization-tucker}
    \end{minipage}\hfill
    \begin{minipage}[c]{0.48\textwidth}
        \centering
        \scriptsize
        \captionof{table}{Comparison of Dice scores (\%)~$\uparrow$ on the ISLES 2022 dataset using 5-fold cross-validation for CPD and Tucker decomposition. Symbols $\bullet$ and $\circ$ denote the presence and absence of a modality, respectively.}
        \label{tab:ablation-tensor-decomp}
        \begin{tabularx}{\linewidth}{cccYY}
        \toprule
        \multirow{2}{*}{\textbf{DWI}} &
        \multirow{2}{*}{\textbf{ADC}} &
        \multirow{2}{*}{\textbf{FLAIR}} &
        \multicolumn{2}{c}{\textbf{Stroke Lesion}} \\
         & & & {CPD} & {Tucker} \\
        \midrule
        $\bullet$ & $\circ$ & $\circ$ & 74.23 & 74.08 \\
        $\circ$ & $\bullet$ & $\circ$ & 42.49 & 42.65 \\
        $\circ$ & $\circ$ & $\bullet$ & 41.09 & 37.80 \\
        $\bullet$ & $\bullet$ & $\circ$ & 74.67 & 74.62 \\
        $\bullet$ & $\circ$ & $\bullet$ & 75.75 & 75.19 \\
        $\circ$ & $\bullet$ & $\bullet$ & 53.70 & 52.35 \\
        $\bullet$ & $\bullet$ & $\bullet$ & 76.82 & 77.07 \\
        \midrule
        \multicolumn{3}{c}{Average} & 62.68 & 61.97 \\
        \bottomrule
        \end{tabularx}
    \end{minipage}
\end{figure}

\clearpage
\section{Statistical analysis}
\label{sec:suppl-stats}

\begin{revblock}
\Cref{tab:significance} reports the patient-level comparison of LARGO against each baseline described in \cref{sec:experiments}. For every patient, the Dice score is averaged over all modality subsets (and tumor regions for BraTS 2018), which yields one paired observation per patient and method. We apply two-sided paired Wilcoxon signed-rank tests and correct the $p$-values with the Benjamini--Hochberg procedure. All aggregate comparisons remain significant after correction; at the level of individual modality subsets, 57 of 60 comparisons on BraTS 2018 and 26 of 28 on ISLES 2022 remain significant.
\end{revblock}

\begin{table}[h]
    \centering
    \caption{\rev{Patient-level statistical comparison of LARGO against each baseline. $\Delta$Dice is the average gain of LARGO in percentage points; ``won'' is the number of patients for which LARGO obtains the higher subset-averaged Dice score; $p$-values are from two-sided paired Wilcoxon signed-rank tests after Benjamini--Hochberg correction.}}
    \label{tab:significance}
    {\scriptsize\begin{tabular*}{0.8\linewidth}{@{\extracolsep{\fill}}lcccccc@{}}
        \toprule
        \multirow{2}{*}{LARGO vs.} & \multicolumn{3}{c}{\textbf{BraTS 2018} (285 patients)} & \multicolumn{3}{c}{\textbf{ISLES 2022} (250 patients)} \\
         & $\Delta$Dice & Won & Corrected $p$ & $\Delta$Dice & Won & Corrected $p$ \\
        \midrule
        mmFormer & $+3.13$ & 243/285 & $2.7\times10^{-29}$ & $+5.79$  & 198/250 & $1.3\times10^{-21}$ \\
        M³AE     & $+2.05$ & 223/285 & $2.0\times10^{-19}$ & $+2.53$  & 181/250 & $3.1\times10^{-14}$ \\
        ShaSpec  & $+0.69$ & 185/285 & $5.8\times10^{-6}$  & $+11.75$ & 240/250 & $3.6\times10^{-40}$ \\
        SimMLM   & $+1.82$ & 207/285 & $9.2\times10^{-16}$ & $+2.65$  & 190/250 & $1.9\times10^{-15}$ \\
        \bottomrule
    \end{tabular*}}
\end{table}

\section{Inference cost and architectural scaling}
\label{sec:suppl-scaling}

\begin{revblock}
This section complements \cref{sec:complexity} of the main paper.
\end{revblock}

\begin{wraptable}[8]{r}{0.42\linewidth}
    \vspace{-14pt}
    \centering
    \caption{\rev{Inference cost of LARGO vs.\ number of available modalities (BraTS 2018, $128^3$ patch).}}
    \label{tab:cost-vs-modalities}
    \setlength{\tabcolsep}{3pt}%
    {\scriptsize\begin{tabular*}{\linewidth}{@{\extracolsep{\fill}}ccc@{}}
        \toprule
        Available modalities & GFLOPs & Time per patch \\
        \midrule
        1 & 1144.3 & 65.8 ms \\
        2 & 1147.9 & 66.1 ms \\
        3 & 1151.5 & 66.2 ms \\
        4 & 1155.1 & 66.6 ms \\
        \bottomrule
    \end{tabular*}}
\end{wraptable}
\leavevmode\begin{revblock}
\Cref{tab:cost-vs-modalities} reports GFLOPs and inference time per $128^3$ patch of LARGO on BraTS 2018 as a function of the number of available modalities. Both vary by less than 1\% (1144.3--1155.1 GFLOPs, 65.8--66.6 ms), since only the uncompressed stem depends on the number of input channels; reconstructing the inner-layer kernels and applying them cost the same for every subset. The four-modality value is the one reported in \cref{tab:comparison-brats}.
\end{revblock}

\begin{wraptable}[10]{r}{0.58\linewidth}
    \centering
    \caption{\rev{Architectural scaling of LARGO with the number of modalities $N$ (BraTS 2018 backbone), with the per-layer rank recomputed from \cref{eq:rank}.}}
    \label{tab:scaling}
    \setlength{\tabcolsep}{3pt}%
    {\scriptsize\begin{tabular*}{\linewidth}{@{\extracolsep{\fill}}cccccc@{}}
        \toprule
        $N$ & $M=2^N-1$ & Params (M) & Stem/head share & GFLOPs & Rank range \\
        \midrule
        4  & 15   & 22.60 & 0.13\%  & 1155.1 & 137--6639 \\
        6  & 63   & 22.74 & 0.77\%  & 1151.5 & 98--6353 \\
        8  & 255  & 23.48 & 3.91\%  & 1126.1 & 45--5419 \\
        10 & 1023 & 27.11 & 16.81\% & 1072.3 & 14--3412 \\
        \bottomrule
    \end{tabular*}}
\end{wraptable}

\leavevmode\begin{revblock}
\Cref{tab:scaling} reports how parameters, the stem/head share, GFLOPs, and the per-layer rank range change when the number of modalities $N$ grows and the rank is recomputed from \cref{eq:rank}. Going from $N=4$ to $N=8$ ($M=255$ subsets) raises the parameter count by only 3.9\%, while GFLOPs decrease with the lower ranks. At $N=10$ the uncompressed stem and head reach 16.8\% of the budget, which motivates factorizing them as discussed in \cref{sec:complexity}. These numbers are architectural evidence and do not establish that accuracy is preserved for $N>4$.
\end{revblock}



\end{document}